\pdfoutput=1
\RequirePackage[T1]{fontenc}
\documentclass[12pt]{article}
\usepackage[skip=1em plus 0.2em minus 0.1em]{parskip}
\usepackage[height=8.85in,width=6.45in]{geometry}

\usepackage[utf8]{inputenc}
\usepackage{amsmath}
\usepackage{amssymb}
\usepackage{mathtools}
\numberwithin{equation}{section}
\usepackage{slashed}
\usepackage{braket}
\usepackage{enumitem}
\usepackage[svgnames]{xcolor}
\usepackage[colorlinks,citecolor=DarkGreen,linkcolor=FireBrick,urlcolor=FireBrick,linktocpage,unicode]{hyperref}

\usepackage{tocloft}

\usepackage[bottom]{footmisc}
\allowdisplaybreaks
\usepackage{graphicx}
\usepackage{tikz}
\usepackage{tikz-cd}
\usepackage{times}
 
\usepackage{bm}
\usepackage{physics}
\usepackage{xcolor}
\usepackage{natbib}
\usepackage{mdframed}
\usepackage{nicefrac}
\usepackage{booktabs}
\usepackage{lipsum}
\usepackage{titlesec}
\usepackage{wrapfig,lipsum,booktabs}
\usepackage{authblk}
\usepackage{blindtext}
\usepackage{subcaption}

\usepackage{algorithm}
\usepackage{algpseudocode}
\newcommand{\commentsymbol}{//}%
\algrenewcommand\algorithmiccomment[1]{\hfill {\footnotesize \commentsymbol{} #1}}

\usepackage{titletoc}
\usepackage{todonotes}
\usepackage{setspace}
\usepackage{dsfont} %
\newcommand{\one}{\mathds{1}}

\usepackage[nameinlink,capitalize,noabbrev]{cleveref}

\newenvironment{claimbox}
{%
  \begin{mdframed}[
    linecolor=black!0,       %
    linewidth=0pt,           %
    backgroundcolor=black!10, %
    innertopmargin=8pt,
    innerbottommargin=8pt,
    innerleftmargin=10pt,
    innerrightmargin=10pt
  ]
}
{\end{mdframed}}

\usepackage{array}
\usepackage{bbm}
\usepackage{makecell}

\usepackage{dashbox}
\usepackage{xcolor}
\usepackage{colortbl}
\definecolor{lightyellow}{rgb}{1.0, 0.95, 0.7}
\definecolor{Blue}{rgb}{0, 0, 0.8}
\definecolor{blue}{rgb}{0,0,1}
\definecolor{darkgreen}{rgb}{0,0.40,0}
\definecolor{firebrick}{rgb}{0.698,0.133,0.133}

\definecolor{colorA}{rgb}{1,0,0}
\definecolor{colorB}{rgb}{0,0.3,1}
\definecolor{colorC}{rgb}{0.9,0.8,0.2}
\definecolor{colorD}{rgb}{0,0.65,0}
\definecolor{lesslightgray}{rgb}{0.5,0.5,0.5}
\definecolor{light-gray}{gray}{0.95}

\let\tilde\widetilde
\let\hat\widehat

\newcommand{\calA}{\mathcal{A}}

\newcommand{\calP}{\mathcal{P}}

\newcommand{\argmin}{\mathop{\mathrm{argmin}}}  
\newcommand{\argmax}{\mathop{\mathrm{argmax}}}

\newcommand{\Softmax}{{\rm{Softmax}}}

\def\R{\mathbb{R}}

\def\I{\mathcal{I}}

\let\cite\citep 

\usepackage{amsthm}
\usepackage[many]{tcolorbox}
\usepackage{etoolbox}
\BeforeBeginEnvironment{proof}{\par\vspace{-1\baselineskip}}
\AfterEndEnvironment  {proof}{\par\vspace{-1.5\baselineskip}}

\newtheoremstyle{theoremstyle}
  {.5\baselineskip} %
  {.5\baselineskip} %
  {}                  %
  {}                  %
  {\bfseries}        %
  {.}                 %
  {1em}               %
  {}                  %

\theoremstyle{theoremstyle}
\newtheorem{theorem}{Theorem}[section]
\newtheorem{lemma}{Lemma}[section]
\newtheorem{corollary}{Corollary}[theorem]

\newtheorem{definition}{Definition}[section]

\newtheorem{remark}{Remark}[section]

\tcolorboxenvironment{theorem}{
  breakable,
  colback=black!10,
  colframe=white,%
  width=\dimexpr\linewidth+10pt\relax,%
  enlarge left by=-5pt,%
  enlarge right by=-5pt,%
  boxsep=5pt,%
  boxrule=0pt,
  left=0pt,right=0pt,top=0pt,bottom=0pt,
  sharp corners,
  before skip=0.5\baselineskip, %
  after skip=0.5\baselineskip,  %
  fonttitle=\bfseries, %
  coltitle=black %
}
\tcolorboxenvironment{remark}{
  blanker,
  breakable,
  before skip=.8\baselineskip,  %
  after  skip=.8\baselineskip   %
}

\tcolorboxenvironment{proposition}{
  breakable,
  colback=black!10,
  colframe=white,%
  width=\dimexpr\linewidth+10pt\relax,%
  enlarge left by=-5pt,%
  enlarge right by=-5pt,%
  boxsep=5pt,%
  boxrule=0pt,
  left=0pt,right=0pt,top=0pt,bottom=0pt,
  sharp corners,
  before skip=0.5\baselineskip, %
  after skip=0.5\baselineskip,  %
  fonttitle=\bfseries, %
  coltitle=black %
}

\tcolorboxenvironment{lemma}{
  breakable,
  colback=black!10,
  colframe=white,%
  width=\dimexpr\linewidth+10pt\relax,%
  enlarge left by=-5pt,%
  enlarge right by=-5pt,%
  boxsep=5pt,%
  boxrule=0pt,
  left=0pt,right=0pt,top=0pt,bottom=0pt,
  sharp corners,
  before skip=0.5\baselineskip, %
  after skip=0.5\baselineskip,  %
  fonttitle=\bfseries, %
  coltitle=black %
}

\tcolorboxenvironment{corollary}{
  breakable,
  colback=black!10,
  colframe=white,%
  width=\dimexpr\linewidth+10pt\relax,%
  enlarge left by=-5pt,%
  enlarge right by=-5pt,%
  boxsep=5pt,%
  boxrule=0pt,
  left=0pt,right=0pt,top=0pt,bottom=0pt,
  sharp corners,
  before skip=0.5\baselineskip, %
  after skip=0.5\baselineskip,  %
  fonttitle=\bfseries, %
  coltitle=black %
}

\tcolorboxenvironment{definition}{
  breakable,
  colback=black!10,
  colframe=white,%
  width=\dimexpr\linewidth+10pt\relax,%
  enlarge left by=-5pt,%
  enlarge right by=-5pt,%
  boxsep=5pt,%
  boxrule=0pt,
  left=0pt,right=0pt,top=0pt,bottom=0pt,
  sharp corners,
  before skip=0.5\baselineskip, %
  after skip=0.5\baselineskip,  %
  fonttitle=\bfseries, %
  coltitle=black %
}
\tcolorboxenvironment{assumption}{
  breakable,
  colback=black!10,
  colframe=white,%
  width=\dimexpr\linewidth+10pt\relax,%
  enlarge left by=-5pt,%
  enlarge right by=-5pt,%
  boxsep=5pt,%
  boxrule=0pt,
  left=0pt,right=0pt,top=0pt,bottom=0pt,
  sharp corners,
  before skip=0.5\baselineskip, %
  after skip=0.5\baselineskip,  %
  fonttitle=\bfseries, %
  coltitle=black %
}
\tcolorboxenvironment{lemma}{
  breakable,
  colback=black!10,
  colframe=white,%
  width=\dimexpr\linewidth+10pt\relax,%
  enlarge left by=-5pt,%
  enlarge right by=-5pt,%
  boxsep=5pt,%
  boxrule=0pt,
  left=0pt,right=0pt,top=0pt,bottom=0pt,
  sharp corners,
  before skip=0.5\baselineskip, %
  after skip=0.5\baselineskip,  %
  fonttitle=\bfseries, %
  coltitle=black %
}

\crefname{theorem}{Theorem}{Theorems}
\crefname{proposition}{Proposition}{Propositions}
\crefname{lemma}{Lemma}{Lemmas}
\crefname{corollary}{Corollary}{Corollaries}
\crefname{definition}{Definition}{Definitions}
\crefname{assumption}{Assumption}{Assumptions}
\crefname{remark}{Remark}{Remarks}
\crefname{problem}{Problem}{Problems}
\crefname{property}{Property}{property}

\tcolorboxenvironment{hypothesis}{
  breakable,
  colback=black!10,
  colframe=white,%
  width=\dimexpr\linewidth+10pt\relax,%
  enlarge left by=-5pt,%
  enlarge right by=-5pt,%
  boxsep=5pt,%
  boxrule=0pt,
  left=0pt,right=0pt,top=0pt,bottom=0pt,
  sharp corners,
  before skip=0.5\baselineskip, %
  after skip=0.5\baselineskip,  %
  fonttitle=\bfseries, %
  coltitle=black %
}
\crefname{hypothesis}{Hypothesis}{Hypothesises}

\tcolorboxenvironment{fact}{
  breakable,
  colback=black!10,
  colframe=white,%
  width=\dimexpr\linewidth+10pt\relax,%
  enlarge left by=-5pt,%
  enlarge right by=-5pt,%
  boxsep=5pt,%
  boxrule=0pt,
  left=0pt,right=0pt,top=0pt,bottom=0pt,
  sharp corners,
  before skip=0.5\baselineskip, %
  after skip=0.5\baselineskip,  %
  fonttitle=\bfseries, %
  coltitle=black %
}
\crefname{fact}{Fact}{Facts}

\tcolorboxenvironment{example}{
  breakable,
  colback=black!10,
  colframe=white,%
  width=\dimexpr\linewidth+10pt\relax,%
  enlarge left by=-5pt,%
  enlarge right by=-5pt,%
  boxsep=5pt,%
  boxrule=0pt,
  left=0pt,right=0pt,top=0pt,bottom=0pt,
  sharp corners,
  before skip=0.5\baselineskip, %
  after skip=0.5\baselineskip,  %
  fonttitle=\bfseries, %
  coltitle=black %
}
\crefname{example}{Example}{Examples}

\tcolorboxenvironment{question}{
  breakable,
  colback=black!10,
  colframe=white,%
  width=\dimexpr\linewidth+10pt\relax,%
  enlarge left by=-5pt,%
  enlarge right by=-5pt,%
  boxsep=5pt,%
  boxrule=0pt,
  left=0pt,right=0pt,top=0pt,bottom=0pt,
  sharp corners,
  before skip=0.5\baselineskip, %
  after skip=0.5\baselineskip,  %
  fonttitle=\bfseries, %
  coltitle=black %
}

\crefname{question}{Question}{Questions}
\numberwithin{equation}{section}
\numberwithin{theorem}{section}
\numberwithin{proposition}{section}
\numberwithin{definition}{section}
\numberwithin{lemma}{section}
\numberwithin{assumption}{section}
\numberwithin{remark}{section}

\usepackage{lipsum}

\newcommand*{\annot}[1]{\tag*{\footnotesize{\textcolor{black!50}{\big(#1\big)}}}}

\makeatletter
\let\save@mathaccent\mathaccent
\newcommand*\if@single[3]{%
    \setbox0\hbox{${\mathaccent"0362{#1}}^H$}%
    \setbox2\hbox{${\mathaccent"0362{\kern0pt#1}}^H$}%
    \ifdim\ht0=\ht2 #3\else #2\fi
}
\newcommand*\rel@kern[1]{\kern#1\dimexpr\macc@kerna}
\newcommand*\widebar[1]{\@ifnextchar^{{\wide@bar{#1}{0}}}{\wide@bar{#1}{1}}}
\newcommand*\wide@bar[2]{\if@single{#1}{\wide@bar@{#1}{#2}{1}}{\wide@bar@{#1}{#2}{2}}}
\newcommand*\wide@bar@[3]{%
    \begingroup
    \def\mathaccent##1##2{%
        \let\mathaccent\save@mathaccent
        \if#32 \let\macc@nucleus\first@char \fi
        \setbox\z@\hbox{$\macc@style{\macc@nucleus}_{}$}%
        \setbox\tw@\hbox{$\macc@style{\macc@nucleus}{}_{}$}%
        \dimen@\wd\tw@
        \advance\dimen@-\wd\z@
        \divide\dimen@ 3
        \@tempdima\wd\tw@
        \advance\@tempdima-\scriptspace
        \divide\@tempdima 10
        \advance\dimen@-\@tempdima
        \ifdim\dimen@>\z@ \dimen@0pt\fi
        \rel@kern{0.6}\kern-\dimen@
        \if#31
        \overline{\rel@kern{-0.6}\kern\dimen@\macc@nucleus\rel@kern{0.4}\kern\dimen@}%
        \advance\dimen@0.4\dimexpr\macc@kerna
        \let\final@kern#2%
        \ifdim\dimen@<\z@ \let\final@kern1\fi
        \if\final@kern1 \kern-\dimen@\fi
        \else
        \overline{\rel@kern{-0.6}\kern\dimen@#1}%
        \fi
    }%
    \macc@depth\@ne
    \let\math@bgroup\@empty \let\math@egroup\macc@set@skewchar
    \mathsurround\z@ \frozen@everymath{\mathgroup\macc@group\relax}%
    \macc@set@skewchar\relax
    \let\mathaccentV\macc@nested@a
    \if#31
    \macc@nested@a\relax111{#1}%
    \else
    \def\gobble@till@marker##1\endmarker{}%
    \futurelet\first@char\gobble@till@marker#1\endmarker
    \ifcat\noexpand\first@char A\else
    \def\first@char{}%
    \fi
    \macc@nested@a\relax111{\first@char}%
    \fi
    \endgroup
    }
\makeatother

\newcommand{\Attn}{{\rm Attn}}

\newcommand{\onehot}[2]{e^{(#1)}_{#2}}
\newcommand{\li}{{\rm Linear}}
\newcommand{\hop}{{\rm Hopfield}}

\newcommand{\hw}{\hat{w}}
\newcommand{\gd}{{\rm GD}}

\definecolor{blue}{named}{black}

\newcommand*{\email}[1]{\footnote{\href{mailto:#1}{\texttt{#1}}}}

\setlist[itemize,enumerate]{
  topsep    = \dimexpr 6pt-1em\relax  plus 1pt minus 1pt,
  itemsep   = .3em plus 2pt,
  parsep    = 0pt plus 1pt,
  partopsep = 0pt
}

\begin{document}

\begin{titlepage}

\begin{flushright}
Last Update: September 21, 2025
\end{flushright}

\begin{center}

{
\LARGE \bfseries %
\begin{spacing}{1.15} %
In‑Context Algorithm Emulation in Fixed‑Weight Transformers
\end{spacing}
}

\vskip 1em
Jerry Yao-Chieh Hu$^{\dagger\ddag*}$
\footnote{\href{mailto:jhu@u.northwestern.edu}{\texttt{jhu@u.northwestern.edu}}; \href{mailto:jhu@ensemblecore.ai}{\texttt{jhu@ensemblecore.ai}}}
\quad
Hude Liu$^{*}$\email{hudeliu0208@gmail.com}
\quad
Jennifer Yuntong Zhang$^{\natural*}$\email{jenniferyt.zhang@mail.utoronto.ca}
\quad
Han Liu$^{\dagger\S}$\email{hanliu@northwestern.edu}

\def\thefootnote{*}
\footnotetext{These authors contributed equally to this work. Part of the work done during JH’s internship at Ensemble AI.
Code is available at \url{https://github.com/MAGICS-LAB/algo_emu}.}

\vskip 1em

{\small
\begin{tabular}{ll}
 $^\dagger\;$Center for Foundation Models and Generative AI, Northwestern University, Evanston, IL 60208, USA\\
 \hphantom{$^\ddag\;$}Department of Computer Science, Northwestern University, Evanston, IL 60208, USA\\
 $^\ddag\;$Ensemble AI, San Francisco, CA 94133, USA\\
 $^\natural\;$Engineering Science, University of Toronto, Toronto, ON M5S 1A4, CA\\
 $^\S\;$Department of Statistics and Data Science, Northwestern University, Evanston, IL 60208, USA
\end{tabular}}

\end{center}
\noindent
We prove that a minimal Transformer with frozen weights emulates a broad class of algorithms by in-context prompting. 
We formalize two modes of in-context algorithm emulation.
In the \textit{task-specific mode}, for any continuous function $f: \mathbb{R} \to \mathbb{R}$, we show the existence of a single-head softmax attention layer whose forward pass reproduces reproduces functions of the form $f(w^\top x - y)$ to arbitrary precision.
This general template subsumes many popular machine learning algorithms (e.g., gradient descent, linear regression, ridge regression).
In the \textit{prompt-programmable mode}, we prove universality: 
a single fixed-weight two-layer softmax attention module emulates all algorithms from the task-specific class (i.e., each implementable by a single softmax attention) via only prompting.
Our key idea is to construct prompts that encode an algorithm’s parameters into token representations, creating sharp dot-product gaps that force the softmax attention to follow the intended computation.
This construction requires no feed-forward layers and no parameter updates. 
All adaptation happens through the
prompt alone. 
Numerical results corroborate our theory.
These findings forge a direct link between in-context learning and algorithmic emulation, and offer a simple mechanism for large Transformers to serve as prompt-programmable libraries of algorithms. 
They illuminate how GPT-style foundation models may swap algorithms via prompts alone, and establish a form of algorithmic universality in modern Transformer models.

\vfill
\textbf{Keywords:} Foundation Model, Transformer Expressiveness, In‑Context Learning, In-Context Universality, Algorithm Emulation, Algorithm Swapping, Universal Approximation
    
\end{titlepage}

{
\setlength{\parskip}{0em}
\setcounter{tocdepth}{2}
\tableofcontents
}
\setcounter{footnote}{0}

\clearpage

\section{Introduction}
\label{sec:intro}
We show that a minimal Transformer architecture with frozen weights is capable of emulating a broad class of algorithms through prompt design alone.
This stylized problem setting isolates the core of in-context computation and provides an analytic lens on fundamental questions in Transformer models:
How do fixed-weight models execute diverse tasks from context alone?
How does a prompt turn into an algorithmic procedure?
How do prompt-encoded parameters and query-key routing realize task identification and stepwise execution?
What minimal architectural ingredients suffice for general in-context capability?
As foundation models rise to prominence in modern AI \cite{bommasani2021opportunities}, these questions are central, since much of their practical utility comes from in‑context learning (prompting) rather than explicit retraining \cite{brown2020language,liu2023pre}.
Against this backdrop,
this work offers a rigorous basis for in‑context \textit{task} learning\footnote{We use ``task'' to highlight algorithm‑level adaptation (to diverse tasks), not mere pattern completion.}, supplies a simple mechanism for Transformers to act as prompt‑programmable algorithm libraries, and shows how GPT‑style models may swap algorithms via prompts alone,
 shedding light on their general-purpose capabilities.

Large Transformer models exhibit ability to adapt to a new task by conditioning on examples or instructions provided in the prompt without any gradient updates.
This capability is known as In-Context Learning (ICL) \cite{min2022rethinking,brown2020language}.
Prior work on Transformer in‑context learning falls into two strands.
One trains models that learn in context for a specific function class \cite{garg2022can,akyurek2022learning,li2023transformers,ahn2023transformers,zhang2024trained}.
The other hand‑engineers Transformers to enact particular algorithms with fixed weights \cite{bai2023transformers,von2023transformers,wu2024incontex}.
In particular, \citet{bai2023transformers} demonstrate that \textit{task-specific} attention layers --- attention mechanisms with weights designed for a given task --- implement a variety of algorithms without gradient updates.
For example, a single Transformer with fixed, task-tailored attention weights achieves near-optimal performance on algorithms such as least-squares regression, ridge regression, lasso, and gradient descent \cite{bai2023transformers,wu2024incontex}.
These results suggest that Transformers are capable of \textit{in-context algorithm emulation}.
Yet these approaches retrain per task or hard‑wire per algorithm.
They do not give a single fixed architecture that is prompt‑programmable across many algorithms with explicit guarantees and minimal components.

To combat this, we advance this line of research by omitting the need for designing a new Transformer block for every algorithm.
We propose a frozen Transformer architecture to emulate a library of attention-based algorithms in context without weight updates.
We achieve this by embedding algorithm-specific information into input prompts.
Specifically, we formalize two emulation modes, and establish explicit guarantees and constructive minimal designs for both.
In the \textit{task‑specific} mode (\cref{sec:examples}),
a dedicated attention module with fixed weights (single- or multi-head) executes one algorithm in context.
In the \textit{prompt‑programmable} mode (\cref{sec:theory}), by contrast,
a single Transformer module with fixed weights
re-programs itself through different prompts to execute multiple algorithms on the fly.
These constructions yield universality and minimality results for in‑context algorithm emulation.
Specifically, we demonstrate a minimalist model of internal algorithm swapping, where prompts serve as the context carrying algorithmic instructions.

\textbf{Contributions.}
We establish a new form of in-context learning universality \emph{for algorithm emulation}, limited to \emph{attention-implementable algorithms}.
Our contributions are four-fold:
\begin{itemize}
    \item \textbf{Task-Specific Emulation of $f(w^\top x- y)x$.}
    A single‑head, single‑layer softmax attention with a linear map universally approximates functions of the form $f(w^\top x-y)x$ for any continuous $f$, with frozen weights and a suitable prompt.
    This general result subsumes, for example, computing per-sample gradients and performing gradient descent updates (by choosing $f$ as a loss derivative), as well as solving linear and ridge regression in one forward pass.

    \item \textbf{Constructive, Interpretable Prompt Design for Algorithm Emulation.}
    We give an explicit prompt design strategy that encodes the target task’s parameters and induces large query-key margins so softmax follows the intended pattern, furnishing an interpretable, verifiable recipe for prompt-programming a fixed attention-only module.

    \item \textbf{A Simple Mechanism for Internal Algorithm Swapping of Transformer Models.}
    Changing only the prompt-encoded algorithm weights swaps the algorithm executed by the fixed attention-only module, without retraining. Theory (finite libraries) and experiments (e.g., Lasso, ridge, linear regression) confirm high-fidelity swapping.
    Altogether, these results shed light on the general-purpose capability of GPT-style Transformer models to select and swap internal routines via prompts (our formal proofs concern attention-only modules).

\end{itemize}

In conclusion, we show a minimalist transformer architecture serve as a general-purpose algorithm emulator in context through prompt design.
Our findings contribute to a sharp theoretical foundation for viewing in-context learning as in-context algorithm emulation.
They suggest that large pretrained softmax attention models (such as GPT-style Transformers) encode a library of algorithms, and swap among them based on prompts.
This is achieved within a unified attention architecture and without any parameter updates.
We believe this perspective opens new opportunities for understanding the emulation ability of Transformer models.

\begin{figure}
    \centering
    \includegraphics[width=\linewidth]{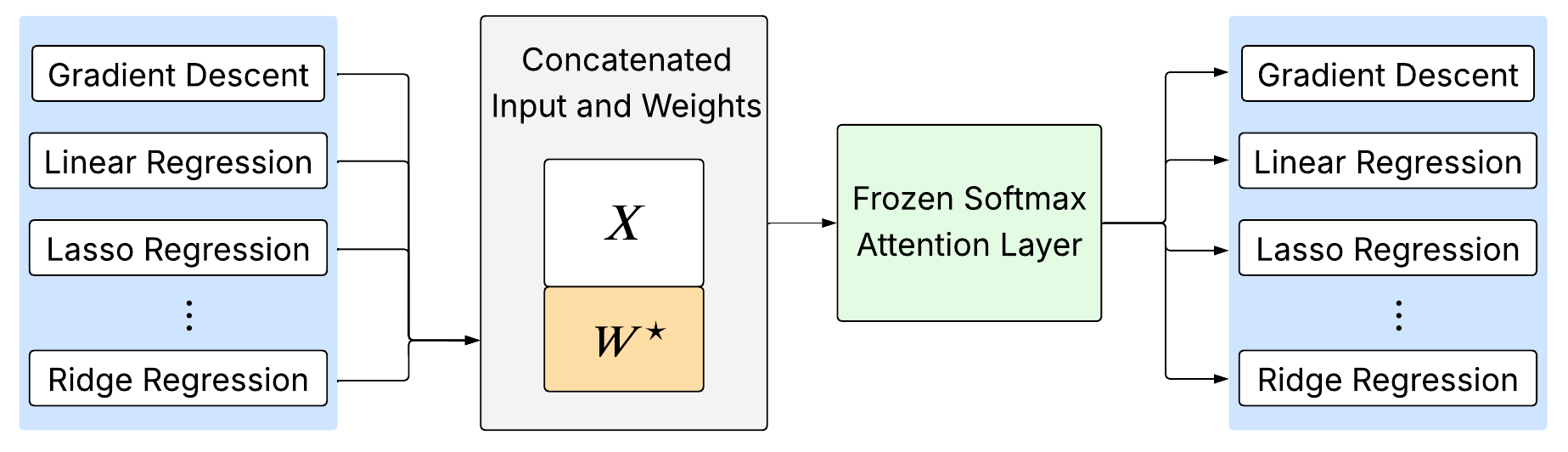}
    \caption{\small\textbf{In-Context Algorithm Emulation Overview.} $X$ represents the data input, and $W^{\star}$ represents the true weights of the each algorithm we aim to emulate.
    We show that even a softmax attention layer suffices to emulate a broad class of algorithms by changing prompt, i.e., the $W^\star$ in the prompt.
    }
    \label{fig:overview}
\end{figure}

\textbf{Organization.}
\cref{sec:preliminaries} presents ideas we build on.
\cref{sec:examples} presents illustrative examples of learning statistical models in-context with \textit{task-specific} attention heads.
\cref{sec:theory} presents our main results.
\cref{sec:proof_sketchs} presents our proof strategies.
\cref{sec:exp} presents numerical validations.

\textbf{Related Work.} Due to page limits, we defer related work discussions to \cref{sec:related_work}.

\textbf{Notations.}
We denote the index set $\{1,\ldots,I\}$ by $[I]$.
We use lowercase letters for vectors and uppercase letters for matrices.
The vector $e_j^{(n)} \in \R^n$ denotes the one-hot vector with $1$ in the $j$-th position and $0$ elsewhere.
We write $X \in \R^{d \times n}$ for the input sequence, where $d$ is the token dimension and $n$ is the sequence length.
We denote the number of attention heads by $H$.
We use $\|\cdot\|_\infty$ and $\|\cdot\|_2$ for the vector $\infty$-norm and $2$-norm, respectively.

\section{Preliminaries: Attention, In-Context Learning and Emulation}
\label{sec:method}

\label{sec:preliminaries}

\textbf{Softmax Attention.}
We define a multi-layer self-attention layer with softmax activation as follows.

\begin{definition}[Softmax Attention Layer]
\label{def:attn}
For any input sequence $X \in \R^{d\times n}$, the multi-head attention output (with $H$ heads) is
\begin{align*}
    \mathrm{Attn}_m(X) = \sum_{h=1}^{H}\underbrace{W_V^{(h)}X}_{d_o\times n}\mathrm{Softmax}(\underbrace{(W_K^{(h)}X)^\top W_Q^{(h)}X}_{n\times n})\underbrace{W_O^{(h)}}_{n\times n_o} \in \R^{d_o\times n_o},
\end{align*}
where $W_K^{(h)} \mathrm{,} W_Q^{(h)} \in \R^{d_h\times d}$, $W_V^{(h)}\in\R^{d_o\times d}$, and $W_O^{(h)}\in\R^{n \times n_o}$ for $h\in[H]$. We use $\mathrm{Attn_s}$ to denote \textit{single-head} self-attention.
\end{definition}

Following the notation of \cite{hu2025universal}, we pick non-identical dimensions for weight matrices $W_K,W_Q,W_V$ for generality of our analysis.

In the common $K:=W_K X$, $Q:=W_Q X$, $V:=W_VX$ notation, 
a single-layer softmax attention  takes a set of key vectors $K = \{k_1,\dots,k_n\}$, value vectors $V=\{v_1,\dots,v_n\}$, and a query vector $q$, to produce an output as a weighted sum of the value vectors. 
The weights on $v_i$ is  $\Softmax(k_i^\top q)$, emphasizing  values whose keys are most similar to the query. 
That is, the softmax attention uses the query as a cue to retrieve the most relevant information from the values (via their keys).

\textbf{Linear Transformation Layer $\li(\cdot)$.}
Throughout this paper, we sometimes compose attention with an additional linear mapping for flexibility.  
Such a linear transformation layer uses learned parameters to increase expressivity in attention-based constructions.  
\begin{definition}
    We use $\li:\R^{p\times n}\to\R^{q\times n}$ for any \emph{column-wise} linear map of the form
    \begin{align}
    \li(Z) = A Z, \quad A\in\R^{q\times p},
    \end{align}
    for any input sequence $Z\in\R^{p\times n}$.
    It is \emph{strictly linear} (no bias). 
    We write $\li$ when dimensions are clear (input/output shapes chosen to match attention).
\end{definition}
This layer preprocesses the input to an attention mechanism or post-processes its output.  
For example, $\Attn_s \circ \li(Z)$ applies a linear projection to $Z$ and then feeds the result into a single-head attention layer.  
Thus $\li(Z)$ enhances the expressivity of attention without weight updates.

\textbf{In-Context Learning Setup.}
In in-context learning, a fixed model (e.g., a pretrained Transformer) performs a new task without parameter updates. 
Formally, 
the model aims to approximate an unknown function $f:X\to Y$ given a few examples of $f$ in the input prompt. 
At inference, we provide $n$ exemplar pairs and a query $x_q$, and concatenate them into a single sequence
\begin{align}\label{eqn:input}
    X := \begin{bmatrix}
        x_1 & x_2 & \cdots & x_n \\
        y_1 & y_2 & \cdots & y_n
    \end{bmatrix} \in\R^{(d+1)\times n}
    \quad\text{and}\quad
    x_q \in \R^{d\times 1}.
\end{align}
Namely, the model receives $(X, x_q)$ as the input prompt. 
The goal of ICL is for the model, given input prompt $(X, x_q)$, to (i) infer $f$ from the exemplars and (ii) apply it to $x_q$ to predict $y_q=f(x_q)$.
All the learning happens in the forward pass through the sequence $X$ in an implicit fashion.

\textbf{Task-Specific Attention.}
Task-specific attention uses fixed parameters to carry out a particular task when the prompt follows the required structure (see \cite{bai2023transformers} for examples.)  
\begin{definition}\label{def:task_att}
    An attention layer is \emph{task-specific} if there exists a prompt family $\calP$ such that, for any prompt $P\in\calP$ constructed from task parameters/data, the attention’s forward pass implements the task’s mapping on the query token(s), with no parameter change.
\end{definition}
In particular, 
we embed the task’s defining transformations (e.g. a linear mapping corresponding to $f$ or part of $f$) into the attention weight matrices. 
Given a well-formed prompt of exemplar and query tokens, the attention selects and combines these tokens to compute the correct output. 
Effectively, this allows an attention layer to approximate diverse functions in context without weight updates.

\textbf{Terminology: Task-Specific vs. Prompt-Programmable In-Context Emulation.}
In-context algorithm emulation refers to executing an algorithm through a forward pass without weight updates. 
The core contribution of this work is to formalize two in-context modes and study their scope:
\begin{itemize}
    \item \textbf{Task-Specific In-Context Emulation}: for each algorithm $\mathcal{A}$, there exists an attention module (possibly multi-head) whose forward pass on a well-formed prompt implements $\mathcal{A}$ on the query token(s). 
    Each algorithm therefore requires its own dedicated parameters.
    
    \item \textbf{Prompt-Programmable In-Context Emulation (via single frozen module)}: there exists a single attention module with fixed weights $\Attn^\star$ such that, for every $\mathcal{A}$ in a target class, a suitable prompt $P_\mathcal{A}$ makes $\Attn^\star$ implement $\mathcal{A}$ on the query token(s). All adaptation occurs through the prompt rather than through weight changes.
    Namely, one $\Attn^\star$ implements a library of algorithms.
\end{itemize}
These modes are complementary: the first reflects the conventional dedicated-module view (e.g., \cite{bai2023transformers}), while the second is stronger --- one \textit{fixed-weight attention module} emulates many algorithms via prompts (our contribution).
In the remainder of the paper, 
\cref{sec:examples} develops the task-specific case.
\cref{sec:theory} establishes the prompt-programmable case by showing how the latter \emph{subsumes} the former via in-context simulation of task-specific modules.

\section{Task-Specific In-Context Algorithm Emulation}
\label{sec:examples}

We present multiple examples demonstrating how softmax attention modules mimic behaviors of  various learning algorithms including gradient descent and linear regression.
We begin with a very general result showing that even a single-layer, single-head attention mechanism is a universal approximator for a broad class of functions defined on the prompt.

\paragraph{In-Context Universal Approximation of $f(w^\top x - y)x$.}
Let $x\in\R^d$, $y\in\R$, $w\in\R^d$, and let $f:\R\to\R$ be continuous.
We consider functions of the form $f(w^\top x-y)x$, where $f$ acts on the residual $w^\top x-y$.
This template is very general: many learning rules for linear models take this form, including many residual/gradient-style updates\footnote{For example, $f(t)=t$ corresponds to the raw residual $ (w^\top x - y)x$, $f(\cdot) = \nabla_w\ell(\cdot)x$ corresponds to per-sample gradients $\nabla_w\ell(w^\top x-y)x$ linear regression or classification with loss $\ell(\cdot)$, and
nonlinear $f$ (sigmoid, step, etc.)  corresponds to perceptron updates or other error-correcting rules.}.
Hence $f(w^\top x-y)x$ subsumes a wide family of residual-driven updates central to machine learning.
Thus, their in-context realization explains much of in-context learning.
To this end, showing that attention is capable of emulating any continuous $f(w^\top x - y)x$ indicates a powerful and general capability.
It means the attention module implements any continuous adjustment or mapping based on the prediction $w^\top x$ and the label $y$.
The next theorem shows how a single-head attention approximates $[f(w^\top x_i-y_i)x_i]_{i=1}^n$ arbitrarily well.

\begin{theorem}[In-Context Emulation of $f(w^\top x - y)x$ with Single-Head Attention]
\label{thm:attn_sim_f}
Let
\begin{align*}
    X := \begin{bmatrix}
        x_1 & x_2 & \cdots & x_n \\
        y_1 & y_2 & \cdots & y_n
    \end{bmatrix} \in\R^{(d+1)\times n}
    \quad\text{and}\quad
    W :=
    \begin{bmatrix}
        w & w & \cdots & w
    \end{bmatrix}\in\R^{d\times n},
\end{align*}
where $x_i \in \R^d$, $y_i \in \R$, and $w \in \R^d$ is the coefficient vector.
Define the input as:
\begin{align}\label{eqn:prompt}
    Z :=
    \begin{bmatrix}
        x_1 & x_2 & \cdots & x_n \\
        y_1 & y_2 & \cdots & y_n \\
        w & w & \cdots & w
    \end{bmatrix}
    =\begin{bmatrix}
        X\\W
    \end{bmatrix} \in \R^{(2d+1)\times n}.
\end{align}
Assume $\max\{ \|X\|_\infty, \|W\|_\infty \} \le B$.
For any continuous function $f:\R \to \R$ and any $\epsilon > 0$, there exists a single-head attention $\Attn_s$ with a linear layer $\li$ such that
\begin{align*}
    \|
    \Attn_s \circ \li(Z)
    -
    \begin{bmatrix}
        f(w^\top x_1-y_1)x_1 &
        \cdots
        &
        f(w^\top x_n-y_n)x_n
    \end{bmatrix}
    \|_\infty
    \leq
    \epsilon,
    \quad\text{for any}\quad
    \epsilon > 0.
\end{align*}
\end{theorem}

\begin{proof}
    See \cref{proof:thm:attn_sim_f} for a detailed proof.
\end{proof}

\cref{thm:attn_sim_f} establishes that even the simplest softmax attention alone suffices to encode any continuous function of the form $f(w^\top x-y)x$ by incorporating weights in the prompt.
A direct implication is by replacing $f$ with the derivatives of differentiable loss function as follows.

\paragraph{Example 1: In-Context Emulation of Single-Step GD.}
Building on
\cref{thm:attn_sim_f}, we show that a softmax attention layer emulates \underline{G}radient \underline{D}escent (GD) in-context.
Fristly, we replace the continuous function $f(\cdot)$ in \cref{thm:attn_sim_f} with $\nabla \ell(\cdot)x$, where $\ell:\R\to\R$ is any differentiable loss function.
We show that the softmax attention emulates \textit{per‐sample gradients} in context.

\begin{corollary}[In-Context Emulation of Per-Sample Gradients]
\label{cor:in-context-GD-attn}
Let $\ell:\R\to\R$ be differentiable and $\ell':\R\to\R$ for its scalar derivative,
$\ell'(t)=\frac{d}{dt}\ell(t)$.
For $z := w^\top x-y$ with $x\in\R^d$, $y\in\R$, $w\in\R^d$, denote $\nabla_w \ell(z) := \ell'(z)x$.
Set $f(\cdot)=\ell'(\cdot)$ in \cref{thm:attn_sim_f}.
With $Z=[X;W]$ as in \eqref{eqn:prompt}, for any $\epsilon>0$, there exist a single-head attention
$\Attn_s(\cdot)$ and a linear map $\li(\cdot)$ such that,
\begin{align*}
\|
\underbrace{\Attn_s\circ\li(Z)}_{=:\widehat{G}\;\;\text{\textit{approximated} per-sample gradient matrix}}
-
\underbrace{[
\ell'(w^\top x_1-y_1)x_1,\cdots,\ell'(w^\top x_n-y_n)x_n
]}_{=: G\;\;\text{\emph{target} per-sample gradient matrix}}
\|_\infty
\le \epsilon.
\end{align*}
\end{corollary}

\cref{cor:in-context-GD-attn} shows that a single-layer single-head softmax attention with a linear map approximates the individual (per-sample) gradient terms $\{\ell'(w^\top x_i-y_i)x_i\}_{i=1}^n$.
Moreover, the layer outputs all per-sample gradient terms in parallel.
Next, we extend \cref{cor:in-context-GD-attn} to show that a fixed attention layer implements the full
gradient-descent update across all samples in-context.

Aggregating the per-sample gradients gives one GD step
\begin{align*}
\hat L_n(w):=\frac{1}{n}\sum_{i=1}^n \ell(w^\top x_i-y_i),
\quad
\nabla \hat L_n(w)=\frac{1}{n}\sum_{i=1}^n \ell'(w^\top x_i-y_i)x_i =: g .
\end{align*}
From \cref{cor:in-context-GD-attn}, let $\hat{G}$ be the attention output and choose the readout  $u:=\frac{1}{n}\mathbf{1}_n\in\R^n$ (equivalently, right-multiply by $W_O=u$ in \cref{def:attn}).
Define the attention estimate of the average gradient as $\hat{g}:=\hat{G} u$.
Then $\hat{g} \approx g$, and the target update is $w_{\rm GD}^+:=w-\eta \nabla \hat{L}_n(w)$.
Feeding $w$ in the prompt and applying the same readout produces a single $d$-dimensional update vector from the layer.
The next corollary states the precise approximation guarantee.

\begin{corollary}[In-Context Emulation of a Single GD Step]
    \label{cor:in-context-sim-gd}
    Let $\ell:\R\to\R$ be differentiable and define $\hat {L}_n(w) := \frac{1}{n}\sum_{i=1}^n \ell(w^\top x_i-y_i)$.
    For any step size $\eta>0$ and any $\epsilon>0$, there exist a single-head attention $\Attn_s$ and a linear map $\li$ such that,
    with $Z=[X;W]$ as in \eqref{eqn:prompt},
    choosing the readout $u:=\tfrac1n\mathbf{1}_n$ \text{(equivalently, right-multiply by $W_O=u$ in \cref{def:attn})}, we have
    \begin{align*}
    \hat{w}_{\rm GD} := (\Attn_s\circ\li(Z))u \in\R^d
    \quad\text{and}\quad
    \|\hat w_{\rm GD} - \underbrace{(w-\eta\nabla \hat L_n(w))}_{w^+_{\rm GD}} \|_\infty \le \epsilon .
    \end{align*}
    \vspace{-2em}
\end{corollary}

\begin{proof}
    See \cref{proof:cor:in-context-sim-gd} for a detailed proof.
\end{proof}
\cref{cor:in-context-sim-gd} shows that a single-layer, single-head softmax attention with a linear map
aggregates the per-sample gradients via the output projection.
It produces a $d$-vector $\hat w_{\rm GD}$ that approximates the GD update $w_{\rm GD}^+ = w - \eta \nabla \hat L_n(w)$.
Notably,
each output column encodes a copy of $w$ together with a scaled per-sample gradient term.
Averaging via the readout $u=\tfrac{1}{n}\mathbf{1}_n$ then recovers $w_{\rm GD}^+$ up to $\epsilon$.

\paragraph{Example 2: In-Context Emulation of Multi-Step GD.}
We extend the single-step construction to show that a multi-layer softmax attention network emulates multi-step gradient descent.
In particular, an $(L{+}1)$-layer transformer approximates $L$ steps of gradient descent.

Stack $(L{+}1)$ copies of the single-head layer from \cref{cor:in-context-sim-gd}.
At layer $t$ ($0\le t<L$), use the readout $u^{(t)} = \tfrac1n\mathbf{1}_n$ and the prompt $Z^{(t)} = [X;W^{(t)}]$ with
$W^{(t)} := [\hat w^{(t)}_{\rm GD}\cdots\hat w^{(t)}_{\rm GD}]$.
Define
\begin{align*}
    \hat w^{(0)}_{\rm GD} := w,
    \quad\text{and}\quad
    \hat w^{(t+1)}_{\rm GD} :=  \Attn_s\circ\li(Z^{(t)})u^{(t)}.
\end{align*}
For the target iterates, set $w^{(0)}_{\rm GD}=w$ and
$w^{(t+1)}_{\rm GD}=w^{(t)}_{\rm GD}-\eta\nabla \hat L_n(w^{(t)}_{\rm GD})$.
By \cref{cor:in-context-sim-gd}, \cref{lem:comp-error-convex-gd} and $\|\cdot\|_\infty\leq\|\cdot\|_2$, we arrive
\begin{align*}
\|\hat w^{(t)}_{\rm GD}-w^{(t)}_{\rm GD}\|_\infty \le t\epsilon,
\quad t\in[L].
\end{align*}

\paragraph{Example 3: In-Context Emulation of Linear Regression.}
We now present the construction for squared loss.
We show that a single-layer softmax attention emulates linear regression in-context.

\begin{corollary}[In-Context Emulation of Linear Regression]
\label{thm:attn_sim_lr}
For any dataset $\{(x_i,y_i)\}_{i=1}^n$ with $x_i\in\R^d$, $y_i\in\R$ and any $\epsilon>0$, there exist a
single-head attention $\Attn_s$, a linear map $\li$, and a readout $u\in\R^n$ such that, with $Z=[X;W]$ as in
\eqref{eqn:prompt} (for any fixed bounded $w$),
\begin{align*}
\hat w_{\text{linear}} := (\Attn_s\circ\li(Z))u \in \R^d,
\quad\text{and}\quad
\|\hat w_{\text{linear}} - w_{\text{linear}}\|_\infty \le \epsilon ,
\end{align*}
where $w_{\text{linear}} := \argmin_{w\in\R^d}\frac{1}{2n}\sum_{i=1}^n(\langle w, x_i\rangle-y_i)^2$.
\end{corollary}

\begin{proof}
    See \cref{proof:thm:attn_sim_lr} for detailed proof.
\end{proof}

\paragraph{Example 4: In-Context Emulation of Ridge Regression.}
We add regularization term and show that a single-layer softmax attention emulates ridge regression with $L_2$ penalty.

\begin{corollary}[In-Context Emulation of Ridge Regression]
\label{thm:attn_sim_rr}
For any dataset $\{(x_i,y_i)\}_{i=1}^n$, any $\lambda\ge0$, and any $\epsilon>0$, there exist a single-head
attention $\Attn_s$, a linear map $\li$, and a readout $u\in\R^n$ such that, with $Z=[X;W]$ as in
\eqref{eqn:prompt} (and the regularization signal included in the prompt),
\begin{align*}
\hat w_{\text{ridge}} := (\Attn_s\circ\li)(Z)u \in \R^d,
\quad\text{and}\quad
\|\hat w_{\text{ridge}} - w_{\text{ridge}}\|_\infty \le \epsilon ,
\end{align*}
where $w_{\text{ridge}} := \argmin_{w\in\R^d}\frac{1}{2n}\sum_{i=1}^n(\langle w, x_i\rangle-y_i)^2+\frac{\lambda}{2}\|w\|_2^2$ with regularization term $\lambda\geq0$.
\end{corollary}

\begin{proof}
    See \cref{proof:thm:attn_sim_rr} for detailed proof.
\end{proof}

So far our constructions in \cref{sec:examples} show that, given freedom to choose parameters per algorithm, attention modules emulate gradient descent, linear regression, ridge regression, and related updates in context.
These results establish the expressive power of task-specific in-context emulation, akin to \cite{bai2023transformers}.
In \cref{sec:theory}, we build on this foundation and prove a stronger universality: a single frozen attention module $\Attn^\star$, via prompt programming, simulates all such task-specific modules.

\section{Prompt-Programmable In-Context Algorithm Emulation}
\label{sec:theory}
This section presents our main results:
softmax attention is capable of (i) emulating \textit{task-specific attention heads} in-context (\cref{sec:in_context_sim_attn}), (ii) emulating statistical models in-context (\cref{sec:att_emu_stat: another construction}), and (iii) emulating any network (with linear projections) in-context (\cref{sec:attn_sim_all_nets}).
Unlike \cref{sec:examples} requiring a separate task-specific module for each algorithm,
here we fix one frozen module $\Attn^\star$ and show that suitable prompts instruct it to emulate every algorithm in the target class.
This establishes universality: one set of weights executes a library of algorithms through prompt programming.

\subsection{In-Context Emulation of Attention}
\label{sec:in_context_sim_attn}

We first specify the input prompt with weight encoding.
\begin{definition}[Input of In-Context Emulation of Attention]
\label{def:input_attn}
Let $X \in \mathbb{R}^{d\times n}$ be the input sequence \eqref{eqn:input}, and let $W^K, W^Q, W^V \in \mathbb{R}^{d_h\times d}$ be the weight matrices of the target attention head to be emulated.
Define the vectorizations
\begin{align*}
    k:= {\rm vec}(W_K) \in \R^{dd_h},\;
    q:= {\rm vec}(W_Q) \in \R^{dd_h},\;
    v:= {\rm vec}(W_V) \in \R^{dd_h},\;
    w:=[k;q;v]\in\R^{3dd_h},
\end{align*}
where $w$ is the concatenation of $k,q,v$.
Finally, define the extended input $X_p$ for in-context emulation of the attention head specified by $W_K,W_Q,W_V$ as
\begin{align*}
        X_p := \begin{bmatrix}
            X\\W_{\rm in}\\I_n
        \end{bmatrix}
        \quad\text{with}\quad
        W_{\rm in} :=
        \begin{bmatrix}
            0\cdot w & 1\cdot w & 2\cdot w & \cdots & (n-1)\cdot w\\
            w & w & w & \cdots & w
        \end{bmatrix}\in\R^{6dd_h}.
\end{align*}
\end{definition}
In other words, $W_{\mathrm{in}}$ is a $2\times n$ block matrix whose $j$-th column consists of $j\cdot w \in \R^{dd_h}$ (in the first block-row) and $w \in \R^{dd_h}$ (in the second block-row), for $j=0,1,\dots,n-1$.
Appending $W_{\mathrm{in}}$ as additional rows to $X$ produces the prompt $X_p$ that encodes the target weights.

Using this weight-encoding prompt, we now design a two-layer attention mechanism to reproduces the effect of the target attention head in-context.

\begin{theorem}[In-Context Emulation of Attention]
\label{thm:attn_sim_attn_multi}
Let $X \in \mathbb{R}^{d\times n}$ be an input sequence, and let $W^K \in \mathbb{R}^{d_h\times d}$, $W^Q \in \mathbb{R}^{d_h\times d}$, $W^V \in \mathbb{R}^{d\times d}$ be the weight matrices of the target attention head we wish to emulate in-context.
For any $\epsilon > 0$, there exists a two-layer attention network --- a multi-head attention layer $\Attn_m$ followed by a single-head attention layer $\Attn_s$ --- such that
\begin{align*}
\|
\Attn_s \circ \Attn_m (
X_p
)
-
W_VX\Softmax((W_KX)^\top W_QX)
\|_\infty
\leq \epsilon,
\quad\text{for any}\quad
\epsilon>0,
\end{align*}
where $X_p$ is the prompt defined in \cref{def:input_attn}.

\end{theorem}

\begin{remark}[Permutation Equivariance]
Our construction keeps the permutation equivariance of attention in its approximation. This means changing the order of columns in $X$ results in an identical change in the order of the columns in $\Attn_s \circ \Attn_m (
X_p
)$.
\end{remark}

\begin{proof}
    See \cref{sec:proof_sketch1} for the proof sketch and \cref{proof:thm:attn_sim_attn_multi}
    for a detailed proof.
\end{proof}

We now provide an alternative formulation of the above result.
In this variant, a single-head attention layer comes first, followed by a multi-head layer with sequence-wise linear projections.
\begin{theorem}[In-Context Emulation of Attention; Alternative Formulation]
\label{thm:attn_sim_attn}
Let $X \in \mathbb{R}^{d\times n}$ be an input sequence, and let $W^K, W^Q, W^V \in \mathbb{R}^{n\times d}$ be the weight matrices of the target attention (assumed to have bounded entries).
For any $\epsilon > 0$, there exists a single-head attention layer $\Attn_s$ followed by a multi-head attention layer with linear projections such that
    \begin{align*}
    \|
    \sum_{j=1}^{3n} \Attn_s \circ \Attn_j\circ\li_j
    \left(\begin{bmatrix}
        X \\
        W_K^\top \\
        W_Q^\top \\
        W_V^\top
    \end{bmatrix}\right)
    -
    \underbrace{W_VX}_{n\times n}\Softmax
    \underbrace{\left(
    (W_KX)^\top W_QX
    \right) }_{n\times n}
    \|_\infty \leq \epsilon.
    \end{align*}

\end{theorem}

\begin{proof}
    See \cref{sec:proof_sketch2} for the proof sketch and  \cref{subsec:Simulation of Attention: Another Construction} for a detailed proof.
\end{proof}

\begin{remark}

    \cref{thm:attn_sim_attn_multi,thm:attn_sim_attn} both establish that a fixed multi-head attention network can approximate any given attention head in-context. We present two versions of the construction using different formulations and analytical techniques.
    In particular, \cref{thm:attn_sim_attn_multi} encodes the target algorithm into the token representations (keeping the sequence length fixed), whereas \cref{thm:attn_sim_attn} achieves a similar effect by encoding the weights as additional tokens in the input sequence (keeping each token’s dimension fixed).
\end{remark}

\begin{remark}
Our constructions may contain non-standard choices, including encoding information along the embedding dimension and using $3n$ parallel attention heads.
We emphasize that the methods apply to approximate a more realistic attention with far fewer hidden dimensions and number of heads in practice.
\cref{sec:exp} provides further details.
\end{remark}

\begin{remark}[Comparison with Prior Work]
    We remark that our results differ from prior work in three key aspects.
    First, we study the practical softmax attention rather than linear or ReLU attention \cite{bai2023transformers,von2023transformers,vladymyrov2024linear}.
    Second, our results in \cref{sec:theory} go beyond task-specific ICL and establish that fixed-weight Transformers are prompt-programmable \cite{bai2023transformers,wu2024incontex,li2025transformers}.
    Third, our results are constructive, providing concrete emulation examples in contrast to prior prompting expressivity \cite{wang2023universality,furuya2024transformers} or Turing-completeness results \cite{perez2021attention,giannou2023looped,qiu2024ask}.
\end{remark}

\textbf{Extension: Modern Hopfield Networks.}
We extend our results to in-context optimization ability of dense associative memory models \cite{ramsauer2020hopfield}  in \cref{subsec:In-Context Application of Statistical Methods by Hopfield Network}.

\subsection{In-Context Emulation of Statistical Methods}
\label{sec:att_emu_stat: another construction}

\cref{thm:attn_sim_attn} shows that a frozen attention module approximates a target attention head by embedding the head’s weights into its input prompt.
We now leverage this idea to emulate a broader class of algorithms.
In essence, we replace the embedded target attention weights with the parameters of an arbitrary statistical method that we aim to emulate.
By the same principle, the fixed attention module then mimics the behavior of diverse statistical models within the in-context learning framework.

\begin{corollary}[In-Context Emulation of Statistical Methods]
\label{thm:attn_sim_stats_methods}
Let $\mathcal{A}$ be the set of all algorithms implementable by a single-layer attention network in-context.
For any finite collection of algorithms $\{a_1, a_2, \dots, a_k\}:= \calA_0 \subseteq \mathcal{A}$, there exists a two-layer attention network (a single-head layer $\Attn_s$ composed with a multi-head layer $\Attn_m$) such that for each $a\in \calA_0$ in the collection
\begin{align*}
    \|
    \sum_{j=1}^{3n}\Attn_s \circ \Attn_j\circ\li_j
    \left(
    \begin{bmatrix}
        X \\
        W^a
    \end{bmatrix}
    \right)
    -
    a(X)
    \|_\infty \leq \epsilon,
\end{align*}
where $W^a$ is the $W$ defined as \cref{def:input_attn} using $W_K^a,W_Q^a,W_V^a$.
\end{corollary}

\begin{proof}
    See \cref{proof:thm:attn_sim_stats_methods} for detailed proof.
\end{proof}

We show that a fixed attention module emulates an arbitrary finite library of in‐context algorithms by varying its prompt.
This result highlights the flexibility of softmax attention: unlike prior work that requires re‐training or fine‐tuning of the model, here we provably achieve task-specific behavior by modifying the input prompt.
In effect, a pretrained Transformer internalizes a small set of fundamental procedures and later deploys them, via prompting, across a wide range of data distributions.
Since the number of distinct algorithms is far smaller than the number of possible datasets, a model that learns a handful of algorithms can leverage them to handle many different scenarios.

\subsection{Attention Makes Every (Linear) Network In-Context}
\label{sec:attn_sim_all_nets}
We now extend the above ideas to show that softmax attention  emulates \emph{any} network (comprised of linear transformations) in-context.
Consider any layer of a neural network that applies a trainable linear map $x \to \Theta x$ with weight matrix $\Theta$.
Our results imply that if $\Theta$ is provided as part of the input sequence, a fixed attention module is capable of approximating this transformation to arbitrary precision.
Hence linear layers in standard architectures are replaceable with attention whose effective weights are encoded in the prompt rather than learned.
This substitution turns the network into an in-context learner in place of, or alongside, conventional training.

\begin{remark}[In‐Context Emulation of Linear Layers]
For example, suppose a model contains a linear layer $f(x) = \Theta x$ with weight matrix $\Theta$.
By including $\Theta$ (appropriately encoded) in the input as in our constructions above, a single softmax attention layer emulates $f(x)$ in-context to arbitrary precision.
In other words, any trainable linear mapping in the original network is replicable with a prompt-programmable attention layer whose parameters are set by the input sequence.
This enables the overall network to adjust that layer’s behavior on-the-fly via prompts, rather than having to learn $\Theta$ through pre-training.

\end{remark}

\section{Proof Sketches}
\label{sec:proof_sketchs}
We present our proof strategies here.

\subsection{Proof Sketch for \texorpdfstring{\cref{thm:attn_sim_attn_multi}}{}}
\label{sec:proof_sketch1}

\begin{figure}
    \centering
    \includegraphics[width=\linewidth]{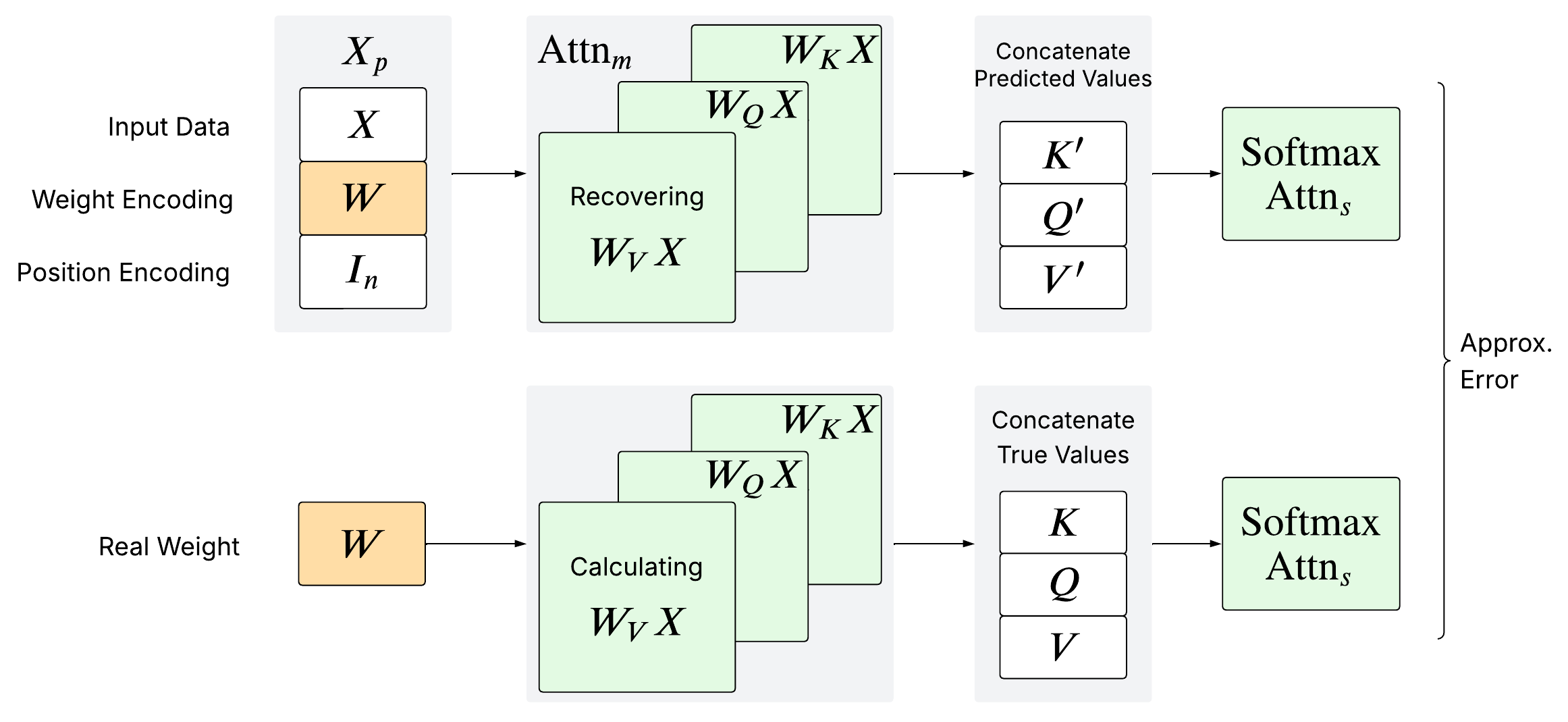}
    \vspace{-1em}
    \caption{\small\textbf{Visualization of Proof Sketch for \texorpdfstring{\cref{thm:attn_sim_attn_multi}}.}. We visualize our proof technique. We combine input data, weight encoding, and position encoding into $X_p$ as input to the multi-head attention $\text{Attn}_m$ to recover approximate key, query, and value representations. We then compare the single-head attention $\text{Attn}_s$ outputs from approximate values with ground truth values to obtain approximation error.}
    \label{fig:proof_sketch_one}
\end{figure}

We construct a two-layer Transformer (single-head layer $\Attn_s$ followed by multi-head layer $\Attn_m$) that replicates the target attention to within any error $\epsilon>0$. 
Recall from \cref{thm:attn_sim_attn_multi}:
\begin{align*}
\|
\overbrace{\Attn_s}^{\textbf{\text{Step 3}}} \circ \underbrace{\Attn_m (
\overbrace{X_p}^{\textbf{\text{Step 1}}}
)}_{\textbf{\text{Step 2}}}
-
W_VX\Softmax((W_KX)^\top W_QX)
\|_\infty 
\leq \epsilon.
\end{align*}
The high-level idea is: (\textbf{\text{Step 1}}) augment the input with a prompt encoding of the target weights $W_K, W_Q, W_V$, (\textbf{\text{Step 2}}) use groups of heads in $\Attn_m$ to recover the matrices $K = W_K X$, $Q = W_Q X$, $V = W_V X$ in-context (up to small error), and (\textbf{\text{Step 3}}) apply $\Attn_s$ with fixed weights to assemble the attention output using these reconstructed $K, Q, V$. 
We then argue in (\textbf{\text{Step 4}}) the approximation error can be made $<\epsilon$ via a stability bound on softmax attention.

\paragraph{Step 1: In-Context Weight Encoding.}
Augment the input $X \in \mathbb{R}^{d \times n}$ by appending special tokens encoding the matrices $W_K, W_Q, W_V$; denote the augmented input as $X_p$. This lets the transformer ``read'' the relevant weight parameters in its attention heads.

Explicitly, embed both the data sequence and the target head into the input:
\begin{align*}
        X_p = \begin{bmatrix}
            X\\W\\I_n
        \end{bmatrix}
        \quad\text{with}\quad
        W :=
        \begin{bmatrix}
            0\cdot w & 1\cdot w & 2\cdot w & \cdots & (n-1)\cdot w\\
            w & w & w & \cdots & w
        \end{bmatrix},
    \end{align*}
where $w=[k^\top,q^\top,v^\top]^\top$ concatenates every row of $W_K,W_Q,W_V$ following \cref{def:input_attn}.
The block $I_n$ provides token-position codes that our construction will exploit.

\paragraph{Step 2: Multi-Head Decomposition for In-Context Recovery of $K,Q,V$.}
We devote the first attention layer ($\Attn_m$) to recovering the key, query, and value matrices that the target attention would compute. 
This layer has a fixed number of heads partitioned into \textit{three groups}, corresponding to $K$, $Q$, and $V$ respectively. 
The goal is to produce, for each data token $x_i$ (the $i$-th column of $X$), an embedding that contains the vectors $k_i := W_K x_i$, $q_i := W_Q x_i$, and $v_i := W_V x_i$. We achieve this by constructing each head to output part of a linear mapping $A_h$, and then combining head outputs within each group. 
Explicitly,
in the first multi-head layer $\Attn_m$, split the heads so that:
\begin{itemize}
    \item A group of heads jointly approximates $W_K X$. By \cite[Theorem~3.2]{hu2025universal}, each head can be constructed to output linear transformations of $X$ and the appended tokens, simulating $k_i^\top x$ for columns $k_i$ of $W_K$.
    \item Another group of heads approximates $W_Q X$ in a similar manner. 
    \item A final group approximates $W_V X$.
\end{itemize}
Concatenate or combine these head outputs so that the final embedding from $\Attn_m(X_p)$ contains (up to small error) the blocks $[K;Q;V]$ for all positions in $X$.

Explicitly, for each hidden row $k_j$ of $W_K$ (similarly $q_j,v_j$) we prepend a token-wise linear map $A_h$ that pulls out this row from $w$ and repeats it $n$ times. The resulting sub-prompt has the form
\begin{align*}
        \begin{bmatrix}
            X\\ k_j \one_{1\times n}\\I_n
        \end{bmatrix},
\end{align*}
so a softmax head can learn a piecewise-constant mapping that returns $k_j^\top X$ up to any error $\epsilon_0$ by the truncated-linear interpolation lemma (\cref{thm:multi-head-truncated_in-context}).
With $H=\lceil 2(b-a)/((n-2)\epsilon_0)\rceil$ heads per group we cover all $d_h$ rows; altogether the $3N=3d_hH$ heads satisfy
\begin{align*}
    \|\underbrace{\sum_{j=1}^{N}\Attn^K_j(X_p)}_{:=K'}-K\|_\infty 
    \le  \epsilon_0,\;
    \|\underbrace{\sum_{j=1}^{N}\Attn^Q_j(X_p)}_{:=Q'}-Q\|_\infty 
    \le\ 
    \epsilon_0, \;
    \|\underbrace{\sum_{j=1}^{N}\Attn^V_j(X_p)}_{:V'}-V\|_\infty 
    \le \epsilon_0.
\end{align*}
We collect these outputs column-wise into
\begin{align*}
    \begin{bmatrix}
        K'\\Q'\\V'
    \end{bmatrix},
    \quad\text{and}\quad
    \|
    \begin{bmatrix}
        K'\\Q'\\V'
    \end{bmatrix}
    -
    \begin{bmatrix}
        K\\Q\\V
    \end{bmatrix}
    \|_\infty \le \epsilon_0.
\end{align*}

\paragraph{Step 3: Single-Head Assembly for Emulated Map.}
We consider the second layer $Attn_s$ as a single-head attention with fixed weights chosen to “read” the $K',Q',V'$ triples from $Z := \Attn_m(X_p)$ and perform the attention formula. 
Explicitly, 
apply a single-head attention layer $\Attn_s$ whose parameters are set to read off the $K$, $Q$, and $V$ sub-blocks in each token embedding:
\begin{align*}
\Attn_s(Z) :=
W_V^{(s)}Z \Softmax((W_K^{(s)} Z)^\top (W_Q^{(s)} Z)).
\end{align*}
For $Z := \Attn_m(X_p)$, we choose fixed weights 
\begin{align*}
   W_K^{(s)}= \begin{bmatrix}
       0_{n\times 2n} & I_n
   \end{bmatrix}, \quad
   W_Q^{(s)} = \begin{bmatrix}
       I_n & 0_{n\times  2n}
   \end{bmatrix}, \quad
   W_V^{(s)} = \begin{bmatrix}
       0_{n\times n} & I_n & 0_{n\times n}
   \end{bmatrix} ,
\end{align*}
so that 
\begin{align*}
    W_K^{(s)}Z \approx W_KX,\quad
    W_Q^{(s)}Z \approx W_QX, \quad W_V^{(s)}Z \approx W_VX.
\end{align*}
Hence,
\begin{align*}
\Attn_s(\Attn_m([Z]))
=\Attn_s(\begin{bmatrix}
        K'\\Q'\\V'
    \end{bmatrix})
= & ~ V'X\Softmax((K'X)^\top Q'X \\
\approx & ~ 
W_V X \Softmax((W_K X)^\top W_QX).
\end{align*}
To be precise, because $K',Q',V'$ differ from $K,Q,V$ by at most $\epsilon_0$, a first-order perturbation argument for softmax (uniform Lipschitz in sup-norm) shows
\begin{align*}
    \| \Attn_s (\begin{bmatrix}
        K'\\Q'\\V'
    \end{bmatrix}) - \Attn_s (\begin{bmatrix}
        K\\Q\\V
    \end{bmatrix})\| \le \epsilon + nB_{KQV} \epsilon_1,
\end{align*}
where $B_{KQV}$ bounds $X,W_K,W_Q,W_V$ and $\epsilon_1=O(\epsilon_0)$.

\paragraph{Step 4: Error Bound.}
Finally, we argue that the approximation can be made arbitrarily precise.
Because each head’s linear approximation can be made arbitrarily close, we can ensure
\begin{align*}
\|\Attn_s\circ \Attn_m([X;W]) - W_VX \Softmax((W_KX)^\top\,W_QX)\|_\infty
\le
\epsilon,
\end{align*}
for any $\epsilon > 0$. This completes the construction, proving in-context emulation of the target attention.
Please see \cref{proof:thm:attn_sim_attn_multi} for a detailed proof and \cref{fig:proof_sketch_one} for proof visualization.

\subsection{Proof Sketch for \texorpdfstring{\cref{thm:attn_sim_attn}}{}}
\label{sec:proof_sketch2}

We outline how to emulate the desired attention step-by-step with a fixed two-layer transformer. 
Similar to \cref{thm:attn_sim_attn_multi} (and \cref{thm:multi-head-truncated_in-context}),
our construction ensures each token’s representation in the intermediate layer carries an approximate copy of its key, query, and value vectors, which the final layer uses to perform the softmax attention. 
All necessary components (including the weight matrices $W_K,W_Q,W_V$ themselves) are encoded into the input, so the network’s weights remain untrained and generic.

\begin{figure}
    \centering
    \includegraphics[width=\linewidth]{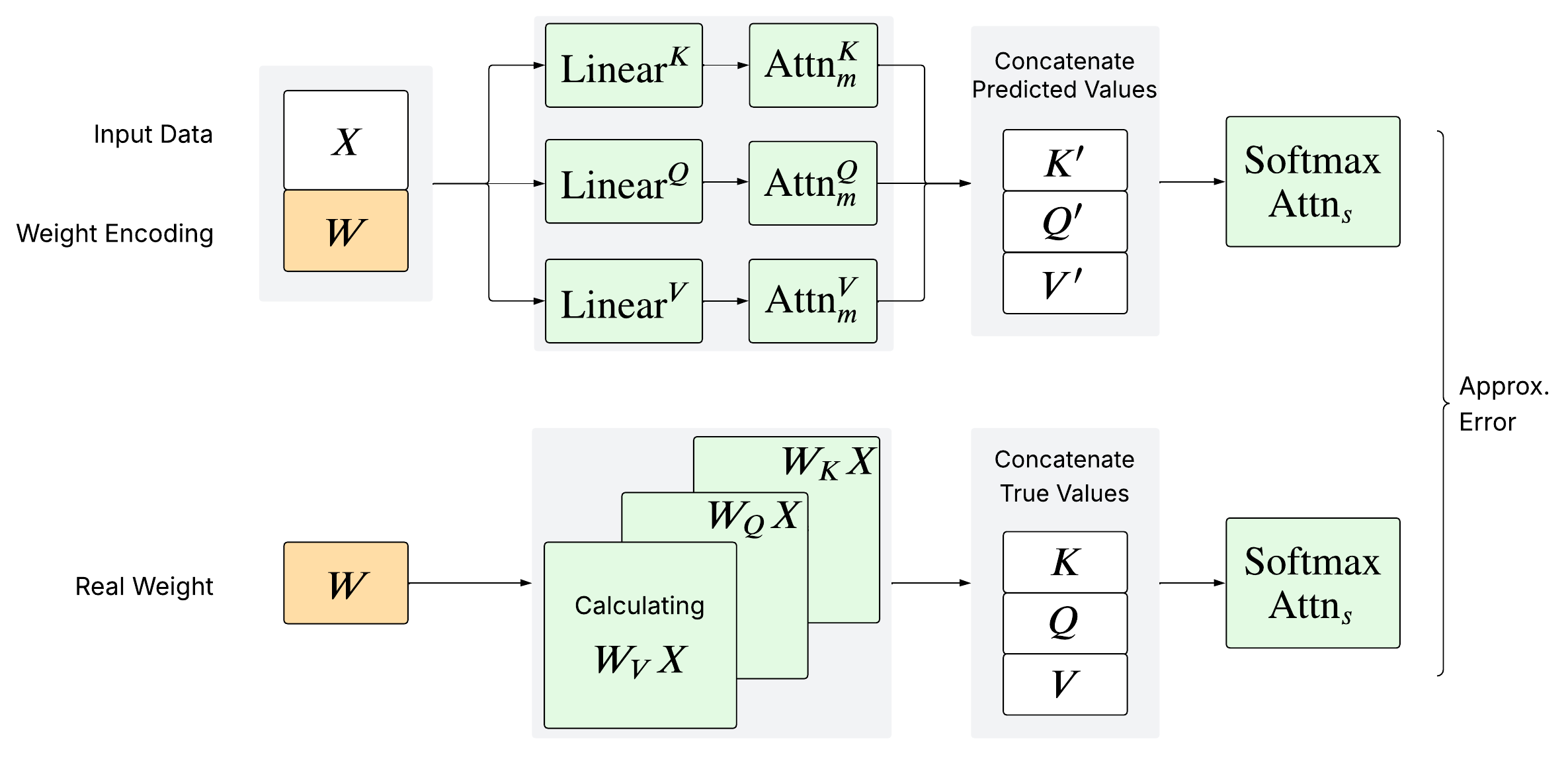}
    \vspace{-1em}
    \caption{\small\textbf{Visualization of Proof Sketch for \texorpdfstring{\cref{thm:attn_sim_attn}}.}.We visualize our proof technique. We combine input data and weight encoding as input. Each key, query, and value has a unique set of linear transformation of input ($\text{Linear}$) and multi-head attention ($\text{Attn}_m$). We feed the  input into each set to attain the approximate key, query, and value representations, respectively. We then compare the single-head attention $\text{Attn}_s$ outputs from approximate values with ground truth values to obtain approximation error.}
    \label{fig:proof_sketch_two}
\end{figure}

\textbf{Step 1: Encoding Weights into the Input.}
Let $X\in\R^{d\times n}$ be the input tokens.
Append a ``weight encoding'' matrix $W$ that contained the rows of $W_K,W_Q,W_V$ (the weight matrices of the target attention head).
This forms a extended input $[X;W]$.
The entries of $X$ and $W$ stay in a bounded range $[a,b]$.
This bound ensures all inner products remain finite.

\textbf{Step 2: Multi-Head Approximation of $K,Q,V$.}
The first layer, $\Attn_m$, has many heads.
Partition them into three groups. One group approximates $K=W_KX$, one approximates $Q=W_Q X$, and one approximates $V=W_VX$.
Then
\begin{itemize}
    \item \textbf{Simulating Dot Products on a 1D Grid.}
    \!Consider a single coordinate $k_j^\top x_c$.
    \!We create reference tokens in $W$ for grid points $t_0<\cdots<t_M$ covering $[a,b]$.
    \!We fix the head's query so it places softmax weight on the reference token whose $t_i$ is closest to $k_j^\top x_c$.
    We set the value vector at that token to encode $t_i$.
    \!Thus, the head output for token $x_c$ approximates $k^\top_j x_c$. 
    \!Fine grids reduce the error.

    \item \textbf{Reconstructing Full $K,Q,V$.}
    Repeat this idea for every coordinates in $K,Q,V$.
    Each coordinate uses one or more heads to approximate $k_j^\top x_c$, $q_j^\top x_c$ or $v_j^\top x_c$.
    Combine these approximations to get matrices $K',Q',V'$.
    The sup norm 
    \begin{align*}
        \|\begin{bmatrix}
            K'\\Q'\\V'
        \end{bmatrix}
        -
        \begin{bmatrix}
            K\\Q\\V
        \end{bmatrix}\|_\infty,
        \quad\text{can be made arbitrarily small.}
    \end{align*}

\end{itemize}

    \textbf{Step 3: Single-Head Assembly of the Attention Output.}
    The second layer, $\Attn_s$, has one head. 
    We set its weight matrices $W^{(s)}_K,W^{(s)}_Q,W^{(s)}_V$
    to pick out $K',Q',V'$ from each token's embedding.
    Then, $\Attn_s$ computes
    \begin{align*}
        V'\Softmax((K')^\top Q') 
        \approx
        W_VX \Softmax((W_K X)^\top W_Q X),
    \end{align*}
    since $K'\approx K$, $Q' \approx Q$ and $V'\approx V$. 

    \textbf{Step 4: Error Bound.}
    Softmax and matrix multiplication are continuous. 
    Small errors in $K',Q',V'$ cause a small error in the final output. 
    By refining the grid (and using enough heads), we make the sup norm error below any $\epsilon>0$.
    Please see \cref{subsec:Simulation of Attention: Another Construction} for a detailed proof and \cref{fig:proof_sketch_two} for proof visualization.

\section{Numerical Studies}
\label{sec:exp}

This section provides numerical results to back up our theory. 
We validate that the frozen softmax attention approximates (i) continuous functions (\cref{subsec:exp_sim_f}), (ii) attention heads (\cref{subsec:exp_sim_attn}), and (iii) statistical algorithms (\cref{subsec:exp_sim_statistical}) using synthetic data. 
Moreover, in \cref{subsec:real_world_exp_sim_statistical}, we show that the attention model approximates statistical models with low approximation error using real-world dataset where the model does not have access to true algorithm weights. 
It illustrates the approximation capabilities of a Transformer in practice.

We conduct all experiments using NVIDIA RTX 2080 GPU. Our code is based on standard PyTorch modules.

\subsection{Proof-of-Concept Experiment on \texorpdfstring{\cref{thm:attn_sim_f}}{}}
\label{subsec:exp_sim_f}

\textbf{Objective: \!Verifying Attention Approximates $f(w^\top x -y)x$.} We investigate accuracy of softmax attention approximating $f(w^\top x -y)x$ by training a single-head softmax attention with linear connection.

\begin{wrapfigure}{r}{0.5\textwidth}
  \vspace{-2em}
  \centering
  \includegraphics[width=0.5\textwidth]{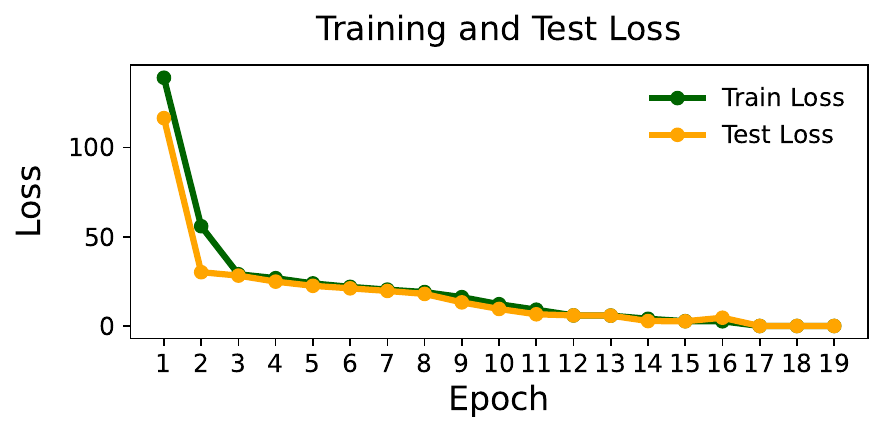}
  \vspace{-2em}
  \caption{\small\textbf{Train and Test Loss.} We report loss (MSE) of the model. We use synthetic data of $50000$ data points with sequence length being $20$ and input dimension being $24$. We set batch size to be $32$, hidden dimension to be $64$, and interpolation point to be $60$. The optimizer used is Adam with learning rate $0.001$. The result shows that only after $19$ epochs of training, both the training and testing loss reduces to almost $0$, i.e. train loss is $0.0028$	and test loss is $0.0003$.}
  \label{fig:attn_sim_f}
  \vspace{-3em}
\end{wrapfigure}

\textbf{Data Generation.} We randomly generate $X\in \R^{n\times d}$ drawn from a normal distribution,  $X \sim 10\cdot N(0,1)-5$. We also generate weight matrix $W \in \R^{n\times d}$ and $y\in\R^{n}$, both randomly drawn from a standard normal distribution, $N(0,1)$. Here, $n$ represents the sequence length and $d$ represents input dimension. The true label is $f(w^\top x -y)x$, where we choose $f(\cdot) = \tanh(\cdot)$.

\textbf{Model Architecture.} We train a single-head attention network with linear transformation to approximate $\tanh(w^\top x -y)x$. We first apply linear transformation to both $[X; y]$ and $W$. We then train the single-head attention model with the linear transformations to approximate our target function as shown in the proof of \cref{thm:attn_sim_f}.

\begin{wrapfigure}{r}{0.5\textwidth}
  \vspace{-3em}
  \centering
  \includegraphics[width=0.5\textwidth]{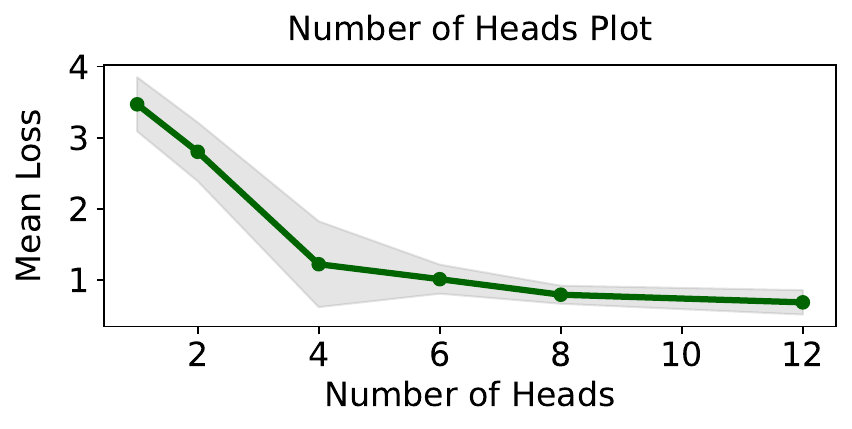}
  \vspace{-2em}
  \caption{\small\textbf{Sensitivity of Attention Emulation to the Number of Heads.} We report loss (MSE) as the mean and one standard deviation (shaded region) over $10$ random seed runs. We use synthetic data of $50000$ data points with sequence length being $20$ and input dimension being $24$. We set batch size to be $32$ and hidden dimension to be $48$. Each multi-head model and the single-head softmax attention layer is trained for $50$ epochs. The optimizer used is Adam with learning rate $0.001$. We visualize the performance (MSE $\pm$ Std) for $1,2,4,6,8,12$ heads.}
  \label{fig:attn_sim}
  \vspace{-4em}
\end{wrapfigure}

\textbf{Results.} As shown in \cref{fig:attn_sim_f}, evaluated on Mean Square Error loss, the model approximates the target $\tanh(w^\top x -y)x$ with minimal error. This experiment proves our theory.

\subsection{Proof-of-Concept Experiment on Emulating Attention Heads}
\label{subsec:exp_sim_attn}
\paragraph{Objective: \!Verifying Approximation Rates.} We investigate the affect of the number of attention heads $H$ on the accuracy of softmax attention approximating softmax attention head.

\paragraph{Data Generation.} We randomly generate a sequence of tokens $X=[x_1, x_2, \cdots, x_n] \in \R^{d\times n}$, where each entry $x_i$ is drawn independently from a normal distribution,  
\begin{align*}
    X \sim 2\cdot N(0, 1)-1.
\end{align*} 
We also generate weight matrices $K = W_K X^\top \in \R^{h\times n}$, $Q = W_Q X^\top \in\R^{h\times n}$, and $V = W_V X^\top \in \R^{d\times n}$. Each parameter matrix is randomly drawn from a standard normal distribution, $N(0, 1)$. Here, $n$ represents the sequence length, $d$ represents token dimension, and $h$ represents hidden dimension. The true label $Y\in\R^{d\times n}$ results from applying a single-layer softmax attention mechanism on inputs $X$, $K$, $Q$, and $V$.

\paragraph{Model Architecture.} We train a multi-layer attention network to approximate softmax attention function. We first train separate multi-head models with linear transformation to approximate $K$, $Q$, and $V$. Then, we use a single-head softmax attention layer to approximate softmax attention function as in the proof.

\begin{wraptable}{r}{0.5\textwidth}
\vspace{-1em}
    \centering
    \caption{\small\textbf{Sensitivity to the Number of Heads.} 
      Emulation MSE (mean $\pm$ std) for multi-head softmax attention with
      $1,2,4,6,8,$ and $12$ heads.}
    \resizebox{0.5\textwidth}{!}{%
      \begin{tabular}{lcccccc}
          \toprule
          \text{Heads} & \textbf{1} & \textbf{2} & \textbf{4} & \textbf{6} & \textbf{8} & \textbf{12} \\
          \midrule
          MSE & 3.469 & 2.802 & 1.222 & 1.012 & 0.793 & 0.686 \\
          Std & 0.381 & 0.413 & 0.603 & 0.204 & 0.127 & 0.171 \\
          \bottomrule
      \end{tabular}%
    }
    \label{tab:sim}
    \vspace{-1em}
\end{wraptable}

\paragraph{Results.} As shown in \cref{fig:attn_sim} and \cref{tab:sim}, the result validates our claim that a multi-head softmax attention mimics a target softmax attention head to arbitrary precision. Moreover, it demonstrates the convergence of multi-head softmax attention emulating softmax-based attention mapping as the number of heads increases. The approximation rate is in the trend of $O(1/H)$ where $H$ is the number of heads. The small and decreasing MSE error indicates that the simple softmax attention model approximates softmax attention head with stability.

\subsection{Proof-of-Concept Experiment on Emulating Statistical Models}
\label{subsec:exp_sim_statistical}
\paragraph{Objective: \!Emulation of Statistical Models.} We investigate the accuracy of a frozen softmax attention approximating statistical models including Lasso, Ridge and linear regression by only varying the input prompts.

\paragraph{Data Generation.} We simulate an in-context dataset by randomly generating a sequence of input tokens $X = [x_1, x_2, \dots, x_n] \in \mathbb{R}^{n \times d}$, where each $x_i$ is independently drawn from a scaled standard normal distribution, 
\begin{align*}
    x_i \sim 2\cdot N(0, 1) - 1. 
\end{align*}
A task-specific prompt vector $w \in \mathbb{R}^{p \times 1}$ is sampled from $N(0,1)$. 
In the case of Lasso, we randomly zero out entries in $w$ with probability $0.5$ to induce sparsity. We generate the output sequence $Y \in \mathbb{R}^{n \times 1}$ via a noisy linear projection: $Y = Xw + \epsilon$, where $\epsilon \sim N(0, \sigma^2)$ is Gaussian noise. For Ridge, we calculate weights using $(X^\top X+\lambda I_d)^{-1} X^\top Y$ with $\lambda=5$.

\paragraph{Model Architecture and Training.} 
We use a mixture of statistical data to train a single-layer attention network with linear transformation. Each input sample consists of $X \in \mathbb{R}^{n \times d}$ and algorithm-specific prompt $w \in \mathbb{R}^p$. We replicate $w$ across the sequence length and concatenate it with $X$ along the feature dimension to obtain an augmented input $[X; w] \in \mathbb{R}^{n \times (d+p)}$. 
We pass it through a multi-head attention layer.
We train the model for $300$ epochs using the Adam optimizer with a learning rate of $0.001$. We use $6$ attention heads, a hidden dimension of $48$, an input dimension of $24$, a batch size of $32$, and $50000$ synthetic samples. After training, we freeze the attention weights, resulting in a fixed softmax attention layer. We evaluate the frozen model on its ability to emulate various statistical algorithms using test data.

\paragraph{Baseline Architecture.} We train three separate attention models for Lasso, Ridge, and linear regression. That is, each attention model weights are adaptive to its corresponding algorithm. We use these models as baselines for comparison with the frozen attention model we propose. All baseline models use the same hyper-parameters as the frozen model.

\paragraph{Results.} As shown in \cref{tab:mse_comparison}, we compare mean MSE and standard deviation over $5$ random seed runs for the frozen attention model against baseline for Lasso, Ridge, and linear regression on the synthetic data. The frozen attention model performs as well as the baseline models trained individually on each algorithm. It achieves lower MSE on Lasso and linear regression tasks compared to their corresponding baselines. It shows that a frozen attention mechanism generalizes across these tasks given task-specific prompts. 
Moreover, the frozen model exhibits lower variance across all tasks, suggesting increased stability and robustness. These results support our claim that a frozen softmax attention layer, when conditioned on task-specific prompts, emulates statistical algorithms in context without much performance degradation.

\begin{table}[ht!]
\centering
\caption{\small\textbf{Comparison Between Baseline and Frozen Attention Layer on Synthetic Dataset.} We compare loss (MSE) as the mean and one standard deviation over $5$ random seed runs for baseline vs. frozen model on different algorithms. We train on $50000$ training data points evaluate on $10000$ testing data points for each algorithm. }
\vspace{.5em}
\begin{tabular}{llccc}
\toprule
\textbf{Model} & \textbf{Lasso} & \textbf{Ridge Regression} & \textbf{Linear Regression} \\
\midrule
    Baseline & 0.068$\pm$0.015 & 0.004$\pm$0.0003 & 0.147$\pm$0.067 \\
    Frozen Attention & 0.059$\pm$0.001 & 0.071$\pm$0.0002 & 0.120$\pm$0.003 \\
\bottomrule
\end{tabular}
\label{tab:mse_comparison}
\end{table}

\subsection{Real-World Experiment on Emulating Statistical Models}
\label{subsec:real_world_exp_sim_statistical}
\paragraph{Objective: \! Real-World Emulation of Statistical Models.} Building on \cref{subsec:exp_sim_statistical}, we use real-world data to investigate the accuracy of a frozen softmax attention emulating algorithms.

\paragraph{Data Collection and Processing.} We collect data from Ames Housing Dataset \cite{ames2025kaggle}. This dataset consists of $2930$ observations and $79$ features. We process the data by log-transforming the target variable, encoding categorical variables with one-hot vectors, replacing missing entries with median values, and standardizing numerical features. The resulting data consists of $262$ features. We fit the processed data to Lasso, Ridge, and linear regression models to obtain algorithm weights as part of the input.

\paragraph{Model Architecture and Training.} 
We use a mixture of statistical data to train a single-layer attention network with linear transformation. The input is passed through a multi-head attention layer with a linear transformation.
We train the model for $300$ epochs using the Adam optimizer with a learning rate of $0.001$. We use $8$ attention heads, a hidden dimension of $524$, and a batch size of $32$. After training, we freeze the attention weights, resulting in a fixed softmax attention layer. The frozen model is then evaluated on its ability to emulate various statistical algorithms using test data. We train the baseline models the same way as the synthetic experiment.

\begin{table}[ht]
\centering
\caption{\small\textbf{Comparison Between Baseline and Frozen Attention Layer on Ames Housing Dataset.} We compare loss (MSE) as the mean and one standard deviation over $5$ random seed runs for baseline vs. frozen model on different algorithms. We train on $80\%$ training data and evaluate on $20\%$ testing data for each algorithm. }
\vspace{.5em}
\begin{tabular}{llccc}
\toprule
\textbf{Model} & \textbf{Lasso} & \textbf{Ridge Regression} & \textbf{Linear Regression} \\
\midrule
    Baseline & 0.0354$\pm$0.0000 & 0.0132$\pm$0.0000 & 0.0288$\pm$0.0000 \\
    Frozen Attention & 	0.0322$\pm$0.0000 & 0.0252$\pm$0.0000 & 0.0250$\pm$0.0000 \\
\bottomrule
\end{tabular}
\label{tab:mse_comparison_ames}
\end{table}

\paragraph{Results.} As shown in \cref{tab:mse_comparison_ames}, we compare mean MSE and standard deviation over $5$ random seed runs for the frozen attention model against baseline for Lasso, Ridge, and linear regression on Ames Housing Data. The results shows the frozen attention model performs as well as the baseline models trained individually. We use an auxiliary network to approximate the required weight encoding. Our experiment validates that the mechanism works even when the exact weights are not supplied in real world scenarios.

\section{Discussion and Conclusion}
\label{sec:conclusion}
We study \emph{in-context algorithm emulation} in fixed-weight Transformers and formalize two modes: task-specific (\cref{sec:examples}) and prompt-programmable algorithm emulation (\cref{sec:theory}). 
For the former, we show that even a single-layer, single-head module suffices for emulating core families (of the form $f(w^\top x-y)x$) such as one-step gradient descent and linear/ridge regression, achieving architectural minimality (\cref{thm:attn_sim_f}). 
For the latter,
we show that a two-layer multi-head softmax attention module emulates a broad class of algorithms by embedding the algorithm’s weights into the input prompt (\cref{thm:attn_sim_attn_multi}). 
Altogether, a fixed softmax attention module becomes a prompt-programmable \emph{library of algorithms}: weights remain frozen, and the prompt selects the routine. 

\textbf{Mechanism.}
The mechanism is constructive. 
By encoding target weights in the input and creating large dot‑product margins, softmax attention routes along the intended computation without weight updates. 
Numerical studies support the theory: on synthetic data the model accurately approximates continuous maps of the form $f(\Braket{w,x} - y)\,x$ and emulates attention heads.
Approximation error decreases as the number of heads grows. 
On a real dataset (Ames Housing), the frozen module-driven by prompts rather than true algorithm weights-achieves low error against standard statistical models.

\textbf{Implications.}
Our results tighten the link between in‑context learning and algorithmic emulation.
Viewing prompts as callable subroutines that select and configure algorithms within a frozen model, we draw three takeaways:
(i) prompt engineering becomes interface design for algorithm selection, (ii) pretraining objectives \emph{could}, in future work, be designed to encourage learning compact libraries of reusable procedures, and (iii) analyses of internal routing help clarify how foundation models select among algorithms.
This lens explains the breadth of in-context generalization, guides prompt design, and motivates new pretraining objectives for more effective algorithm installation and utilization.

\section*{Acknowledgments}
JH would like to thank Alex Reneau, Yingyu Liang, Mimi Gallagher, Sara Sanchez, T.Y. Ball, Dino Feng and Andrew Chen for valuable conversations; Hong-Yu Chen, Maojiang Su, Weimin Wu, Zhao Song and Pin-Yu Chen for collaborations on related topics; the Red Maple Family for support; and Jiayi Wang for facilitating experimental deployments.
The authors would like to thank the anonymous reviewers and program chairs for constructive comments.

JH is partially supported by Ensemble AI and Northwestern University.
Han Liu is partially supported by NIH R01LM1372201, NSF
AST-2421845, Simons Foundation
MPS-AI-00010513, AbbVie , Dolby and Chan Zuckerberg Biohub Chicago Spoke Award.
This research was supported in part through the computational resources and staff contributions provided for the Quest high performance computing facility at Northwestern University which is jointly supported by the Office of the Provost, the Office for Research, and Northwestern University Information Technology.
The content is solely the responsibility of the authors and does not necessarily represent the official
views of the funding agencies.

Typeset with a modified LaTeX template of 1712.09542 [hep-th] by Yuji Tachikawa \cite{tachikawa2020gauging}.

\newpage
\appendix
\label{sec:append}
\part*{Appendix}
{
\setlength{\parskip}{-0em}
\startcontents[sections]
\printcontents[sections]{ }{1}{}
}

\section{Related Work}
\label{sec:related_work}
Our results diverge from prior findings on Transformer universality and in-context learning.

\subsection{Core Related Work}

\paragraph{Universal Approximation.} 
Prior studies establish that Transformers approximate arbitrary sequence-to-sequence functions, but they do not address in-context learning and often assume complex architectures. 
For example, \citet{yun2019transformers} prove that deep multi-head Transformers with feed-forward layers are universal approximators of continuous sequence-to-sequence functions. Subsequent advances tighten this finding: 
\citet{kajitsuka2023transformers,hu2024fundamental} show that even a single-layer Transformer realizes any continuous sequence function.
However, these results treat Transformers as parametric function approximators.
The model requires re-training and re-prompting to adapt to a new target function instead of handling multiple tasks through context. 
In contrast, we prove that a minimal Transformer architecture, even a single-layer, single-head attention module with no feed-forward network, emulates a broad class of algorithms without weight updates by varying its prompt. 
This result achieves a new level of generality through context alone (i.e. prompt-based conditioning) despite a fixed minimalist model.

\paragraph{In-Context Learning and Algorithm Emulation.}
Another line of recent theory bridges Transformers with in-context learning by designing model components to carry out specific algorithms. 
For example, \citet{bai2023transformers} show that Transformers execute a broad range of standard algorithms in-context, but each algorithm requires a distinct, tailored attention head. 
In comparison, we extend this approach by showing that one fixed attention mechanism emulates any specialized attention head via prompt encoding. 
Rather than crafting a different attention module for each algorithm, a single frozen softmax-based attention layer takes its instructions from the prompt to perform all tasks in context.
This minimal model thus becomes a unified and compact in-context algorithm emulator. It switches behaviors by changing only its input prompt, setting it apart from earlier approaches that required per-task reparameterization.

\subsection{Broader Discussion}

\paragraph{Universal Approximation and Expressivity of Transformers.}
Transformers exhibit strong expressive power as sequence models. Recent theory shows even minimal Transformer architectures approximate broad classes of functions. \citet{kajitsuka2023transformers,hu2024fundamental} prove a single-layer, single-head Transformer can memorize any finite dataset perfectly. \citet{kajitsuka2023transformers} achieve this with low-rank attention matrices, while \citet{hu2024fundamental} use attention matrices of any rank. Adding two small feed-forward layers makes it a universal approximator for continuous sequence functions under permutation-equivariance. More recently, \citet{hu2025universal} show self-attention layers alone are universal approximators. Specifically, two attention-only layers approximate continuous sequence-to-sequence mappings, and even a single softmax-attention layer suffices for universal approximation. 
Similarly, \citet{liu2025attention} also demonstrate that one single-head attention connected with linear transformations is sufficient to approximate any continuous function in $L_\infty$ norm.
These results eliminate the need for feed-forward networks, improving on earlier constructions. Overall, these findings highlight the inherent expressiveness of minimal attention mechanisms.

\paragraph{Transformers as In-Context Learners and Algorithm Emulators.}
Large Transformers also learn in-context by conditioning on examples in their prompts, without updating weights \cite{brown2020language}. Recent work formally explains this by showing attention-based models implement standard learning algorithms internally. \citet{bai2023transformers} construct Transformer heads executing algorithms such as linear regression, ridge regression, Lasso, and gradient descent steps, achieving near-optimal predictions. \citet{wu2024incontex} further build Transformers explicitly simulating multiple gradient descent iterations for training deep neural networks, with provable convergence guarantees. 
Empirical and theoretical studies confirm Transformers internalize learning algorithms when meta-trained on task families. \citet{garg2022can} show meta-trained Transformers mimic classical algorithms, such as ordinary least squares regression, in-context. 
Similarly, \citet{akyurek2022learning,von2023transformers,zhang2024trained} analyze Transformers trained on linear regression tasks and demonstrate their outputs mimic gradient descent steps precisely. Overall, existing literature shows that sufficiently trained or carefully designed Transformers emulate step-by-step computations of standard algorithms through prompt conditioning.

\paragraph{Prompt Tuning.}
Prompt‑tuning adapts frozen models by learning a short continuous prefix \cite{lester2021power,li2021prefix,liu2023ptuning}. It keeps backbone weights fixed and updates only prompt embeddings. Our setting is stricter: prompts are hand‑designed, not learned, and we give exact approximation bounds. Thus we expose the theoretical limit of prompt control: a single frozen softmax head can mimic any task‑specific head.

\paragraph{Encoding Context Along Embedding Dimension.}
Recent work in in-context learning explores encoding and manipulating context in the embedding space rather than sequence dimension. For example, \citet{liu2024incontextvectorsmakingcontext} propose In-Context Vectors for steering the model's behavior by adding task-specific vectors along the embedding space. \citet{zhuang2025vectoriclincontextlearningcontinuous} extend this idea by showing that manipulating embedding vectors such as interpolation makes in-context learning more controllable. \citet{abernethy2023mechanismsampleefficientincontextlearning} showcase that appending additional information along the embedding dimension allows the model to perform sample-efficient in-context learning.

\paragraph{Comparison to Our Work.}
The above results demonstrate the versatility of Transformer networks, but they require task‑specific weights, training, or learned prompts. 
For instance, \citet{bai2023transformers} design a different task-specific head for each algorithm of interest, raising the question of whether a single fixed attention mechanism could instead serve as a universal emulator for any algorithm given the right prompt. 
Our work directly addresses this question. 
In contrast, we prove one fixed softmax head emulates any specialized head through prompt encoding alone. 
No additional weights or training are required. 
Even the simplest attention (one layer, one head) acts as a universal algorithm emulator when given the right prompt, shifting focus from architecture to prompt design.

\clearpage
\section{Proofs of Main Text}

To prepare our proofs, we state the following axillary definitions and lemmas.

\begin{definition}[Truncated Linear Function]
\label{def:range}
    We define the truncated linear function as follows:
    \begin{align*}
    {\rm Range}_{[a,b]}(x)
    = \left\{
    \begin{aligned}
        a & \quad x \leq a,\\
        x & \quad  a \leq x \leq b, \\
        b & \quad  b \leq x.
    \end{aligned}
    \right.
    \end{align*}
\end{definition}
Intuitively, ${\rm Range}_{[a,b]}(\cdot)$ is the part of a linear function whose value is in $[a,b]$.

We then define the interpolation points in $[a,b]$ that are used in later proofs.
\begin{definition}[Interpolation]
\label{def:interpolation}
    Let $[a,b] \subset \R$ be an interval with $a \leq b$ and let $p \in \mathbb{N}^*$ be a positive integer.
    We define
    \begin{align*}
        \tilde{L}_0^{[a,b]}:=a,
        \quad
        \tilde{L}_p^{[a,b]}:=b,
        \quad
        \tilde{L}_i^{[a,b]}
        :=
        a + \frac{i}{p}\,(b-a),
        \quad i=[p-1].
    \end{align*}
    Hence, $\tilde{L}_0 < \tilde{L}_1 < \cdots < \tilde{L}_p$ forms a uniform partition of $[a,b]$.
    We also write
    \begin{align*}
        \Delta L := \tilde{L}^{[a,b]}_i - \tilde{L}^{[a,b]}_{i-1}, \quad i \in [p].
    \end{align*}
    We often omit the superscript $[a,b]$ when the context is clear.
\end{definition}

We also propose the following lemma to show Hardmax property that is capable of being approximated by Softmax.

\begin{lemma}[Lemma F.1 in \cite{hu2025universal}: Approximating Hardmax with Finite-Temperature Softmax]
\label{lem:Soft_to_Hard}
Let $x = [x_1, x_2, \dots, x_n] \in \mathbb{R}^n$, $\epsilon>0$.
Define $\Softmax_\beta(\cdot)$ as
\begin{align*}
    \Softmax_\beta(x) \coloneqq [\frac{\exp(\beta x_1)}{\sum_{j=1}^{n}\exp(\beta x_j)},\cdots,\frac{\exp(\beta x_n)}{\sum_{j=1}^{n}\exp(\beta x_j)} ].
\end{align*}
The following statements hold:
\begin{itemize}
    \item \textbf{Case of a Unique Largest Entry.}
    Assume $x_1 = \max_{i \in [n]} x_i$ is unique, and  $x_2 = \max_{i \in [n] \setminus \{1\}} x_i$.
    Then, if $\beta \ge (\ln(n-1) - \ln(\epsilon))/(x_1 - x_2)$,
    we have
    \begin{align*}
      \Bigl\| \Softmax_\beta(x) - e_1 \Bigr\|_\infty
      &\le \epsilon,
    \end{align*}
    where $e_1 \in \mathbb{R}^n$ is the one-hot vector corresponding to to the maximal entry of $x$ (i.e., $x_1$.)

    \item \textbf{Case of Two Largest Entries (Tied or Separated by $\delta$).}
    Assume $x_1$ and $x_2$ are the first and second largest entries, respectively, with $\delta = x_1 - x_2 \ge 0$.
    Let $x_3$ be the third largest entry and is smaller than $x_1$ by a constant $\gamma>0$ irrelevant to the input.
    Then, if $\beta \ge (\ln(n-2) - \ln \epsilon )/\gamma$,
    we have
    \begin{align*}
      \Bigl\|
        \Softmax_\beta(x)
        - \frac{1}{1 + e^{-\beta \delta}} e_1
           - \frac{e^{-\beta \delta}}{1 + e^{-\beta \delta}} e_2
      \Bigr\|_\infty
      &\le \epsilon.
    \end{align*}
\end{itemize}

\end{lemma}

The following technical lemma is used in the proof of \cref{thm:multi-head-truncated_in-context}.

\begin{lemma}[Refined Version of Lemma F.2 in \cite{hu2025universal}: Cases of All Heads in ${\rm Attn}^H$]
\label{lem:all_heads_cases}
For $a\in [\tilde{L}_0,\tilde{L}_{H(n-2)}]$.
For any $h\in [H]$, define three cases of the relationship between $a$ and $h$
\vspace{.5em}
\begin{itemize}
    \item \textbf{Case 1:}
        $a \in [\tilde{L}_{(h-1)(n-2)}, \tilde{L}_{h(n-2)-1}]$,
    \item \textbf{Case 2:}
    $a \notin [\tilde{L}_{(h-1)(n-2)-1}, \tilde{L}_{h(n-2)}]$.
    \item \textbf{Case 3:}
    $a \in [\tilde{L}_{(h-1)(n-2)-1}, \tilde{L}_{(h-1)(n-2)}]\cup [\tilde{L}_{h(n-2)-1}, \tilde{L}_{h(n-2)}]$.
\end{itemize}
\vspace{1em}
These cases includes all possible situation.
Then for all $h$, only two cases exists
\begin{itemize}
    \item $a$ falls in Case 1 for an $h$ and Case 2 for all others.
    \item $a$ falls in Case 3 for two adjacent $h$ and Case 2 for all others.
\end{itemize}
\end{lemma}

\begin{proof}
Because $a\in [\tilde{L}_0,\tilde{L}_{H(n-2)}]$ and
\begin{align*}
    [\tilde{L}_0,\tilde{L}_{H(n-2)}] =
    \cup_{h=1}^H
    [\tilde{L}_{(h-1)(n-2)}, \tilde{L}_{h(n-2)}],
\end{align*}

we have
\begin{align}
    a \in [\tilde{L}_{(h_a-1)(n-2)}, \tilde{L}_{h_a(n-2)}]
\end{align}
for an arbitrary $h_a$.

This leads to only two possible cases
\begin{itemize}
    \item Case 1*: $a \in [\tilde{L}_{(h_a-1)(n-2)}, \tilde{L}_{h_a(n-2)-1}]$.
    \item Case 2*: $a \in [\tilde{L}_{h_a(n-2)-1}, \tilde{L}_{h_a(n-2)}]$.
\end{itemize}

\paragraph{Case 1*: $a \in [\tilde{L}_{(h_a-1)(n-2)}, \tilde{L}_{h_a(n-2)-1}]$.}
Because $a \in [\tilde{L}_{(h_a-1)(n-2)}, \tilde{L}_{h_a(n-2)-1}]$,  for $h\neq h_a$, we have
\begin{align*}
& ~ \tilde{L}_{h(n-2)-2},\tilde{L}_{h(n-2)} < \tilde{L}_{(h_a-1)(n-2)}, \quad h < h_a \\
& ~ \tilde{L}_{h(n-2)+1},\tilde{L}_{(h-1)(n-2)-1} \geq \tilde{L}_{h_a(n-2)-1}, \quad h > h_a.
\end{align*}
Thus
\begin{align*}
& ~ [\tilde{L}_{(h_a-1)(n-2)}, \tilde{L}_{h_a(n-2)-1}] \cap [\tilde{L}_{(h-1)(n-2)-1}, \tilde{L}_{h(n-2)}] = \emptyset \\
& ~ [\tilde{L}_{(h_a-1)(n-2)}, \tilde{L}_{h_a(n-2)-1}] \cap [\tilde{L}_{(h-1)(n-2)-1}, \tilde{L}_{h(n-2)}] = \emptyset
\end{align*}
for all $h\neq h_a$.

This means that $a$ does not fall into Case 1 nor Case 3 for other $h\in [H]$.
Thus $a$ has to fall into Case 2 for other $h$.

\paragraph{Case 2*: $a \in [\tilde{L}_{(h_a-1)(n-2)}, \tilde{L}_{(h_a-1)(n-2)+1}]\cup [\tilde{L}_{h_a(n-2)-1}, \tilde{L}_{h_a(n-2)}]$.}
Without loss of generality, assume $a$ to be in the left half $[\tilde{L}_{(h_a-1)(n-2)}, \tilde{L}_{(h_a-1)(n-2)+1}]$.
Because
\begin{align*}
& ~ [\tilde{L}_{(h_a-1)(n-2)}, \tilde{L}_{(h_a-1)(n-2)+1}] = [\tilde{L}_{(h_a-1)(n-2)-1}, \tilde{L}_{(h_a-1)(n-2)}],\annot{Case 3 of $h_a-1$}\\
& ~ [\tilde{L}_{(h_a-1)(n-2)}, \tilde{L}_{(h_a-1)(n-2)+1}] = [\tilde{L}_{(h_a-1)(n-2)-1}, \tilde{L}_{(h_a-1)(n-2)}],\annot{Case 3 of $h_a$}
\end{align*}
this means $a$ falls into Case 3 for $h_a$ and $h_a-1$.

This completes the proof.
\end{proof}

We are now ready to prove the refined version of  \cite[Theorem 3.2]{hu2025universal}.

\begin{theorem}[Multi-Head Attention Approximate Truncated Linear Models In-Context]
\label{thm:multi-head-truncated_in-context}
Let $X\in\R^{d\times n}$ be the input.
Fix real numbers $a < b$, and let the truncation operator ${\rm Range}_{[a,b]}(\cdot)$ follow \cref{def:range}.
Let $w_s$ denote the linear coefficient of the in-context truncated linear model.
Define $W_s$ as
\begin{align*}
W_s :=
\begin{bmatrix}
0 \cdot w_s & 1\cdot w_s & \cdots & (n-1)\cdot w_s \\
w_s & w_s & \cdots & w_s
\end{bmatrix} \in\R^{2d\times n}.
\end{align*}
For a precision parameter $p>n$ with $\epsilon = O(1/p)$, number of head $H = p/(n-2)$
there exists a single-layer, $H$-head self-attention ${\rm Attn}^{H}$ with a linear transformation $A:\R^{d\times n}\to \R^{(d+n)\times n}$, such that $\mathrm{Attn}^{H}\circ A: \mathbb{R}^{d \times n} \to \mathbb{R}^{d_o \times n}$ satisfies, for any $i\in [n]$,
\begin{align*}
    \| {\rm Attn}^H\circ
    A(X)_{:,i} - {\rm Range}_{[a,b]}(w_s^\top x_i) e_{\tilde{k}_i}\|_\infty
    \leq
    \underbrace{\max\{\abs{a}, \abs{b}\} \cdot \epsilon_0}_{\text{finite-$\beta$ softmax error}} + \underbrace{\frac{b-a}{(n-2)H}}_{\text{interpolation error}}.
\end{align*}
Here $e_{\tilde{k}_i}$ is the one-hot vector with a $1$ at position $\tilde{k}_i$-th index and $0$ elsewhere, and
\begin{align*}
    \tilde{k}_i = G(k_i) \in [d_o],
    \quad \text{with} \quad
     k_i = \argmin_{k \in \{0,1,\cdots,p-1\}}(-2x_i^\top w_i-2t_i+\tilde{L}_0+\tilde{L}_k)\cdot k,
\end{align*}
where $G: [p]\to[d_o]$ denotes any set-to-constant function sending each selected interpolation index $k_i$ into an appropriate integer  $\tilde{k}_i\in [d_o]$ for $i\in [n]$.

\end{theorem}

\begin{proof}

Define $A:\R^{d\times n}\to \R^{(3d+n)\times n}$ for the input sequence $X$ as
\begin{align*}
    A(X) :=
    \underbrace{
    \begin{bmatrix}
        I_{3d}\\
        0_{n \times 3d}
    \end{bmatrix}
    }_{\text{token-wise linear}}
    \begin{bmatrix}
        X \\
        W_s
    \end{bmatrix}
    +
    \underbrace{
    \begin{bmatrix}
        0_{3d\times n}\\
        I_n
    \end{bmatrix}
    }_{\text{positional encoding}}
    = \begin{bmatrix}
        X\\
        W_s\\
        I_n
    \end{bmatrix} \in \R^{(3d+n) \times n}.
\end{align*}
Thus, A is a token-wise linear layer augmented with positional encoding, as it applies a linear projection to each token and then adds a unique per-token bias.

Let $p$ be a precision parameter, without loss of generality, let it be divisible by $n-2$ and denote $p/(n-2)$ as $H$.

Now we define the multi-head attention ${\rm Attn}$ of $H$ heads.
Denote $\ell_k \coloneqq k(\tilde{L}_k + \tilde{L}_0)$ for $k\in [p]$ following \cref{def:interpolation}.
We denote the $h$-th head as ${\rm Attn}_h$, and define the weight matrices as
\begin{align*}
    &W_K^{(h)} =
    -\beta
    \begin{bmatrix}
    0_{d\times d} & -2I_d & -2[(h-1)(n-2)-1]I_d & 0 & 0 & \cdots & 0\\
    0_{1\times d} & 0_{1\times d} & 0_{1\times d} & \ell_{(h-1)(n-2)-1} & \ell_{(h-1)(n-2)} & \cdots & \ell_{h(n-2))}
    \end{bmatrix},
    \\
    &W_Q^{(h)} =
    \begin{bmatrix}
        I_{d} & 0_{d\times 2d} & 0_{d\times n}\\
        0_{1\times d} & 0_{1\times 2d} & 1_{1\times n}
    \end{bmatrix},
    \\
    &W_V^{(h)} =
    \begin{bmatrix}
        0_{d_o\times (3d+1)} & \tilde{L}_{(h-1)(n-2)}e_{\tilde{k}_{(h-1)(n-2)}} & \tilde{L}_{(h-1)(n-2)+1}e_{\tilde{k}_{(h-1)(n-2)}+1} & \cdots & \tilde{L}_{h(n-2)-1}e_{\tilde{k}_{h(n-2)-1}} & 0_{d_o}
    \end{bmatrix},
\end{align*}
for every $h \in [H]$.

Here $\beta>0$ is a coefficient we use to control the precision of our approximation.
The attention reaches higher precision as $\beta$ gets larger.

With the construction of weights, we are also able to calculate the $K,~Q,~V$ matrices in ${\rm Attn}$
\begin{align}
\label{eqn:mutihead_k}
    K^{(h)}
    := & ~
    W^{(h)}_K A(X) \notag \\
    = & ~
    W^{(h)}_K \cdot
    \begin{bmatrix}
        X\\
        W_s\\
        I_n
    \end{bmatrix} \notag\\
    = & ~
    W^{(h)}_K
    \cdot
    \begin{bmatrix}
        x_1 & x_2 & \cdots & x_n\\
        0 \cdot w_s & 1\cdot w_s & \cdots & (n-1)\cdot w_s \\
        w_s & w_s & \cdots & w_s \\
        \onehot{n}{1} & \onehot{n}{2} &  \cdots & \onehot{n}{n}
    \end{bmatrix} \notag\\
    = & ~
    -\beta
    \begin{bmatrix}
    0_{d\times d} & -2I_d & -2[(h-1)(n-2)-1]I_d & 0 & 0 & \cdots & 0\\
    0_{1\times d} & 0_{1\times d} & 0_{1\times d} & \ell_{(h-1)(n-2)-1} & \ell_{(h-1)(n-2)} & \cdots & \ell_{h(n-2))}
    \end{bmatrix} \notag\\
    & ~ \cdot
    \begin{bmatrix}
        x_1 & x_2 & \cdots & x_n\\
        0 \cdot w_s & 1\cdot w_s & \cdots & (n-1)\cdot w_s \\
        w_s & w_s & \cdots & w_s \\
        \onehot{n}{1} & \onehot{n}{2} &  \cdots & \onehot{n}{n}
    \end{bmatrix}%
    \notag \\
    = & ~
    -\beta
    \begin{bmatrix}
        -2[(h-1)(n-2)-1]w_s & -2(h-1)(n-2)w_s & \cdots & -2h(n-2)w_s \\
        \ell_{(h-1)(n-2)-1} & \ell_{(h-1)(n-2)} & \cdots & \ell_{h(n-2)}
    \end{bmatrix} \in \R^{(d+1)\times n},
\end{align}
    where the last equality comes from multiplying $X$ with $0$, thus this is a extraction of non-zero entries in $W_K$.

For $Q$, we have
\begin{align}\label{eqn:mutihead_q}
    Q^{(h)}
    := & ~
    W^{h}_Q A(X) \notag\\
    = & ~
    \begin{bmatrix}
        I_{d} & 0_{d\times 2d} & 0_{d\times n}\\
        0_{1\times d} & 0_{1\times 2d} & 1_{1\times n}
    \end{bmatrix}
    \cdot
    \begin{bmatrix}
        X\\
        W_s\\
        I_n
    \end{bmatrix}%
    \notag \\
    = & ~
    \begin{bmatrix}
        I_d \cdot X + 0_{d\times 2d} \cdot W_s + 0_{d\times n} \cdot I_n \\
        0_{1\times d}\cdot X + 0_{1\times 2d}\cdot W_s + 1_{1\times n} \cdot I_n
    \end{bmatrix} \notag \\
    = & ~
    \begin{bmatrix}
        X\\
        1_{1\times n}
    \end{bmatrix} \in \R^{(d+1)\times n}.
\end{align}

For $V$, we have
\begin{align}\label{eqn:mutihead_v}
    V^{(h)}
    := & ~
    W^{(h)}_V A(X) \notag \\
    = & ~
    \begin{bmatrix}
        0_{d_o\times (3d+1)} & \tilde{L}_{(h-1)(n-2)}e_{\tilde{k}_{(h-1)(n-2)}} & \cdots & \tilde{L}_{h(n-2)-1}e_{\tilde{k}_{h(n-2)-1}} & 0_{d_o}
    \end{bmatrix}
    \cdot
    \begin{bmatrix}
        X \\
        W_s \\
        I_n
    \end{bmatrix} \notag \\
    = & ~
    \underbrace{0}_{d_o\times 3d}\cdot \begin{bmatrix}
        X \\
        W_s
    \end{bmatrix} +
    \underbrace{\begin{bmatrix}
        0_{d_o} & \tilde{L}_{(h-1)(n-2)}e_{\tilde{k}_{(h-1)(n-2)}} &  \cdots & \tilde{L}_{h(n-2)-1}e_{\tilde{k}_{h(n-2)-1}} & 0_{d_o}
    \end{bmatrix}}_{d_o \times n} \cdot I_n \notag \\
    = & ~
    \begin{bmatrix}
        0_{d_o} & \tilde{L}_{(h-1)(n-2)}e_{\tilde{k}_{(h-1)(n-2)}} & \tilde{L}_{(h-1)(n-2)+1}e_{\tilde{k}_{(h-1)(n-2)}+1} & \cdots & \tilde{L}_{h(n-2)-1}e_{\tilde{k}_{h(n-2)-1}} & 0_{d_o}
    \end{bmatrix} \notag \\
     \in&~\R^{d_0\times n}.
\end{align}

Given that all $\tilde{k}_j$, where $j\in[p]$, share the same  number in $[d_o]$, we denote this number by $k_G$.

Hence we rewrite $V^{(h)}$ as
\begin{align*}
V^{(h)}
=
\begin{bmatrix}
        0_{d_o} & \tilde{L}_{(h-1)(n-2)}e_{k_G} & \tilde{L}_{(h-1)(n-2)+1}e_{k_G} & \cdots & \tilde{L}_{h(n-2)-1}e_{k_G} & 0_{d_o}
    \end{bmatrix}.
\end{align*}

We define $m_v$ as
\begin{align*}
    m_v := \max\{|a|,|b|\}.
\end{align*}
By the definition of $V^{(h)}$, we have
\begin{align}\label{eqn:max_v}
\|V\|_\infty \leq \max_{i\in [P]}\{\tilde{L}_i\} \leq m_v.
\end{align}

\begin{claimbox}
\begin{remark}[Intuition of the Construction of $V^{(h)}$]
As previously mentioned, $\tilde{L}_i$, for $ i\in [p]$, are all the interpolation points.
In this context, $V^{(h)}$ encompasses the $(n-2)$ elements of these interpolations (i.e., $(h-1)(n-2)$ to $h(n-2)-1$).
Meanwhile, the value on the two ends of $V^{h}$ are both set to $0_{d_o}$, because we suppress the head and let it output $0$ when the input $X$ is not close enough to the interpolations of the head.
\end{remark}
\end{claimbox}

Now we are ready to calculate the output of each ${\rm Attn}_h$
\begin{align*}
    & ~ {\rm Attn}_h(A(X))
    \\
    = & ~ V^{(h)}\Softmax((K^{(h)})^\top Q^{(h)}) \nonumber\\
    = & ~
    V^{(h)}
    \Softmax
    \left(
    -\beta
    \begin{bmatrix}
        -2[(h-1)(n-2)-1]w & -2(h-1)(n-2)w & \cdots & -2h(n-2)w \\
        \ell_{(h-1)(n-2)-1} & \ell_{(h-1)(n-2)} & \cdots & \ell_{h(n-2)}
    \end{bmatrix}^\top
    \begin{bmatrix}
        X \\
        1_{1\times n}
    \end{bmatrix}
    \right),
\end{align*}
where last line is by plug in \eqref{eqn:mutihead_k} and \eqref{eqn:mutihead_q}.
Note the $i$-th column of the attention score matrix (the $\Softmax$ nested expression) is equivalent to the following expressions
\begin{align}
    & ~ \Softmax((K^{(h)})^\top Q^{(h)})_{:,i} \notag\\
    = & ~
    \Softmax\left(-\beta
        \begin{bmatrix}
        -2[(h-1)(n-2)-1]w & -2(h-1)(n-2)w & \cdots & -2h(n-2)w \\
        \ell_{(h-1)(n-2)-1} & \ell_{(h-1)(n-2)} & \cdots & \ell_{h(n-2)}
    \end{bmatrix}^\top
    \begin{bmatrix}
        X \\
        1_{1\times n}
    \end{bmatrix}
    \right)_{:,i} \notag \\
    = & ~
    \Softmax
    \left(-\beta
    \begin{bmatrix}
        -2[(h-1)(n-2)-1]w^\top x_i +\ell_{(h-1)(n-2)-1}\\
        -2(h-1)(n-2)w^\top x_i+ \ell_{(h-1)(n-2)}\\
        \vdots\\
        -2h(n-2)w^\top x_i +\ell_{h(n-2)}
    \end{bmatrix}
    \right) \annot{pick column $i$} \\
    = & ~
    \Softmax
   \left(-\beta
    \begin{bmatrix}
        [(h-1)(n-2)-1](-2w^\top x_i +\tilde{L}_{(h-1)(n-2)-1}+\tilde{L}_0)-2[(h-1)(n-2)-1]t\\
        (h-1)(n-2)(-2w^\top x_i +\tilde{L}_{(h-1)(n-2)}+\tilde{L}_0)-2(h-1)(n-2)t\\
        \vdots\\
        h(n-2)(-2w^\top x_i +\tilde{L}_{h(n-2)}+\tilde{L}_0)-2h(n-2)t
    \end{bmatrix}
    \right) \annot{By $\ell_k = k(\tilde{L}_k + \tilde{L}_0) - 2kt$}\\
    = & ~
    \Softmax\left(-\frac{\beta}{\Delta L}
    \begin{bmatrix}
        (-2x_i^\top w-2t+\tilde{L}_0+\tilde{L}_{(h-1)(n-2)-1})\cdot [(h-1)(n-2)-1]\Delta L\\
        (-2x_i^\top w-2t+\tilde{L}_0+\tilde{L}_{(h-1)(n-2)})\cdot (h-1)(n-2)\Delta L \\
        \vdots\\
        (-2x_i^\top w-2t+\tilde{L}_0+\tilde{L}_{h(n-2)})\cdot h(n-2)\Delta L
    \end{bmatrix}
    \right) \annot{By mutiplying and dividing by $\Delta L$}\\
    = & ~
    \Softmax\left(
    -\frac{\beta}{\Delta L}
    \begin{bmatrix}
        (-2x_i^\top w-2t+\tilde{L}_0+\tilde{L}_{(h-1)(n-2)-1})\cdot (\tilde{L}_{(h-1)(n-2)-1}-\tilde{L}_0) \\
        (-2x_i^\top w-2t+\tilde{L}_0+\tilde{L}_{(h-1)(n-2)})\cdot (\tilde{L}_{(h-1)(n-2)}-\tilde{L}_0) \\
        \vdots \\
        (-2x_i^\top w-2t+\tilde{L}_0+\tilde{L}_{h(n-2)})\cdot (\tilde{L}_{h(n-2)}-\tilde{L}_0)
    \end{bmatrix}
    \right) \annot{By $k\Delta L=\tilde L_k-\tilde L_0$}\\
    = & ~
    \Softmax\left(
    -\frac{\beta}{\Delta L}
    \begin{bmatrix}
        (-2x_i^\top w-2t)\cdot \tilde{L}_{(h-1)(n-2)-1}+(\tilde{L}_{(h-1)(n-2)-1})^2+(x_i^\top w+t)^2 \\
        (-2x_i^\top w-2t)\cdot \tilde{L}_{(h-1)(n-2)}+(\tilde{L}_{(h-1)(n-2)})^2+(x_i^\top w+t)^2 \\
        \vdots \\
        (-2x_i^\top w-2t)\cdot \tilde{L}_{h(n-2)}+(\tilde{L}_{h(n-2)})^2+(x_i^\top w+t)^2 \\
    \end{bmatrix}
    \right) \notag \\
    = & ~
    \Softmax\left(
    -\frac{\beta}{\Delta L}
    \begin{bmatrix}
        (x_i^\top w+t-\tilde{L}_{(h-1)(n-2)-1})^2 \\
        (x_i^\top w+t-\tilde{L}_{(h-1)(n-2)})^2 \\
        \vdots \\
        (x_i^\top w+t-\tilde{L}_{h(n-2)})^2 \\
    \end{bmatrix}
    \right).
    \label{eqn:mul_softmax_k_q}
\end{align}
Here, the second-last equality arises from the fact that the softmax function is shift-invariant, allowing us to subtract and add a constant across all coordinates.
To be more precise, we first expand the product for $k$-th coordinate of the column vector
\begin{align*}
    & ~ (-2x_i^\top w - 2t + \tilde L_0 + \tilde L_k)(\tilde L_k - \tilde L_0) \\
    = & ~ (-2x_i^\top w - 2t)L_k + L_0L_k + L_k^2 -(-2x_i^\top w - 2t)L_0 -L_0^2- L_0L_k \\
    = & ~(-2x_i^\top w - 2t)L_k  + L_k^2 -\underbrace{(-2x_i^\top w - 2t)L_0 -L_0^2}_{\text{constant across the column vector}}.
\end{align*}
Then, dropping the constant and adding another constant $(x_i^\top w + t)^2$ across all coordinates, the above equation becomes
\begin{align*}
    & ~(-2x_i^\top w - 2t)L_k  + L_k^2 + (x_i^\top w + t)^2
    = (x_i^\top w + t - L_k)^2.
\end{align*}

Hence we finish the derivation of \eqref{eqn:mul_softmax_k_q}.
Thus we have
\begin{align}
\label{eq:expression_attn_matrix}
    {\rm Attn}_h(A(X))_{:,i} = V^{(h)}
    \Softmax\left(
    -\frac{\beta}{\Delta L}
    \begin{bmatrix}
        (x_i^\top w+t-\tilde{L}_{(h-1)(n-2)-1})^2 \\
        (x_i^\top w+t-\tilde{L}_{(h-1)(n-2)})^2 \\
        \vdots \\
        (x_i^\top w+t-\tilde{L}_{h(n-2)})^2 \\
    \end{bmatrix}
    \right).
\end{align}

For a specific $h$, we calculate the result of \eqref{eq:expression_attn_matrix} column by column.
Let $X_i$ denote any column (token) of the matrix $X$.
We partition the situation at each column (token) into three distinct cases:
\begin{itemize}
    \item \textbf{Case 1:}
    $w^\top X_i+t$ is strictly within the interpolation range of ${\rm Attn}_h$ ($X\in[\tilde{L}_{(h-1)(n-2)}, \tilde{L}_{h(n-2)-1}]$).
    This excludes the following range at the edge of the interpolation range of
    \begin{align*}
       [\tilde{L}_{(h-1)(n-2)-1}, \tilde{L}_{(h-1)(n-2)}]\cup [\tilde{L}_{h(n-2)-1}, \tilde{L}_{h(n-2)}].
    \end{align*}

    \item \textbf{Case 2:}
    $w^\top X_i+t$ is not within the interpolation range of ${\rm Attn}_h$:
    \begin{align*}
    w^\top X_i+t\notin [\tilde{L}_{(h-1)(n-2)-1}, \tilde{L}_{h(n-2)}].
    \end{align*}

    \item \textbf{Case 3:}
    $w^\top X_i+t$ is on the edge (region) of the interpolation range of ${\rm Attn}_h$:
    \begin{align*}
         w^\top X_i+t \in [\tilde{L}_{(h-1)(n-2)-1}, \tilde{L}_{(h-1)(n-2)}]\cup [\tilde{L}_{h(n-2)-1}, \tilde{L}_{h(n-2)}].
    \end{align*}
\end{itemize}

Two remarks are in order.
\vspace{1em}
\begin{claimbox}
\begin{remark}[Cases of a Single Head Attention]
The $H$ heads split the approximation of the truncated linear map across disjoint intervals. For head $h$,
\begin{align*}
    \|{\rm Attn}_h(X) - {\rm Range}_{[a+\frac{b-a}{p}((h-1)(n-2)-1),a+\frac{b-a}{p}h(n-2)]}(X)\|_\infty \leq
    \epsilon_1,
\end{align*}
where $\epsilon>0$ is arbitrarily small.

With this understanding, $w^\top X_i + t$:
\begin{itemize}
    \item \textbf{Case 1:}
    falls into the interior of the interpolation range of the $h$-th head ${\rm Attn}_h$, denoted as ${\rm Range}_{[a+(b-a)((h-1)(n-2)-1)/p,a+(b-a)h(n-2)/p]}$.

    \item \textbf{Case 2:}
    remains outside of the interpolation range of the $h$-th head ${\rm Attn}_h$.

    \item \textbf{Case 3:}
    falls on the boundary of the interpolation range of the $h$-th head ${\rm Attn}_h$.
\end{itemize}
\end{remark}
\end{claimbox}

\begin{claimbox}
\begin{remark}[Cases of All Attention Heads]
According to \cref{lem:all_heads_cases}, for all heads in ${\rm Attn}^H$, there are two possible cases:
\vspace{0.5em}
    \begin{itemize}
        \item \textbf{Case 1*:} $x$ falls into Case 1 for a head, and Case 2 for all other heads.
        \vspace{0.5em}
        \item \textbf{Case 2*:} $x$ falls into Case 3 for two heads with adjacent interpolation ranges, and Case 2 for other heads.
    \end{itemize}
\vspace{0.5em}
This also means that when Case 1 appears in ${\rm Attn}^H$, the situation of all head in ${\rm Attn}^H$ falls into Case 1*.
And when Case 3 appears in ${\rm Attn}^H$, the situation of all head in ${\rm Attn}^H$ falls into Case 2*.
Thus, We discuss Case 2* in the discussion of Case 3.
\end{remark}
\end{claimbox}

\paragraph{Case 1: $X_i \in[\tilde{L}_{(h-1)(n-2)}, \tilde{L}_{h(n-2)-1}]$.}

In this case, our goal is to demonstrate this attention head outputs a value close to ${\rm Range}_{[a,b]}(w^\top X_i+t)$.

Let $\tilde{L}_{s}$ and $\tilde{L}_{s+1}$ be the two interpolants such that
\begin{align}
\label{eq:s_near_target}
    w^\top X_i +t \in [\tilde{L}_{s},\tilde{L}_{s+1}].
\end{align}

Then, $s$ and $s+1$ are also the labels of the two largest entries in
\begin{align*}
    -\frac{\beta}{\Delta L}
    \begin{bmatrix}
        (w^\top X_i +t-\tilde{L}_{(h-1)(n-2)-1})^2 \\
        (w^\top X_i +t-\tilde{L}_{(h-1)(n-2)})^2 \\
        \vdots \\
        (w^\top X_i +t-\tilde{L}_{h(n-2)})^2 \\
    \end{bmatrix},
\end{align*}
since
\begin{align*}
    & ~ \argmax_{k\in \{(h-1)(n-2)-1,h(n-2)\}}
    -\frac{\beta}{\Delta L}(w^\top X_i+t-\tilde{L}_{k})^2 \\
    = & ~
    \argmin_{k\in \{(h-1)(n-2)-1,h(n-2)\}}(w^\top X_i+t-\tilde{L}_{k})^2\\
    = & ~
    \argmin_{k\in \{(h-1)(n-2)-1,h(n-2)\}}|w^\top X_i+t-\tilde{L}_{k}|.
\end{align*}

We also note that the distance of $w^\top X_i +t$ to interpolants beside $\tilde{L}_s$ and $\tilde{L}_{s+1}$ differs from $w^\top X_i+t$ for at least $\tilde{L}_s-\tilde{L}_{s-1} = (b-a)/p$ or $\tilde{L}_{s+1}-\tilde{L}_{s} = (b-a)/p$.

This is equivalent to the occasion when $x_1 - x_3$ in \cref{lem:Soft_to_Hard} is larger than
\begin{align*}
    & \max \Big\{\frac{\beta}{\Delta L}
    (
    w^\top X_i +t - \tilde{L}_{s-1})^2
    -
    (w^\top X_i +t - \tilde{L}_s
    )^2
    ,
    \frac{\beta}{\Delta L}
    (
    w^\top X_i +t - \tilde{L}_{s+2})^2
    -
    (w^\top X_i +t - \tilde{L}_{s+1}
    )^2
    \Big\} \\
    & \geq
    \frac{\beta}{\Delta L} \cdot (\frac{b-a}{p})^2,
\end{align*}
which is invariant to $X_i$.

Thus according to \cref{lem:Soft_to_Hard} and the fact that the $s$ and $s+1$ are the two largest entries in the $i$-th column of the attention score matrix, we have
\begin{align*}
    \Bigl\|
    \Softmax\left(
    -\frac{\beta}{\Delta L}
    \begin{bmatrix}
        (w^\top X_i +t-\tilde{L}_{(h-1)(n-2)-1})^2 \\
        (w^\top X_i +t-\tilde{L}_{(h-1)(n-2)})^2 \\
        \vdots \\
        (w^\top X_i +t-\tilde{L}_{h(n-2)})^2 \\
    \end{bmatrix}
    \right)
    - \frac{1}{1 + e^{-\beta \delta}} \underbrace{e_{s}}_{n\times 1}
        - \frac{e^{-\beta \delta}}{1 + e^{-\beta \delta}} \underbrace{e_{s+1}}_{n\times 1}
    \Bigr\|_\infty
    &\le \epsilon_2,
\end{align*}
for any $\epsilon_2 > 0$.

This yields that
\begin{align*}
    & ~\Bigl\|
    V
    \Softmax\left(
    -\frac{\beta}{\Delta L}
    \begin{bmatrix}
        (w^\top X_i +t-\tilde{L}_{(h-1)(n-2)-1})^2 \\
        (w^\top X_i +t-\tilde{L}_{(h-1)(n-2)})^2 \\
        \vdots \\
        (w^\top X_i +t-\tilde{L}_{h(n-2)})^2 \\
    \end{bmatrix}
    \right)
    -
    V\frac{1}{1 + e^{-\beta \delta}} e_{s}
        - V\frac{e^{-\beta \delta}}{1 + e^{-\beta \delta}} e_{s+1}
    \Bigr\|_\infty  \\
    \leq & ~
    \Bigl\|
    \Softmax\left(
    -\frac{\beta}{\Delta L}
    \begin{bmatrix}
        (w^\top X_i +t-\tilde{L}_{(h-1)(n-2)-1})^2 \\
        (w^\top X_i +t-\tilde{L}_{(h-1)(n-2)})^2 \\
        \vdots \\
        (w^\top X_i +t-\tilde{L}_{h(n-2)})^2 \\
    \end{bmatrix}
    \right)
    - \frac{1}{1 + e^{-\beta \delta}} e_{s}
        - \frac{e^{-\beta \delta}}{1 + e^{-\beta \delta}} e_{s+1}
    \Bigr\|_\infty \cdot \|V\|_\infty \\
    \leq & ~ \|V\|_\infty \epsilon_2.
\end{align*}

This is equivalent to
\begin{align}
    & ~ \|V\Softmax(K^\top Q)_{:,i}
    -
    \frac{1}{1 + e^{-\beta \delta}} \tilde{L}_{(h-1)(n-2)+s-1}e_{k_G}
    -
    \frac{e^{-\beta \delta}}{1 + e^{-\beta \delta}} \tilde{L}_{(h-1)(n-2)+s}e_{k_G}
    \|_\infty \notag \\
    \leq & ~
    \|V\|_\infty \cdot \epsilon_2
    \annot{By $\|AB\|\le \|A\|\cdot \|B\|$}\\
    \leq & ~
    m_v\epsilon_2,
    \label{eq:VsoftmaxKTQ-Lh}
\end{align}
where the last line is by \eqref{eqn:max_v}.

From \eqref{eq:s_near_target}, we derive that
\begin{align}
    & ~ \|
    \frac{1}{1 + e^{-\beta \delta}} \tilde{L}_{(h-1)(n-2)+s-1}
    +
    \frac{e^{-\beta \delta}}{1 + e^{-\beta \delta}} \tilde{L}_{(h-1)(n-2)+s}
    -
    (w^\top X_i+t)e_{k_G}
    \|_\infty \nonumber\\
    \leq & ~
    \|
    \frac{1}{1 + e^{-\beta \delta}} (\tilde{L}_{(h-1)(n-2)+s-1}
    -
    (w^\top X_i+t)e_{k_G})
    \|_\infty
    +
    \|
    \frac{e^{-\beta \delta}}{1 + e^{-\beta \delta}} (\tilde{L}_{(h-1)(n-2)+s}
    -
    (w^\top X_i+t))
    \|_\infty \annot{By convex combination of $(w^\top X_i+t)$ and triangle inequality} \\
    \leq & ~
    \frac{1}{1 + e^{-\beta \delta}} \cdot \frac{b-a}{p}
    +
    \frac{e^{-\beta \delta}}{1 + e^{-\beta \delta}} \cdot \frac{b-a}{p} \annot{By \eqref{eq:s_near_target}} \\
    = & ~
    \frac{b-a}{p} .
    \label{eq:Lh-f}
\end{align}

Combing \eqref{eq:VsoftmaxKTQ-Lh} and \eqref{eq:Lh-f} yields
\begin{align}
    & ~ \|
    V\Softmax(K^\top Q)_{:,i}
    -
    (w^\top X_i+t)
    \|_\infty \nonumber\\
    \leq & ~
    \|
    V\Softmax(K^\top Q)_{:,i}
    -
    \frac{1}{1 + e^{-\beta \delta}} \tilde{L}_{(h-1)(n-2)+s-1}
    -
    \frac{e^{-\beta \delta}}{1 + e^{-\beta \delta}} \tilde{L}_{(h-1)(n-2)+s}
    \|_\infty \nonumber\\
    & ~ +
    \|
    \frac{1}{1 + e^{-\beta \delta}} \tilde{L}_{(h-1)(n-2)+s-1}
    +
    \frac{e^{-\beta \delta}}{1 + e^{-\beta \delta}} \tilde{L}_{(h-1)(n-2)+s}
    -
    (w^\top X_i+t)e_{k_G}
    \|_\infty \annot{By triangle inequality} \\
    \leq & ~
    \label{eq:case1_output}
    m_v\epsilon_2+\frac{b-a}{p},
\end{align}
where the first inequality comes from adding and subtracting the interpolation points' convex combination and then applying triangle inequality.

\paragraph{Case 2: $X\notin [\tilde{L}_{(h-1)(n-2)-1}, \tilde{L}_{h(n-2)}]$.}

In this case, $X_i$ falls out of the range of interpolation covered by ${\rm Attn}_h$.

Without loss of generality, suppose $w^\top X_i+t$ to lie left to the range of interpolation of ${\rm Attn}_h$.

This yields that $\tilde{L}_{(h-1)(n-2)-1}$ is the closest interpolant within ${\rm Attn}_h$ to $w^\top X_i+t$.
Furthermore, the second closest interpolant $\tilde{L}_{(h-1)(n-2)}$ is at least further for at least $(b-a)/p$, which is a constant irrelevant to $X_i$

Then by \cref{lem:Soft_to_Hard}, we have
\begin{align*}
    \Bigl\|
    \Softmax\left(
    -\frac{\beta}{\Delta L}
    \begin{bmatrix}
        (w^\top X_i +t-\tilde{L}_{(h-1)(n-2)-1})^2 \\
        (w^\top X_i +t-\tilde{L}_{(h-1)(n-2)})^2 \\
        \vdots \\
        (w^\top X_i +t-\tilde{L}_{h(n-2)})^2 \\
    \end{bmatrix}
    \right)
    - \underbrace{e_1}_{n\times 1}
    \Bigr\|_\infty
    \leq
    \epsilon_3,
\end{align*}
for any $\epsilon_3>0$.

This yields that
\begin{align*}
    & ~ \|
    V\Softmax\left(
    -\frac{\beta}{\Delta L}
    \begin{bmatrix}
        (w^\top X_i +t-\tilde{L}_{(h-1)(n-2)-1})^2 \\
        (w^\top X_i +t-\tilde{L}_{(h-1)(n-2)})^2 \\
        \vdots \\
        (w^\top X_i +t-\tilde{L}_{h(n-2)})^2 \\
    \end{bmatrix}
    \right)
    - V\underbrace{e_1}_{n\times 1}
    \|_\infty \\
    \leq & ~
    \|V\|_\infty \cdot \epsilon_3
    \annot{By $\|AB\|\le \|A\|\cdot \|B\|$}\\
    \leq & ~
    m_v\epsilon_3,
\end{align*}
where the last line is by \eqref{eqn:max_v}.

This is equivalent to
\begin{align}
    \label{eq:case2_output}
    \Bigl\|
    V\Softmax\left(
    -\frac{\beta}{\Delta L}
    \begin{bmatrix}
        (w^\top X_i +t-\tilde{L}_{(h-1)(n-2)-1})^2 \\
        (w^\top X_i +t-\tilde{L}_{(h-1)(n-2)})^2 \\
        \vdots \\
        (w^\top X_i +t-\tilde{L}_{h(n-2)})^2 \\
    \end{bmatrix}
    \right)
    - 0_{d_o}
    \Bigr\|_\infty
    \leq
    m_v\epsilon_3.
\end{align}

\paragraph{Case 1*.}
According to \cref{lem:all_heads_cases}, when Case 1 occurs for one head in the $H$ heads of ${\rm Attn}^H$, all other head will be in Case 2.

Combining with the result in Case 2, we have the output of all heads as
\begin{align*}
& ~ \|{\rm Attn}^H(A(X))_{:,i}-(w^\top X_i+t)e_{k_G}\|_\infty \\
= & ~ %
\|\sum_{h_0\in [H]/\{h\}}{\rm Attn}_{h_0}\circ A(X)_{:,i}\|_\infty
+
\|{\rm Attn}_h\circ A(X)_{:,i}-(w^\top X_i+t)e_{k_G}\|_\infty\\
= & ~ %
(H-1)m_v\epsilon_3 + m_v\epsilon_2+\frac{b-a}{p}
\annot{By \eqref{eq:case1_output} and \eqref{eq:case2_output}}\\
= & ~
(H-1)m_v\epsilon_3 + m_v\epsilon_2+\frac{b-a}{H(n-2)}
.
\end{align*}

Setting $\epsilon_2,\epsilon_3$ to be
\begin{align*}
& ~ \epsilon_2 =\frac{\epsilon_0}{2}
\\
& ~ \epsilon_3 = \frac{\epsilon_0}{2(H-1)m}
\end{align*}
yields the final result.

\paragraph{Case 3 (and Case 2*): $X \in [\tilde{L}_{(h-1)(n-2)-1}, \tilde{L}_{(h-1)(n-2)}]\cup [\tilde{L}_{h(n-2)-1}, \tilde{L}_{h(n-2)}]$.}

In this case, $w^\top X_i + t$ is the boundary of the interpolation range of ${\rm Attn}_{h_0}$.
By \cref{lem:all_heads_cases}, it should also fall on the boundary of a head with neighboring interpolation range.
Without loss of generality,  we set it to be ${\rm Attn}_{h_0-1}$.
Furthermore, \cref{lem:all_heads_cases} indicates that $w^\top X_i +t$ should fall on no other interpolation range of any heads beside ${\rm Attn}_{h_0}$ and ${\rm Attn}_{h_0-1}$.

Combining this with case 2, we have
\begin{align*}
    {\rm Attn}^H(A(X))_{:,i}
    = & ~
    \sum_{h=1}^H {\rm Attn}_h\circ A(X)_{:,i}\\
    & ~ \in
    [(-(H-2)m_v\epsilon_3 +
    {\rm Attn}_{h_0}\circ A(X)_{:,i}
    +
    {\rm Attn}_{h_0-1}\circ A(X)_{:,i}),\\
    &~~~ \quad
    ((H-2)m_v\epsilon_3 +
    {\rm Attn}_{h_0}\circ A(X)_{:,i}
    +
    {\rm Attn}_{h_0-1}\circ A(X)_{:,i})]
    \annot{By \eqref{eq:case2_output}}.
\end{align*}

By \cref{lem:Soft_to_Hard}, let $\delta$ denote
\begin{align*}
    \delta =
    \tilde{L}_{(h-1)(n-2)+s}-(w^\top X_i +t)e_{k_G}-[\tilde{L}_{(h-1)(n-2)+s}-(w^\top X_i +t)e_{k_G}],
\end{align*}
we have
\begin{align*}
    \|
    \Softmax((K^{(h)})^\top Q^{(h)})
    -
    (\frac{1}{1 + e^{-\beta \delta}} e_{1}
    +
    \frac{e^{-\beta \delta}}{1 + e^{-\beta \delta}} e_{2})
    \|
    \leq
    \epsilon_4,
\end{align*}
and
\begin{align*}
    \|
    \Softmax((K^{(h-1)})^\top Q^{(h-1)})
    -
    (\frac{1}{1 + e^{-\beta \delta}} e_{n-1}
    +
    \frac{e^{-\beta \delta}}{1 + e^{-\beta \delta}} e_{n})
    \|
    \leq
    \epsilon_5,
\end{align*}
for any $\epsilon_4,\epsilon_5 >0$.

Thus we have
\begin{align*}
    & ~ \|V^{(h)}\Softmax((K^{(h)})^\top Q^{(h)})
    +
    V^{(h-1)}\Softmax((K^{(h-1)})^\top Q^{(h-1)})\\
    & ~ \hspace{6em} -
    V(\frac{1}{1 + e^{-\beta \delta}} e_{1}
    +
    \frac{e^{-\beta \delta}}{1 + e^{-\beta \delta}} e_{2}
    +
    \frac{1}{1 + e^{-\beta \delta}} e_{n-1}
    +
    \frac{e^{-\beta \delta}}{1 + e^{-\beta \delta}} e_{n})
    \|_\infty \\
    \leq & ~
    \|V\|_\infty(\epsilon_4+\epsilon_5).
\end{align*}
This is equivalent to
\begin{align*}
    & ~ \|V^{(h)}\Softmax((K^{(h)})^\top Q^{(h)})
    +
    V^{(h-1)}\Softmax((K^{(h-1)})^\top Q^{(h-1)})\\
    & ~~ -
    (\frac{1}{1 + e^{-\beta \delta}}\cdot 0
    +
    \frac{e^{-\beta \delta}}{1 + e^{-\beta \delta}} e_{k_G}\tilde{L}_{(h-1)(n-2)+s}
    +
    \frac{1}{1 + e^{-\beta \delta}} e_{k_G}\tilde{L}_{(h-1)(n-2)+s-1}
    +
    \frac{e^{-\beta \delta}}{1 + e^{-\beta \delta}} e_{k_G})\cdot 0
    \|_\infty \\
    \leq & ~
    \|V\|_\infty \cdot (\epsilon_4+\epsilon_5).
\end{align*}

Thus we have
\begin{align*}
    & ~ \|V^{(h)}\Softmax((K^{(h)})^\top Q^{(h)})
    +
    V^{(h-1)}\Softmax((K^{(h-1)})^\top Q^{(h-1)})\\
    & ~ \quad\quad\quad -
    (
    \frac{e^{-\beta \delta}}{1 + e^{-\beta \delta}} e_{k_G}\tilde{L}_{(h-1)(n-2)+s}
    +
    \frac{1}{1 + e^{-\beta \delta}} e_{k_G}\tilde{L}_{(h-1)(n-2)+s-1}
    )
    \|_\infty \\
    \leq & ~
    \|V\|_\infty(\epsilon_4+\epsilon_5),
\end{align*}
which implies
\begin{align}\label{eqn:mul_attn_two)close}
    & ~ \|
    \sum_{h=1}^H {\rm Attn}_h(A(X))_{:,i}
    -
    (
    \frac{e^{-\beta \delta}}{1 + e^{-\beta \delta}} e_{k_G}\tilde{L}_{(h-1)(n-2)+s}
    +
    \frac{1}{1 + e^{-\beta \delta}} e_{k_G}\tilde{L}_{(h-1)(n-2)+s-1}
    )
    \|_\infty \notag\\
    \leq & ~
    (H-2)m_v\epsilon_3+\|V\|_\infty(\epsilon_4+\epsilon_5).
\end{align}

Finally, since
\begin{align*}
    \|
    \frac{e^{-\beta \delta}}{1 + e^{-\beta \delta}} e_{k_G}\tilde{L}_{(h-1)(n-2)+s}
    +
    \frac{1}{1 + e^{-\beta \delta}} e_{k_G}\tilde{L}_{(h-1)(n-2)+s-1}
    -
    (w^\top X_i +t)e_{k_G}
    \|_\infty
    \leq
    \frac{b-a}{p} \annot{By \eqref{eq:Lh-f}},
\end{align*}
combining with \eqref{eqn:mul_attn_two)close}, we have
\begin{align*}
    & ~ \|
    \sum_{h=1}^H {\rm Attn}_h(A(X))_{:,i}
    -
    (w^\top X_i +t)e_{k_G}
    \|_\infty \\
    \leq & ~
    \|
    \sum_{h=1}^H {\rm Attn}_h(A(X))_{:,i}
    -
    (\frac{e^{-\beta \delta}}{1 + e^{-\beta \delta}} e_{k_G}\tilde{L}_{(h-1)(n-2)+s}
    +
    \frac{1}{1 + e^{-\beta \delta}} e_{k_G}\tilde{L}_{(h-1)(n-2)+s-1})
    \|_\infty \\
    & ~  +
    \|
    (\frac{e^{-\beta \delta}}{1 + e^{-\beta \delta}} e_{k_G}\tilde{L}_{(h-1)(n-2)+s}
    +
    \frac{1}{1 + e^{-\beta \delta}} e_{k_G}\tilde{L}_{(h-1)(n-2)+s-1})
    -
    (w^\top X_i +t)e_{k_G}
    \|_\infty \annot{By triangle inequality}\\
    \leq & ~
    \frac{b-a}{p}
    +
    (H-2)m_v\epsilon_3+\|V\|_\infty(\epsilon_4+\epsilon_5) \\
    \leq & ~
    \frac{b-a}{H(n-2)}+(H-2)\max\{|a|,|b|\}\epsilon_3+\max\{|a|,|b|\}(\epsilon_4+\epsilon_5).
\end{align*}

Setting $\epsilon_3,\epsilon_4,\epsilon_5$ to be
\begin{align*}
& ~ \epsilon_3 = \frac{\epsilon_0}{3(H-2)} \\
& ~ \epsilon_4 = \epsilon_5 = \frac{\epsilon_0}{3}
\end{align*}
yields the final result.

This completes the proof.
\end{proof}

\begin{lemma}[Attention Prepended with Token-Wise Linear Transformation is Still a Transformer]
\label{lem:still_attention}
For any attention $\Attn$ and any linear transformation $A$, $\Attn\circ A$ is still an attention.
\end{lemma}
\begin{proof}
We denote the transformation matrix of $A$ also as $M_A$.
Denote the attention $\Attn$ as
\begin{align*}
\Attn(Z) :=
W_VZ \Softmax((W_KZ )^\top W_Q Z).
\end{align*}
Then we have
\begin{align*}
\Attn\circ A(Z)
=
W_V M_A Z
\Softmax((W_K M_AZ )^\top W_QM_A Z).
\end{align*}
It is a new attention with parameters $W_KM_A$,$W_QM_A$ and $W_VM_A$.
\end{proof}

\begin{lemma}[Lemma $14$ in \cite{bai2023transformers}: Composition of Error for Approximating Convex GD]
\label{lem:comp-error-convex-gd}
Suppose $f:\R^d\to\R$ is a convex function. Let $ w^\star\in\argmin_{w\in\R^d} f(w)$, $R\geq 2\|w^\star\|_2$, and assume that $\nabla f$ is $L_f$-smooth on $B_2^d(R)$. Let sequences $\{\hw^\ell\}_{\ell\ge 0}\subset \R^d$ and $\{w_{\gd}^\ell\}_{\ell\ge 0}\subset \R^d$ be given by $\hw^{0}=w_{\gd}^0=\mathbf{0}$,
\begin{align*}
\left\{
\begin{aligned}
    & \hw^{\ell+1}=\hw^\ell-\eta \nabla f(\hw^\ell)+\epsilon^\ell, \qquad \|\epsilon^{\ell}\|_2 \leq \epsilon, \\
    & w_{\gd}^{\ell+1} = w_{\gd}^\ell - \eta\nabla f(w_{\gd}^\ell),
\end{aligned}
\right.
\end{align*}
for all $\ell\ge 0$. Then as long as $\eta\leq 2/L_f$, for any $0\leq L\leq R/(2\epsilon)$, it holds that $\|\hw^L - w_{\gd}^L\|_2\leq L\epsilon$ and $\|\hw^L\|_2\leq \frac{R}2+L\epsilon\leq R$.
\end{lemma}

\begin{corollary}[Corollary A.2 in \cite{bai2023transformers}: Gradient Descent for Smooth and Strongly Convex Function]
\label{cor:gd-smooth-strongly-convex}
    Suppose $L:\R^d\to\R$ is a $\alpha$-strongly convex and $\beta$-smooth for som $0<\alpha\leq\beta$. Then, the gradient descent iterates $w_{\gd}^{(t+1)} := w_{\gd}^t - \eta\nabla L(w_{\gd}^L)$ with learning rage $\eta=1/\beta$ and initialization $w_{\gd}^0\in\R^d$ satisfies for any $t\geq1$,
    \begin{align*}
        &\|w_{\mathrm{GD}}^t - w^\star\|_2^2 \le \exp(-\frac{t}{\kappa}) \cdot \|w_{\mathrm{GD}}^0 - w^\star\|_2^2, \\  &L(w_{\mathrm{GD}}^t) - L(w^\star) \le \frac{\beta}{2} \exp(-\frac{t}{\kappa}) \cdot \|w_{\mathrm{GD}}^0 - w^\star\|_2^2,
    \end{align*}
    where $\kappa:=\beta/\alpha$ is the condition number of $L$, and $w^\star := \argmin_{w\in\R^d} L(w)$.
\end{corollary}

\subsection{Proof of \texorpdfstring{\cref{thm:attn_sim_f}}{}}
\label{proof:thm:attn_sim_f}

\begin{theorem}[In-Context Emulation of $f(w^\top x - y)x$ with Single-Head Attention;  \cref{thm:attn_sim_f} Restate]
Let
\begin{align*}
    X := \begin{bmatrix}
        x_1 & x_2 & \cdots & x_n \\
        y_1 & y_2 & \cdots & y_n
    \end{bmatrix} \in\R^{(d+1)\times n}
    \quad\text{and}\quad
    W :=
    \begin{bmatrix}
        w & w & \cdots & w
    \end{bmatrix}\in\R^{d\times n},
\end{align*}
where $x_i \in \R^d$, $y_i \in \R$, and $w \in \R^d$ is the coefficient vector.
Define the input as:
\begin{align}
    Z :=
    \begin{bmatrix}
        x_1 & x_2 & \cdots & x_n \\
        y_1 & y_2 & \cdots & y_n \\
        w & w & \cdots & w
    \end{bmatrix}
    =\begin{bmatrix}
        X\\W
    \end{bmatrix} \in \R^{(2d+1)\times n}.
\end{align}
Assume $\max\{ \|X\|_\infty, \|W\|_\infty \} \le B$.
For any continuous function $f:\R \to \R$ and any $\epsilon > 0$, there exists a single-head attention $\Attn_s$ with a linear layer $\li$ such that
\begin{align*}
    \|
    \Attn_s \circ \li(Z)
    -
    \begin{bmatrix}
        f(w^\top x_1-y_1)x_1 &
        \cdots
        &
        f(w^\top x_n-y_n)x_n
    \end{bmatrix}
    \|_\infty
    \leq
    \epsilon,
    \quad\text{for any}\quad
    \epsilon > 0.
\end{align*}
\end{theorem}

\begin{proof}

We define the linear transformation of $Z$ as a concatenation of two functions:
\begin{align*}
    \li(Z) := [\underbrace{\li_x(X)}_{(2d+n+2)\times n(P+1)}  \quad \underbrace{\li_w(W)}_{(2d+n+2)\times n}] \in\R^{(2d+n+2) \times (n(P+1)+n)},
\end{align*}
where $\li_x$ and $\li_w$ are defined as below.

We define $\li_w$ as:
\begin{align*}
    \li_w(W) :=
    \begin{bmatrix}
        I_d \\
        0_{(d+n+2)\times d}
    \end{bmatrix}
    W
    +
    \begin{bmatrix}
        0_{d\times n} \\
        -1_{1 \times n}\\
        0_{d\times n}\\
        -1_{1 \times n}\\
        I_n
    \end{bmatrix}
    =
    \begin{bmatrix}
        W\\
        -1_{1 \times n}\\
        0_{d\times n}\\
        -1_{1 \times n}\\
        I_n
        \end{bmatrix}\in\R^{(2d+n+2)\times n}.
\end{align*}

We define $\li_x$ as:
\begin{align*}
    & ~ \li_x(X) \\
    := & ~
    \sum_{i=1}^n
    \underbrace{\begin{bmatrix}
        I_{d+1}\\
        0_{(d+1+n)\times (d+1)}
    \end{bmatrix}}_{(2d+n+2)\times (d+1)}
    \underbrace{X}_{(d+1)\times n}
    \underbrace{\begin{bmatrix}
        0_{n\times (i-1)(P+1)} & 2L_0\onehot{n}{i} & 2L_1\onehot{n}{i} & \cdots & 2L_P\onehot{n}{i} & 0_{n\times(n-i)(P+1)}
    \end{bmatrix}}_{n \times n(P+1)} \\
    & ~ +
    \sum_{i=1}^n
    \underbrace{\begin{bmatrix}
        0_{(d+1)\times d} & 0_{(d+1)}\\
        I_d & 0_d\\
        0_{(n+1)\times d} & 0_{n+1}
    \end{bmatrix}}_{(2d+n+2)\times (d+1)}
     X
     \begin{bmatrix}
        0_{n\times (i-1)(P+1)} & f(L_0)\onehot{i}{i} & f(L_1)\onehot{i}{i} & \cdots & f(L_P)\onehot{i}{i} & 0_{n\times(n-i)(P+1)}
    \end{bmatrix}
     \\
     & ~ +
     \underbrace{\begin{bmatrix}
     0_{(2d+1)\times (P+1)} & \cdots &0_{(2d+1)\times (P+1)} \\
         S & \cdots & S \\
         ({2dB^2+B-\ln\epsilon_0})\onehot{n}{1}1_{1\times (P+1)}
         &
         \cdots &
         ({2dB^2+B-\ln\epsilon_0})\onehot{n}{n}1_{1\times (P+1)}
     \end{bmatrix}}_{(2d+n+2)\times n(P+1)}\\
     = & ~
     \begin{bmatrix}
         T_1 & T_2 & \cdots & T_n
     \end{bmatrix},
\end{align*}
where
\begin{align*}
    & ~1_{1\times (P+1)} := \begin{bmatrix}
        1 & 1 & \cdots & 1
    \end{bmatrix} \in\R^{1\times (P+1)}, \\
    & ~S :=
    \begin{bmatrix}
        -L_0^2 & -L_1^2 & \cdots & L_P^2
    \end{bmatrix}\in\R^{1\times (P+1)},\\
    & ~T_i :=
    \begin{bmatrix}
        2L_0x_i & 2L_1x_i & \cdots & 2L_Px_i \\
        2L_0y_i & 2L_1y_i & \cdots & 2L_Py_i \\
        f(L_0)x_i & f(L_1)x_i & \cdots & f(L_P)x_i\\
        -L_0^2 & -L_1^2 & \cdots & -L_P^2 \\
        ({2dB^2+B-\ln\epsilon_0})\onehot{n}{i} & ({2dB^2+B-\ln\epsilon_0})\onehot{n}{i} & \cdots & ({2dB^2+B-\ln\epsilon_0})\onehot{n}{i}
    \end{bmatrix}\in\R^{(2d+n+2)\times (P+1)}.
\end{align*}

Here $\epsilon_0$ is a parameter that we will designate later according to $\epsilon$.

Now construct $W_K,W_Q,W_V$ to be:
\begin{align*}
W_Q := & ~  \begin{bmatrix}
    0_{n(P+1)\times n} \\
    I_{n}
\end{bmatrix} \in \R^{(n(P+1)+n)\times n}, \\
W_K := & ~ \begin{bmatrix}
    I_{n(P+1)} \\
    0_{n\times n(P+1)}
\end{bmatrix} \in\R^{(n(P+1)+n)\times n(P+1)},\\
W_V^\top:= & ~
\begin{bmatrix}
    0_{d \times (d+1)} & I_d & 0_{d \times (n+1)}\\
\end{bmatrix}\in\R^{d \times (2d+n+2)}.
\end{align*}

Therefore,
\begin{align*}
    \Attn_s \circ \li(Z)
    = & ~
    W_V\li(Z)
    \Softmax( (
    \li(Z)W_K)^\top
    \li(Z) W_Q
    ), \\
\end{align*}
where
\begin{align*}
\Softmax((
    \li(Z)W_K)^\top
    \li(Z) W_Q
    )
= & ~
\Softmax
(
\begin{bmatrix}
         T_1 & T_2 & \cdots & T_n
\end{bmatrix}^\top
\begin{bmatrix}
        W\\
        -1_{1 \times n}\\
        0_{d\times n}\\
        -1_{1 \times n}\\
        I_n
    \end{bmatrix}
).
\end{align*}

This is equivalent to:
\begin{align*}
((\li(Z)W_K)^\top
    \li(Z) W_Q
)_{:,c}
= & ~
\begin{bmatrix}
    T_1^\top \\
    T_2^\top\\
    \vdots \\
    T_n^\top
\end{bmatrix}
\cdot
\begin{bmatrix}
w \\
-1\\
0_d\\
-1\\
\onehot{n}{c}
\end{bmatrix}\\
= & ~
\begin{bmatrix}
    M_{1,c} \\
    M_{2,c} \\
    \vdots \\
    M_{n,c}
\end{bmatrix},
\end{align*}
where
\begin{align*}
    M_{i,c} := & ~
    T_i^\top
    \cdot
    \begin{bmatrix}
    w\\
    -1\\
    0_d\\
    -1\\
    \onehot{n}{i}
    \end{bmatrix}\\
    = & ~
    \begin{bmatrix}
        2L_0x_i^\top w-2L_0y_i-L_0^2+ ({2dB^2+B-\ln\epsilon_0}) \one_{i=c} \\
        2L_1x_i^\top w-2L_1y_i-L_1^2+ ({2dB^2+B-\ln\epsilon_0}) \one_{i=c}\\
        \cdots \\
        2L_Px_i^\top w-2L_Py_i-L_P^2+ ({2dB^2+B-\ln\epsilon_0}) \one_{i=c}
    \end{bmatrix},
\end{align*}
where $i\in[n]$ and $c\in[n]$, and $\one_{i=c}$ represents the indicator function of $i=c$.

This means that
\begin{align*}
    & ~ \Softmax(
    (
    \li(Z)W_K)^\top
    \li(Z) W_Q
    )_{:,c} \\
    = & ~
    \Softmax
    (
    \begin{bmatrix}
    M_{1,c} \\
    M_{2,c} \\
    \vdots \\
    M_{n,c}
    \end{bmatrix}
    ) \\
    = & ~
    \sum_{i=1}^n\sum_{j=1}^P
    \frac{\exp{
    2L_{j}x_i^\top w-2L_{j}y_i-L_0^2+ ({2dB^2+B-\ln\epsilon_0}) \one_{i=c}}
    }{\sum_{i'=1}^n\sum_{j'=0}^P
    \exp{2L_{j'}x_{i'}^\top w-2L_{j'}y_{i'}-L_{j'}^2+ ({2dB^2+B-\ln\epsilon_0}) \one_{i=c}}
    }\onehot{nP}{(i-1)P+j}.
\end{align*}

Thus we have
\begin{align*}
& ~ \Attn_s \circ \li(Z)_{:,c}\\
= & ~
W_V\li(Z)
    \Softmax(
(\li(Z)W_K)^\top
    \li(Z) W_Q
)_{:,c} \annot{$W_v$ only retrieves the $(d+2)$-th row in $T_i$}\\
= & ~
\begin{bmatrix}
    F_1 &\cdots & F_n
\end{bmatrix}
\sum_{i=1}^n\sum_{j=1}^P
    \frac{\exp{
    2L_{j}x_i^\top w-2L_{j}y_i-L_j^2+ ({2dB^2+B-\ln\epsilon_0}) \one_{i=c}}
    }{\sum_{i'=1}^n\sum_{j'=0}^P
    \exp{2L_{j'}x_{i'}^\top w-2L_{j'}y_{i'}-L_{j'}^2+ ({2dB^2+B-\ln\epsilon_0}) \one_{i=c}}
    }\onehot{nP}{(i-1)P+j}
\\
= & ~
\sum_{i=1}^n\sum_{j=0}^P
    \frac{\exp{
    2L_{j}x_i^\top w-2L_{j}y_i-L_j^2+ ({2dB^2+B-\ln\epsilon_0}) \one_{i=c}}
    }{\sum_{i'=1}^n\sum_{j'=0}^P
    \exp{2L_{j'}x_{i'}^\top w-2L_{j'}y_{i'}-L_{j'}^2+ ({2dB^2+B-\ln\epsilon_0}) \one_{i=c}}
    }f(L_j)x_i,
\end{align*}
where $F$ is:
\begin{align*}
F_i
:=
\begin{bmatrix}
f(L_0)x_i & f(L_1)x_i & \cdots & f(L_P)x_i
\end{bmatrix}.
\end{align*}

For every $i \in [n]$, if $i\neq c$, we have
\begin{align*}
    & ~ \sum_{j=0}^P
    \frac{\exp{
    2L_{j}x_i^\top w-2L_{j}y_i-L_j^2+ ({2dB^2+B-\ln\epsilon_0}) \one_{i=c}}
    }{\sum_{i'=1}^n\sum_{j'=0}^P
    \exp{2L_{j'}x_{i'}^\top w-2L_{j'}y_{i'}-L_{j'}^2+ ({2dB^2+B-\ln\epsilon_0}) \one_{i=c}}
    } \\
    = & ~
    \sum_{j=0}^P
    \frac{\exp{
    2L_{j}x_i^\top w-2L_{j}y_i-L_j^2
    }}{\sum_{i'=1}^n\sum_{j'=0}^P
    \exp{2L_{j'}x_{i'}^\top w-2L_{j'}y_{i'}-L_{j'}^2+ ({2dB^2+B-\ln\epsilon_0}) \one_{i=c}}
    }\\
    < & ~
    \sum_{j=0}^P
    \frac{\exp{
    2L_{j}x_i^\top w-2L_{j}y_i-L_j^2
    }}{\sum_{j'=0}^P
    \exp{2L_{j'}x_{i'}^\top w-2L_{j'}y_{i'}-L_{j'}^2+ ({2dB^2+B-\ln\epsilon_0})}}
    \annot{only taking the $i' = c$ part}
    \\
    < & ~
    \sum_{j=0}^P
    \frac{\exp{
    2dB^2+B
    }}{P
    \exp({2dB^2+B-\ln\epsilon_0})}
    =
    \epsilon_0.
\end{align*}

For $i = c$, since
\begin{align*}
    \sum_{i\neq c}^n\sum_{j=0}^P
    \frac{\exp{
    2L_{j}x_i^\top w-2L_{j}y_i-L_j^2+ ({2dB^2+B-\ln\epsilon_0}) \one_{i=c}}
    }{\sum_{i'=1}^n\sum_{j'=0}^P
    \exp{2L_{j'}x_{i'}^\top w-2L_{j'}y_{i'}-L_{j'}^2+ ({2dB^2+B-\ln\epsilon_0}) \one_{i=c}}
    }\leq (n-1)\epsilon_0,
\end{align*}

we have
\begin{align*}
    & ~
    \frac{\sum_{j=0}^P\exp{
    2L_{j}x_c^\top w-2L_{j}y_c-L_j^2+ ({2dB^2+B-\ln\epsilon_0})}
    }{\sum_{i'=1}^n\sum_{j'=0}^P
    \exp{2L_{j'}x_{i'}^\top w-2L_{j'}y_{i'}-L_{j'}^2+ ({2dB^2+B-\ln\epsilon_0}) \one_{i=c}}
    }\\
    = & ~
    \sum_{j=0}^P
    \frac{\exp{
    2L_{j}x_c^\top w-2L_{j}y_c-L_j^2+ ({2dB^2+B-\ln\epsilon_0})}
    }{\sum_{i'=1}^n\sum_{j'=0}^P
    \exp{2L_{j'}x_{i'}^\top w-2L_{j'}y_{i'}-L_{j'}^2+ ({2dB^2+B-\ln\epsilon_0}) \one_{i=c}}
    }\\
    \geq & ~  1-(n-1)\epsilon_0.
\end{align*}

Thus for the parts in the weighted sum output that corresponds to rows in $M_{:,c}$ in the attention score matrix, we have
\begin{align*}
    & ~ \|
    \sum_{j=0}^P
    \frac{\exp{
    2L_{j}x_c^\top w-2L_{j}y_c-L_j^2+ ({2dB^2+B-\ln\epsilon_0})}
    }{\sum_{i'=1}^n\sum_{j'=0}^P
    \exp{2L_{j'}x_{i'}^\top w-2L_{j'}y_{i'}-L_{j'}^2+ ({2dB^2+B-\ln\epsilon_0}) \one_{i=c}}
    }f(L_j)x_c
    -f(x_c^\top w-y_c)x_c\|_\infty \\
    = & ~ %
    \|
    \sum_{j=0}^P
    \frac{\exp{
    2L_{j}x_c^\top w-2L_{j}y_c-L_j^2+ ({2dB^2+B-\ln\epsilon_0})}
    }{\sum_{j'=0}^P\exp{
    2L_{j'}x_c^\top w-2L_{j'}y_c-L_{j'}^2+ ({2dB^2+B-\ln\epsilon_0})}}(f(L_j)x_c-f(x_c^\top w-y_c)x_c) \\
    & ~ \quad\quad \cdot
    \frac{\sum_{j'=0}^P\exp{
    2L_{j'}x_c^\top w-2L_{j'}y_c-L_{j'}^2+ ({2dB^2+B-\ln\epsilon_0})}}
    {\sum_{i'=1}^n\sum_{k=0}^P
    \exp{2L_{k}x_{i'}^\top w-2L_{k}y_{i'}-L_{k}^2+ ({2dB^2+B-\ln\epsilon_0}) \one_{i=c}}
    }\\
    & ~ -
    (1-\frac{\sum_{j'=0}^P\exp{
    2L_{j'}x_c^\top w-2L_{j'}y_c-L_{j'}^2+ ({2dB^2+B-\ln\epsilon_0})}}
    {\sum_{i'=1}^n\sum_{k=0}^P
    \exp{2L_{k}x_{i'}^\top w-2L_{k}y_{i'}-L_{k}^2+ ({2dB^2+B-\ln\epsilon_0}) \one_{i=c}}
    } )
    f(x_c^\top w-y_c)x_c
    \|_\infty\\
    \leq & ~ %
    \sum_{j=0}^P
    \frac{\exp{
    2L_{j}x_c^\top w-2L_{j}y_c-L_j^2+ ({2dB^2+B-\ln\epsilon_0})}
    }{\sum_{j'=0}^P\exp{
    2L_{j'}x_c^\top w-2L_{j'}y_c-L_{j'}^2+ ({2dB^2+B-\ln\epsilon_0})}}|f(L_j)-f(x_c^\top w-y_c)|\cdot d\|x_c\|_\infty
    \\
    & ~ -
    (1-\frac{\sum_{j'=0}^P\exp{
    2L_{j'}x_c^\top w-2L_{j'}y_c-L_{j'}^2+ ({2dB^2+B-\ln\epsilon_0})}}
    {\sum_{i'=1}^n\sum_{k=0}^P
    \exp{2L_{k}x_{i'}^\top w-2L_{k}y_{i'}-L_{k}^2+ ({2dB^2+B-\ln\epsilon_0}) \one_{i=c}}
    })|f(x_c^\top w-y_c)|\|x_c\|_\infty\\
    \leq & ~ %
    \sum_{j=0}^P
    \frac{\exp{
    2L_{j}x_c^\top w-2L_{j}y_c-L_j^2+ ({2dB^2+B-\ln\epsilon_0})}
    }{\sum_{j'=0}^P\exp{
    2L_{j'}x_c^\top w-2L_{j'}y_c-L_{j'}^2+ ({2dB^2+B-\ln\epsilon_0})}}|f(x_c^\top w-y_c)|\|x_c\|_\infty \\
    & ~+
    (n-1)\epsilon_0 B_f\|x_c\|_\infty\\
    = & ~ %
    \sum_{j=0}^P
    \frac{\exp{
    2L_{j}x_c^\top w-2L_{j}y_c-L_j^2}
    }{\sum_{j'=0}^P\exp{
    2L_{j'}x_c^\top w-2L_{j'}y_c-L_{j'}^2}}|f(L_j)-f(x_c^\top w-y_c)|\|x_c\|_\infty
    +
    (n-1)\epsilon_0 B_f\|x_c\|_\infty\\
    =& ~
    \sum_{j=0}^P
    \frac{\exp{
    -\beta(x_c^\top w -y_c -L_j)^2}
    }{\sum_{j'=0}^P\exp{
    -\beta(x_c^\top w -y_c -L_{j'})^2}}|f(L_j)-f(x_c^\top w-y_c)|\|x_c\|_\infty+
    (n-1)\epsilon_0 B_f\|x_c\|_\infty,
\end{align*}
where we define $B_f:=|f|$ as the bound for $f$.

For any $\epsilon_1>0$, set $\Delta L$ to be sufficiently small such that
\begin{align*}
    |f(x)-f(y)|\leq \epsilon_1,
\end{align*}
when $|x-y|\leq \Delta L$.

Then when $\beta$ is sufficiently large, we have
\begin{align*}
    \sum_{|L_i - (x_c^\top w -y_c)|>\Delta L}
    \frac{\exp{
    -\beta(x_c^\top w -y_c -L_j)^2}
    }{\sum_{j'=0}^P\exp{
    -\beta(x_c^\top w -y_c -L_{j'})^2}}\leq\epsilon_2,
\end{align*}
for any $\epsilon_2>0$.

Thus
\begin{align*}
    & ~ \sum_{j=0}^P
    \frac{\exp{
    -\beta(x_c^\top w -y_c -L_j)^2}
    }{\sum_{j'=0}^P\exp{
    -\beta(x_c^\top w -y_c -L_{j'})^2}}|f(L_j)-f(x_c^\top w-y_c)| \\
    = & ~
    \sum_{|L_i - (x_c^\top w -y_c)|>\Delta L}
    \frac{\exp{
    -\beta(x_c^\top w -y_c -L_j)^2}
    }{\sum_{j'=0}^P\exp{
    -\beta(x_c^\top w -y_c -L_{j'})^2}}|f(L_j)-f(x_c^\top w-y_c)| \\
    & ~ +
    \sum_{|L_i - (x_c^\top w -y_c)|\leq\Delta L}
    \frac{\exp{
    -\beta(x_c^\top w -y_c -L_j)^2}
    }{\sum_{j'=0}^P\exp{
    -\beta(x_c^\top w -y_c -L_{j'})^2}}|f(L_j)-f(x_c^\top w-y_c)| \\
    \leq & ~
    \epsilon_2\cdot 2B_f
    +
    \epsilon_1.
\end{align*}
This completes the proof.
\end{proof}

\subsection{Proof of \texorpdfstring{\cref{cor:in-context-sim-gd}}{}}
\label{proof:cor:in-context-sim-gd}

\begin{corollary}[In-Context Emulation of a Single GD Step; \cref{cor:in-context-sim-gd} Restate]
    Let $\ell:\R\to\R$ be differentiable and define $\hat {L}_n(w) := \frac{1}{n}\sum_{i=1}^n \ell(w^\top x_i-y_i)$.
    For any step size $\eta>0$ and any $\epsilon>0$, there exist a single-head attention $\Attn_s$ and a linear map $\li$ such that,
    with $Z=[X;W]$ as in \eqref{eqn:prompt},
    choosing the readout $u:=\tfrac1n\mathbf{1}_n$ \text{(equivalently, right-multiply by $W_O=u$ in \cref{def:attn})}, we have
    \begin{align*}
    \hat{w}_{\rm GD} := (\Attn_s\circ\li(Z))u \in\R^d
    \quad\text{and}\quad
    \|\hat w_{\rm GD} - \underbrace{(w-\eta\nabla \hat L_n(w))}_{w^+_{\rm GD}} \|_\infty \le \epsilon .
    \end{align*}
\end{corollary}

\begin{proof}
    From  \cref{cor:in-context-GD-attn}, we derive that $\|(\Attn_s\circ\li)_i-\nabla \ell(w^\top x_i-y_i) x_i\|_\infty\leq\epsilon$ for all $i\in[n]$.

    Therefore,
    \begin{align*}
        \hat{w} &= w+\frac{1}{n}\sum_{j=1}^n(\Attn_s\circ\li)_j  \\
        &= w-\frac{\eta}{n}\sum_{j=1}^n\nabla \ell(w^\top x_j-y_j) x_j + \epsilon^\prime \\
        &= w- \eta\nabla\hat{L_n}(w)+ \epsilon^\prime \\
        &= w_{\text{GD}}+ \epsilon^\prime
    \end{align*}

This completes the proof.
\end{proof}

\subsection{Proof of \texorpdfstring{\cref{thm:attn_sim_lr}}{}}
\label{proof:thm:attn_sim_lr}

\begin{theorem}[In-Context Emulation of Linear Regression;  \cref{thm:attn_sim_lr} Restate]
    For any dataset $\{(x_i,y_i)\}_{i=1}^n$ with $x_i\in\R^d$, $y_i\in\R$ and any $\epsilon>0$, there exist a
    single-head attention $\Attn_s$, a linear map $\li$, and a readout $u\in\R^n$ such that, with $Z=[X;W]$ as in
    \eqref{eqn:prompt} (for any fixed bounded $w$),
    \begin{align*}
    \hat w_{\text{linear}} := (\Attn_s\circ\li(Z))u \in \R^d,
    \quad\text{and}\quad
    \|\hat w_{\text{linear}} - w_{\text{linear}}\|_\infty \le \epsilon ,
    \end{align*}
    where $w_{\text{linear}} := \argmin_{w\in\R^d}\frac{1}{2n}\sum_{i=1}^n(\langle w, x_i\rangle-y_i)^2$.
\end{theorem}

\begin{proof}
    From \cref{cor:in-context-sim-gd}, we know that $\|\hat{w}^l- w_{\gd}^l\|_\infty\leq\epsilon/2$ for all $l\in[L]$. Note that $\frac{1}{2n}\sum_{i=1}^n(\langle w, x_i\rangle-y_i)^2$ is convex and smooth which satisfies the precondition for \cref{cor:gd-smooth-strongly-convex}. Therefore, from \cref{cor:gd-smooth-strongly-convex}, using $\|\cdot\|_\infty\leq\|\cdot\|_2$, we derive that $\|w_{\text{GD}}^l-w_{\text{linear}}^l\|_\infty\leq\epsilon/2$. Thus, $\|\Attn - w_{\text{linear}}\|_\infty\leq\|\hat{w}^l- w_{\gd}^l\|_\infty+\|w_{\text{GD}}^l-w_{\text{linear}}^l\|_\infty\leq\epsilon$ by triangle inequality. This completes the proof.
\end{proof}

\subsection{Proof of \texorpdfstring{\cref{thm:attn_sim_rr}}{}}
\label{proof:thm:attn_sim_rr}

\begin{theorem}[Restate of \cref{thm:attn_sim_rr}: In-Context Emulation of Ridge Regression]
    For any input-output pair $(x_i , y_i)$, where $x_i\in\R^d, y_i\in\R, i\in[n]$, and any $\epsilon>0$, there exists a single-layer Attention network with linear connection $\Attn$ such that
    \begin{align*}
        \|\Attn - w_{\text{ridge}}\|_\infty \leq \epsilon,
    \end{align*}
    where $w_{\text{ridge}} := \argmin_{w\in\R^d}\frac{1}{2n}\sum_{i=1}^n(\langle w, x_i\rangle-y_i)^2+\frac{\lambda}{2}\|w\|_2^2$ with regularization term $\lambda\geq0$. This completes the proof.
\end{theorem}

\begin{proof}
    From \cref{cor:in-context-sim-gd}, we know that $\|\hat{w}^l- w_{\gd}^l\|_\infty\leq\epsilon/2$ for all $l\in[L]$. Note that $\frac{1}{2n}\sum_{i=1}^n(\langle w, x_i\rangle-y_i)^2+\frac{\lambda}{2}\|w\|_2^2$ is convex and smooth which satisfies the precondition for \cref{cor:gd-smooth-strongly-convex}. Therefore, from \cref{cor:gd-smooth-strongly-convex}, using $\|\cdot\|_\infty\leq\|\cdot\|_2$, we derive that $\|w_{\text{GD}}^l-w_{\text{ridge}}^l\|_\infty\leq\epsilon/2$. Thus, $\|\Attn - w_{\text{ridge}}\|_\infty\leq\|\hat{w}^l- w_{\gd}^l\|_\infty+\|w_{\text{GD}}^l-w_{\text{ridge}}^l\|_\infty\leq\epsilon$ by triangle inequality. This completes the proof.
\end{proof}

\subsection{Proof of \texorpdfstring{\cref{thm:attn_sim_attn_multi}}{}}
\label{proof:thm:attn_sim_attn_multi}

\begin{theorem}[In-Context Emulation of Attention;  \cref{thm:attn_sim_attn_multi} Restate]
Let $X \in \mathbb{R}^{d\times n}$ be an input sequence, and let $W^K \in \mathbb{R}^{d_h\times d}$, $W^Q \in \mathbb{R}^{d_h\times d}$, $W^V \in \mathbb{R}^{d\times d}$ be the weight matrices of the target attention head we wish to emulate in-context.
For any $\epsilon > 0$, there exists a two-layer attention network --- a multi-head attention layer $\Attn_m$ followed by a single-head attention layer $\Attn_s$ --- such that
\begin{align*}
\|
\Attn_s \circ \Attn_m (
X_p
)
-
W_VX\Softmax((W_KX)^\top W_QX)
\|_\infty
\leq \epsilon,
\quad\text{for any}\quad
\epsilon>0,
\end{align*}
where $X_p$ is the prompt defined in \cref{def:input_attn}.
\end{theorem}

\begin{proof}

For the input, we first append it with the positional encoding $I_n$.
This result is denoted as $X_p$. It writes out as
\begin{align*}
X_p :=
\begin{bmatrix}
    X\\
    W\\
    I_n
\end{bmatrix} \in\R^{(d+6dd_h+n)\times n}.
\end{align*}

Denote $a$ and $b$ to be the minimum and maximum values that the inner product $w\top x$ can take.

We separate the heads in $\Attn_m$ into three groups
\begin{align*}
& ~ \Attn_j^K := \Attn_j,\quad  j\in [N],\annot{Responsible for calculating $K$}\\
& ~ \Attn_j^Q := \Attn_{N+j},\quad  j\in [N],\annot{Responsible for calculating $Q$}\\
& ~ \Attn_j^V := \Attn_{2N+j},\quad  j\in [N]\annot{Responsible for calculating $V$},
\end{align*}
where $N$ is defined as
\begin{align*}
    N := \lceil \frac{2(b-a) }{(n-2)\epsilon_0}\rceil \cdot d_h.
\end{align*}

For simplicity, we define
\begin{align*}
    H := \lceil \frac{2(b-a) }{(n-2)\epsilon_0}\rceil.
\end{align*}
We prove our claim by adopting \cref{thm:multi-head-truncated_in-context} to construct heads to approximate in-context linear transformations ($k_j^\top x,q_j^\top x,v_j^\top x,j\in [d_h]$).
We now construct linear transformations $A_h$ attached before the attentions given by \cref{thm:multi-head-truncated_in-context}. By \cref{lem:still_attention}, when prepended with $A_h$, these attentions are still attentions.
Thus we are able to use them to construct the heads in $\Attn_m$.

Construct $A_h$ as
\begin{align*}
A_h
:=
\begin{bmatrix}
I_d & 0_{d\times 3d_h} & 0_{d\times 3d_h} & 0_{d\times n} \\
0_{d\times d} & E_h & K_hE_h & 0_{d\times n} \\
0_{n\times d} & 0_{n\times 3d_h} & 0_{n\times 3d_h} & I_n
\end{bmatrix}\in\R^{(2d+n)\times(d+6dd_h+n)},
\end{align*}
where
\begin{align*}
& ~ E_h
:=
\begin{bmatrix}
0_{d\times (d[h/H])} & I_d & 0_{d\times (3dd_h-d[h/H]-d)}
\end{bmatrix} \in\R^{d\times 3d_h},\\
& ~ K_h := [(h \% H-1)(n-2)-1].
\end{align*}
Here $h\%H$ is used to denote the remainder of dividing $h$ by $H$. We define $\%$ such that instead of the the common $(kH) \% H = 0$,
\begin{align*}
    kH \% H = H,
    \quad \text{for all } \quad
    k \in \mathbb{N}^+.
\end{align*}

Applying $A$ to the input yields
\begin{align*}
A_h\cdot
X_p
:= & ~
\begin{bmatrix}
I_d & 0_{d\times 3d_h} & 0_{d\times 3d_h} & 0_{d\times n} \\
0_{d\times d} & E_h & K_hE_h & 0_{d\times n} \\
0_{n\times d} & 0_{n\times 3d_h} & 0_{n\times 3d_h} & I_n
\end{bmatrix}
\cdot
\begin{bmatrix}
    X\\
    W\\
    I_n
\end{bmatrix}\\
= & ~
\begin{bmatrix}
    X \\
[E_h \quad K_hE_h] \cdot W\\
    I_n
\end{bmatrix}.
\end{align*}

In the above equation, $[E_h\quad K_hE_h] \cdot W$ is equivalent to
\begin{align*}
\begin{bmatrix}
    E_h & K_hE_h
\end{bmatrix}
W
= & ~
\begin{bmatrix}
    E_h & K_hE_h
\end{bmatrix}
\begin{bmatrix}
        0\cdot w & 1\cdot w & 2\cdot w & \cdots & (n-1)\cdot w\\
        w & w & w & \cdots & w
\end{bmatrix}\\
= & ~
E_h
\begin{bmatrix}
        0\cdot w & 1\cdot w & 2\cdot w & \cdots & (n-1)\cdot w\\
\end{bmatrix}
+
K_hE_h
\begin{bmatrix}
        w & w & w & \cdots & w
\end{bmatrix}\\
= & ~
\begin{bmatrix}
   K_hE_hw & (K_h +1)E_hw & \cdots & (K_h+n-1)E_hw
\end{bmatrix},
\end{align*}
where
\begin{align}
\label{eq:Ehw_output}
E_h w
= & ~
\begin{bmatrix}
0_{d\times (d[h/H])} & I_d & 0_{d\times (3dd_h-d[h/H]-d)}
\end{bmatrix}
\begin{bmatrix}
    k \\
    q \\
    v
\end{bmatrix}\notag\\
= & ~
\begin{cases}
k_{\lceil h/H\rceil}, & ~ 1\leq h\leq N\\
q_{\lceil h/H\rceil-N},&~ N < h \leq 2N\\
v_{\lceil h/H\rceil-2N},& ~ 2N\leq h \leq
3N\end{cases}.
\end{align}

When $1\leq h\leq N$, we compute $A_h\cdot X_p$ as
\begin{align*}
A_h\cdot X_p =
\begin{bmatrix}
    X\\
    k_{\lceil h/H\rceil}\cdot1_{1\times n} \\
    I_n
\end{bmatrix}.
\end{align*}
This means every $h$ in $\{jH+1,\cdots,jH+H\}$, $(j\in\{0,1,\cdots, d_h\})$ has the same $A_h\cdot X_p$, which is
\begin{align*}
\begin{bmatrix}
    X\\
    k_j\cdot1_{1\times n} \\
    I_n
\end{bmatrix}.
\end{align*}

By \cref{thm:multi-head-truncated_in-context}, there exists a multi-head attention $\Attn_j'$ of $H$ heads that takes the input of the shape of $A_h \cdot X_p $, such that the output satisfies
\begin{align*}
\| %
\Attn_j'(A_h\cdot X_p)
-
k_j^\top X\onehot{3N}{j}
\|_\infty
\leq
\epsilon_0,
\end{align*}
for any $\epsilon_0 >0$.

We use $\Attn_j'^{(s)}$ to denote the $s$-th head of $\Attn_j'$, and thus we define the $h$-th head for calculating $K$ to be
\begin{align*}
\Attn_h^K (Z) := \Attn_j'^{(h-jH)}(A_hZ) \annot{$Z\in\R^{(d+6dd_h+n)\times n}$ denotes any input}.
\end{align*}

By \cref{lem:still_attention}, $\Attn_h^K (Z)$ is still an attention.

Thus, we have
\begin{align*}
\| %
\Attn_j^K(X_p)
-
k_j^\top X\onehot{3N}{j}
\|_\infty
\leq
\epsilon_0.
\end{align*}

This means that
\begin{align}
\label{eq:K}
\| %
\sum_{j\in[N]}\Attn_j^K(X_p)
-
\begin{bmatrix}
        K \\
        0_{n\times n}\\
        0_{n\times n}
    \end{bmatrix}
\|_\infty
\leq
\epsilon_0.
\end{align}

Similarly for $Q$, by \eqref{eq:Ehw_output}, we have
\begin{align*}
    E_hw
    =
    q_j,
    \quad \text{where} \quad j = \lceil h/H\rceil - N,
\end{align*}
and
\begin{align*}
    A_h\cdot X_p =
\begin{bmatrix}
    X\\
    q_j\cdot1_{1\times n} \\
    I_n
\end{bmatrix}.
\end{align*}

By \cref{thm:multi-head-truncated_in-context}, there exists a multi-head attention $\Attn_j''$ for each $j\in\{0,1,\cdots d_h-1\}$ such that
\begin{align*}
    \| %
\Attn_j''(A_h\cdot X_p)
-
q_j^\top X\onehot{3N}{j+N}
\|_\infty
\leq
\epsilon_0.
\end{align*}

Then we construct $\Attn_j^Q$ in a way similar to $\Attn_j^K$
\begin{align*}
    \Attn_j^Q(Z) :=
    \Attn_j''(A_h\cdot Z).
\end{align*}
By \cref{lem:still_attention}, $\Attn_j^Q(Z)$ is an attention.

Thus, we have
\begin{align*}
\| %
\Attn_j^Q(X_p)
-
q_j^\top X\onehot{3N}{j+N}
\|_\infty
\leq
\epsilon_0.
\end{align*}

This means that
\begin{align}
\| %
\label{eq:Q}
\sum_{j\in[N]]}\Attn_j^Q(X_p)
-
\begin{bmatrix}
        0_{n\times n}\\
        Q\\
        0_{n\times n}
    \end{bmatrix}
\|_\infty
\leq
\epsilon_0.
\end{align}

Finally, for $V$, with completely analogous construction to that of $K$ and $Q$, there exists a construction of $\Attn^V_j,~j\in \{0,1,\cdots, d_h-1\} $ such that
\begin{align*}
\| %
\Attn_j^V(X_p)
-
v_j^\top X\onehot{3N}{j+2N}
\|_\infty
\leq
\epsilon_0.
\end{align*}

This means that
\begin{align}
\label{eq:V}
\| %
\sum_{j\in[N]}\Attn_j^V(X_p)
-
\begin{bmatrix}
        0_{n\times n}\\
        0_{n\times n}\\
        V
    \end{bmatrix}
\|_\infty
\leq
\epsilon_0.
\end{align}

Combining \eqref{eq:K}, \eqref{eq:Q} and \eqref{eq:V} we have
\begin{align*}
\|
\sum_{j\in [3N]}\Attn_j(X_p) -
\begin{bmatrix}
    K \\
    Q\\
    V
\end{bmatrix}
\|_\infty \leq \epsilon_0.
\end{align*}
\paragraph{In-Context Calculation of Attention.}
Construct $\Attn_s$ to be
\begin{align*}
    \Attn_s(Z)
    :=
    \underbrace{
    \begin{bmatrix}
        0_{n\times 2n} & I_n
    \end{bmatrix}}_{W_V}
    Z
    \Softmax(
    (
    \underbrace{\begin{bmatrix}
        I_n & 0_{n\times 2n}
    \end{bmatrix}}_{W_K}
    Z)^\top
    \underbrace{
    \begin{bmatrix}
        0_{n\times n} & I_n & 0_{n\times n}
    \end{bmatrix}
    }_{W_Q}
    Z
    ).
\end{align*}

We have
\begin{align*}
    \Attn_s\left(
    \begin{bmatrix}
        K\\
        Q\\
        V
    \end{bmatrix}
    \right)
    =
    \underbrace{W_VX}_{d_h\times n}\Softmax
    \underbrace{\left(
    (W_KX)^\top W_QX
    \right) }_{n\times n}.
\end{align*}

Now we estimate the error due to the difference between the output of $\sum_{j=1}^{3n}
\Attn_j\circ\li_j\left(
\begin{bmatrix}
    X^\top&
    W_K&
    W_Q&
    W_V
\end{bmatrix}^\top
\right)$ and
$
\begin{bmatrix}
    K^\top & Q^\top & V^\top
\end{bmatrix}^\top$.

Define
\begin{align*}
\begin{bmatrix}
    K' \\
    Q' \\
    V'
\end{bmatrix}
:=
    \sum_{j=1}^{3n}
\Attn_j\circ\li_j\left(
    \begin{bmatrix}
        X \\
        W_K^\top\\
        W_Q^\top\\
        W_V^\top
    \end{bmatrix}
    \right).
\end{align*}
This is the same as
\begin{align*}
    \sum_{j=1}^n \Attn^K_j\circ\li^K_j\left(
    \begin{bmatrix}
        X \\
        W_K^\top\\
        W_Q^\top\\
        W_V^\top
    \end{bmatrix}
    \right)
    =
    \begin{bmatrix}
        K' \\
        0_{n\times n}\\
        0_{n\times n}
    \end{bmatrix}, \\
    \sum_{j=1}^n \Attn^Q_j\circ\li^Q_j\left(
    \begin{bmatrix}
        X \\
        W_K^\top\\
        W_Q^\top\\
        W_V^\top
    \end{bmatrix}
    \right)
    =
    \begin{bmatrix}
        0_{n\times n}\\
        Q' \\
        0_{n\times n}
    \end{bmatrix},\\
    \sum_{j=1}^n \Attn^V_j\circ\li^V_j\left(
    \begin{bmatrix}
        X \\
        W_K^\top\\
        W_Q^\top\\
        W_V^\top
    \end{bmatrix}
    \right)
    =
    \begin{bmatrix}
        0_{n\times n}\\
        0_{n\times n}\\
        V'
    \end{bmatrix}.
\end{align*}

In this definition, we have
\begin{align*}
    & ~ \Attn_s\left(
    \begin{bmatrix}
        K'\\
        Q'\\
        V'
    \end{bmatrix}
    \right)
    -
    \Attn_s\left(
    \begin{bmatrix}
        K\\
        Q\\
        V
    \end{bmatrix}
    \right)\\
    = & ~
    V'\Softmax
    \left(
    K'^\top Q'
    \right)
    -
    V\Softmax
    \left(
    K^\top Q
    \right) \\
    = & ~
    (V'-V)\Softmax
    \left(
    K'^\top Q'
    \right)+
    V(\Softmax
    \left(
    K'^\top Q'
    \right)-
    \Softmax
    \left(
    K^\top Q
    \right)).
\end{align*}

This yields that
\begin{align}
& ~ \left\|
    \Attn_s\left(
    \begin{bmatrix}
        K'\\
        Q'\\
        V'
    \end{bmatrix}
    \right)
    -
    \Attn_s\left(
    \begin{bmatrix}
        K\\
        Q\\
        V
    \end{bmatrix}
    \right)
    \right\|_\infty \notag\\
        = & ~
        \left\|
        (V'-V)\Softmax
        \left(
        K'^\top Q'
        \right)+
        V(\Softmax
        \left(
        K'^\top Q'
        \right)-
        \Softmax
        \left(
        K^\top Q
        \right))
        \right\|_\infty \notag\\
    \leq & ~
    \left\|
    (V'-V)\Softmax
    \left(
    K'^\top Q'
    \right)\right\|_\infty
    +
    \left\|
    V(\Softmax
    \left(
    K'^\top Q'
    \right)-
    \Softmax
    \left(
    K^\top Q
    \right))
    \right\|_\infty \annot{Triangle inequality}\notag\\
        \leq & ~ %
        \label{eq:dif_ats'-ats}
        \epsilon_0
        +
        n\left\|
        V\right\|_\infty
        \left\|\Softmax
        \left(
        K'^\top Q'
        \right)-
        \Softmax
        \left(
        K^\top Q
        \right)
        \right\|_\infty.
\end{align}
The last inequality is because that $\Softmax(K'^\top Q')$ has each column summing up to $1$, which means
\begin{align*}
(V'-V)\Softmax
\left(
K'^\top Q'
\right)_{:,j}  ,
\end{align*}
is a weighted sum of the columns in $V'-V$.

Thus we have
\begin{align*}
\|(V'-V)\Softmax
\left(
K'^\top Q'
\right)_{:,j} \|_\infty
\leq & ~
\|V-V'\|_\infty\leq \epsilon_0.
\end{align*}
Because the average of columns has a smaller maximal entry than the original columns.

Since $X,W_K,W_Q,W_V$ are all bounded in infinite norm, there multiplications are also bounded in infinite norm. We denote this bound as $B_{KQV}$.

Now we calculate the error of the last term in the above inequality
\begin{align*}
    & ~ |(\Softmax
    \left(
    K'^\top Q'
    \right)-
    \Softmax
    \left(
    K^\top Q
    \right))_{i,j}| \\
    = & ~
    |\frac{e^{K_i'\cdot Q_j'}}{\sum_{i'=1}^ne^{K_{i'}' \cdot Q_j'}}
    -
    \frac{e^{K_{i} \cdot Q_j}}{\sum_{i'=1}^ne^{K_{i'}\cdot Q_j}}|\\
    = & ~
    |\frac{e^{K_i'\cdot Q_j'}-e^{K_{i} \cdot Q_j}}{\sum_{i'=1}^ne^{K_{i'}' \cdot Q_j'}}|
    +
    |e^{K_{i} \cdot Q_j}(\frac{1}{\sum_{i'=1}^ne^{K_{i'}'\cdot Q_j'}}-\frac{1}{\sum_{i'=1}^ne^{K_{i'}\cdot Q_j}})|\\
    \leq & ~
    \frac{e^{K_i'\cdot Q_j'}}{\sum_{i'=1}^ne^{K_{i'}' \cdot Q_j'}}|1-e^{K_i'\cdot Q_j'-K_i\cdot Q_j}|+e^{K_iQ_j}\frac{1}{(\sum_{i'=1}^ne^{K_{i'}' \cdot Q_j'})(\sum_{i'=1}^ne^{K_{i'}\cdot Q_j})}|\sum_{i'=1}^ne^{K_{i'}\cdot Q_j}-\sum_{i'=1}^ne^{K_{i'}'\cdot Q_j'}| \\
    \leq & ~
    |1 - e^{\epsilon_0(K_i'+Q_j')+\epsilon_0^2}|+|1-\frac{\sum_{i'=1}^ne^{K_{i'}'\cdot Q_j'}}{\sum_{i'=1}^ne^{K_{i'}\cdot Q_j}}| \\
    \leq & ~
   2|1 - e^{\epsilon_0(K_i'+Q_j')+\epsilon_0^2}| \annot{the largest deviation ratio between $e^{K_{i'}\cdot Q_j}$ and $e^{K_{i'}'\cdot Q_j'}$ are invariant of $i$}\\
   \leq & ~
   2|1-e^{2\epsilon_0(B_{KQV}+\frac{1}{2})}|
,
\end{align*}
where $K_i'$, $Q_i'$ denote the columns in $K'$ and $Q'$ respectively.

Thus for any $\epsilon_1>0$, when $\epsilon_0$ satisfies
\begin{align*}
    \epsilon_0 \leq & ~ \frac{\ln(1+\frac{\epsilon_1}{2})}{2B_{KQV}+1},
\end{align*}
we have
\begin{align*}
|(\Softmax
    \left(
    K'^\top Q'
    \right)-
    \Softmax
    \left(
    K^\top Q
    \right))_{i,j}|
    \leq \epsilon_1.
\end{align*}

Combining this with \eqref{eq:dif_ats'-ats} yields
\begin{align*}
\left\|
    \Attn_s\left(
    \begin{bmatrix}
        K'\\
        Q'\\
        V'
    \end{bmatrix}
    \right)
    -
    \Attn_s\left(
    \begin{bmatrix}
        K\\
        Q\\
        V
    \end{bmatrix}
    \right)
    \right\|_\infty
    \leq & ~
        \epsilon_0
        +
        nB_{KQV}\epsilon_1
\end{align*}

When take $\epsilon_1$ and $\epsilon_0$ to be infinitely small, this right hand side tends to $0$.

This completes the proof.
\end{proof}

\subsection{Proof of \texorpdfstring{\cref{thm:attn_sim_attn}}{}}
\label{subsec:Simulation of Attention: Another Construction}

\begin{theorem}[Restate of \cref{thm:attn_sim_attn}]
\label{proof:thm:attn_sim_attn}
    Let $X \in \mathbb{R}^{d\times n}$ be an input sequence, and let $W^K, W^Q, W^V \in \mathbb{R}^{n\times d}$ be the weight matrices of the target attention (assumed to have bounded entries).
    For any $\epsilon > 0$, there exists a single-head attention layer $\Attn_s$ followed by a multi-head attention layer with linear projections such that
    \begin{align*}
    \|
    \sum_{j=1}^{3n} \Attn_s \circ \Attn_j\circ\li_j
    \left(\begin{bmatrix}
        X \\
        W_K^\top \\
        W_Q^\top \\
        W_V^\top
    \end{bmatrix}\right)
    -
    \underbrace{W_VX}_{n\times n}\Softmax
    \underbrace{\left(
    (W_KX)^\top W_QX
    \right) }_{n\times n}
    \|_\infty \leq \epsilon.
    \end{align*}
\end{theorem}

\begin{proof}

Denote $W_K,W_Q,W_V$ as
\begin{align*}
    W_K = [k_1,k_2\cdots,k_n]^\top, \quad
    W_Q = [q_1,q_2\cdots,q_n]^\top, \quad
    W_V= [v_1,v_2\cdots,v_n]^\top.
\end{align*}

We express the input as
\begin{align*}
\begin{bmatrix}
        X \\
    W_K^\top\\
    W_Q^\top\\
    W_V^\top
    \end{bmatrix} =
\begin{bmatrix}
        x_1 & x_2 & \cdots & x_n \\
        k_1 & k_2 & \cdots & k_n \\
        q_1 & q_2 & \cdots & q_n \\
        v_1 & v_2 & \cdots & v_n
\end{bmatrix} \in\R^{4d\times n},
\end{align*}
where $x_j$, $k_j$, $q_j$ and $v_j$ are all $d$ dimensional vectors for $j\in [n]$.

\paragraph{Interpolations.}
Let $B$ denote the bound of $X,W_K,W_Q,W_V$ in infinite norm
\begin{align*}
B= \max(\|X\|_\infty,\|W_K\|_\infty,\|W_Q\|_\infty,\|W_V\|_\infty).
\end{align*}

Then all $k_i^\top x_j$, $q_i^\top x_j$, $v_i^\top x_j$ are bounded by $d\cdot B^2$, for all $i, j\in[n]$.

Define $L_0 := -dB^2$, $L_P:=dB^2$ and
\begin{align*}
L_i := \frac{iL_P+(P-i)L_0}{P},
\end{align*}
where $P$ is the number of interpolation steps, meaning how many divisions are made between $L_0$ and $L_N$, and is a parameter that decides attention size, thereby influencing the precision of approximation.

We use $\Delta L$ to denote the length of the interval between two neighboring interpolations.
Obviously, we have
\begin{align*}
    \Delta L := \frac{L_P-L_0}{P} = \frac{2dB^2}{P}.
\end{align*}

\paragraph{In-Context Calculation of $K$.}

The first layer of attention consists of $3n$ heads.
Among them, heads with labels from $1$ to $n$ are responsible for the calculation of $K$.

Label each head in the $n$ heads as $\Attn_j^K$ for $j\in [n]$ ($j$ denotes the $k_j^\top x$ that we are interpolating to obtain the value).
This writes out as
\begin{align}
\label{eq:attnKj_output}
\Attn_j^K := \Attn_j,\quad  j\in [n].
\end{align}

In addition, for clarity of structure, we also show what the other heads are used for
\begin{align*}
& ~ \Attn_j^Q := \Attn_{n+j},\quad  j\in [n],\annot{Responsible for calculating $Q$}\\
& ~ \Attn_j^V := \Attn_{2n+j},\quad  j\in [n]\annot{Responsible for calculating $V$}.
\end{align*}

We define the linear transformation $\li_j:\R^{4d\times n}\to\R^{(2d+3)\times (P+1)}$ attached before $\Attn_j^K$ as:
\begin{align*}
    \li_j(Z)
    := & ~
    \begin{bmatrix}
    0_{d\times d} & 0_{d\times d} & 0_{d\times 2d}\\
    0_{d\times d} & I_d & 0_{d\times 2d}\\
    0_{3\times d} & 0_{3\times d} & 0_{3\times 2d}\\
    \end{bmatrix}
    \underbrace{Z}_{4d\times n}
    \underbrace{
    \begin{bmatrix}
        2L_{0}\onehot{n}{j} & 2L_{1}\onehot{n}{j} &
        \cdots &
        2L_{P} \onehot{n}{j}
    \end{bmatrix}
    }_{n\times P}
    +\\
    & ~
    \begin{bmatrix}
    I_d & 0_{d\times d} & 0_{d\times 2d}\\
    0_{d\times d} & 0_{d\times d} & 0_{d\times 2d}\\
    0_{3\times d} & 0_{3\times d} & 0_{3\times 2d}\\
    \end{bmatrix}
    Z
    \begin{bmatrix}
        I_n & 0_{n \times (P+1-n)}
    \end{bmatrix}
    +
    \begin{bmatrix}
        0_{2d\times (P+1)} \\
        M_1 \\
        M_L
    \end{bmatrix} \in\R^{(2d+3)\times (P+1)},
\end{align*}
where $M_1, M_L$ are
\begin{align*}
    M_1
    := & ~
    \underbrace{
    \begin{bmatrix}
        1_{1\times n} & 0_{1\times (P-n+1)}
    \end{bmatrix}}_{1\times (P+1)}, \\
    M_L
    := & ~
    \underbrace{
    \begin{bmatrix}
    L_0 & L_1 & \cdots & L_P\\
    -L^2_{0} & -L^2_{1} & \cdots & -L^2_{P}
    \end{bmatrix}
    }_{2\times (P+1)}.
\end{align*}

The $\li_j^K$ layer outputs
\begin{align}
\label{eq:linear_ij}
    & ~\li_j^K\left(
    \begin{bmatrix}
        X \\
        W_K^\top\\
        W_Q^\top\\
        W_V^\top
    \end{bmatrix}
    \right) \notag\\
    =&  ~
    \underbrace{\begin{bmatrix}
        0_d & 0_d & \cdots & 0_d\\
        2L_{0}k_j & 2L_{1}k_j &
        \cdots &
         2L_{P}k_j\\
        0_3 & 0_3 & \cdots & 0_3\\
    \end{bmatrix}}_{(2d+3) \times (P+1)} +
    \underbrace{
    \begin{bmatrix}
        x_1 & x_2 & \cdots & x_n & 0_d & \cdots & 0_d\\
        0_d & 0_d & \cdots & 0_d & 0_d & \cdots & 0_d \\
        0_3 & 0_3 & \cdots & 0_3 & 0_3 & \cdots & 0_3\\
    \end{bmatrix}
    }_{(2d+3)\times (P+1)}
    +
    \underbrace{
    \begin{bmatrix}
        0_{2d\times (P+1)} \\
        M_1 \\
        M_L
    \end{bmatrix}
    }_{(2d+3)\times (P+1)}\notag\\
    = & ~
    \begin{bmatrix}
        x_1 & x_2 & \cdots & x_n &
        0_d & \cdots & 0_d\\
        2L_{0}k_j & 2L_{1}k_j &
        \cdots &
        2L_{n-1}k_j & 2L_nk_j & \cdots &  2L_{P}k_j\\
        1 & 1 & \cdots & 1 & 0 & \cdots & 0\\
        L_0 & L_1 & \cdots & L_{n-1} & L_n & \cdots & L_P\\
        -L^2_{0} & -L^2_{1} &
        \cdots &
        -L^2_{n-1} & -L^2_n & \cdots & -L^2_{P}
    \end{bmatrix} \in\R^{(2d+3)\times (P+1)}.
\end{align}

We now construct $\Attn_j^K$ which takes $\eqref{eq:linear_ij}$ as input.
We denote the parameters of $\Attn_j^K$ as $W^{(j;K)}_K$, $W^{(j;K)}_Q$, $W^{(j;K)}_V$ and $W^{(j;K)}_O$.

We construct the $W^{(j;K)}_K, W^{(j;K)}_Q$ to be
\begin{align*}
    W^{(j;K)}_K
    := & ~
    \beta
    \begin{bmatrix}
        0_{d\times d} & I_d & 0_{d\times 1} & 0_{d\times 1} & 0_{d\times 1} \\
        0_{1\times d} & 0_{1\times d} & 0 & 0 & 1
    \end{bmatrix}
    \in\R^{(d+1)\times (2d+3)},\\
    W^{(j;K)}_Q
    := & ~
    \begin{bmatrix}
        I_d & 0_{d\times d} & 0_{d\times 1} & 0_{d\times 1} & 0_{d\times 1} \\
        0_{1\times d} & 0_{1\times d} & 1 & 0 & 0
    \end{bmatrix}
    \in\R^{(d+1)\times (2d+3)},
\end{align*}
where $\beta\in\R$ is a temperature coefficient of Softmax function.

This yields the $K^{(j)},Q^{(j)}$ of $\Attn_j^K$ to be:
\begin{align*}
    K^{(j)} := & ~
    W^{(j;K)}_K \cdot
    \li_j^K\left(
    \begin{bmatrix}
        X \\
        W_K^\top\\
        W_Q^\top\\
        W_V^\top
    \end{bmatrix}
    \right)\\
    = & ~
    \beta\begin{bmatrix}
        0_{d\times d} & I_d & 0_{d\times 1} & 0_{d\times 1} & 0_{d\times 1} \\
        0_{1\times d} & 0_{1\times d} & 0 & 0 & 1
    \end{bmatrix}
    \cdot
    \begin{bmatrix}
        x_1 & x_2 & \cdots & x_n & \cdots & 0_d\\
        2L_{0}k_j & 2L_{1}k_j &
        \cdots &
        2L_{n-1}k_j & \cdots & 2L_{P}k_j\\
        1 & 1 & \cdots & 1 & \cdots & 1\\
        L_0 & L_1 & \cdots & L_{n-1} & \cdots & L_P\\
        -L^2_{0} & -L^2_{1} &
        \cdots &
        -L^2_{n-1} & \cdots & -L^2_{P}
    \end{bmatrix} \annot{By \eqref{eq:linear_ij}}\\
    = & ~
    \beta\begin{bmatrix}
        2L_{0}k_j & 2L_{1}k_j &
        \cdots &
        2L_{n-1}k_j & \cdots & 2L_{P}k_j\\
        -L^2_{0} & -L^2_{1} &
        \cdots &
        -L^2_{n-1} & \cdots & -L^2_{P}
    \end{bmatrix} \in\R^{(d+1)\times(P+1)},
\end{align*}
and
\begin{align*}
    Q^{(j)} := & ~
    W^{(j;K)}_Q \li^K_j
    \left(
    \begin{bmatrix}
        X \\
        W_K^\top\\
        W_Q^\top\\
        W_V^\top
    \end{bmatrix}
    \right)\\
    = & ~
    \begin{bmatrix}
        I_d & 0_{d\times d} & 0_{d\times 1} & 0_{d\times 1} & 0_{d\times 1} \\
        0_{1\times d} & 0_{1\times d} & 1 & 0 & 0
    \end{bmatrix}
    \cdot
    \begin{bmatrix}
        x_1 & x_2 & \cdots & x_n & \cdots & 0_d\\
        2L_{0}k_j & 2L_{1}k_j &
        \cdots &
        2L_{n-1}k_j & \cdots & 2L_{P}k_j\\
        1 & 1 & \cdots & 1 & \cdots & 1\\
        L_0 & L_1 & \cdots & L_{n-1} & \cdots & L_P\\
        -L^2_{0} & -L^2_{1} &
        \cdots &
        -L^2_{n-1} & \cdots & -L^2_{P}
    \end{bmatrix} \annot{By \eqref{eq:linear_ij}}\\
    = & ~
    \begin{bmatrix}
        x_1 & x_2 & \cdots & x_n & 0_{d\times (P-n+1)} \\
        1 & 1 & \cdots & 1 & 0_{1\times (P-n+1)}
    \end{bmatrix}\in\R^{(d+1)\times(P+1)}.
\end{align*}

Combining the above result of $K^{(j)}$ and $Q^{(j)}$, we calculate $\Softmax((K^{(j)})^\top Q^{(j)})$ as
\begin{align}
\label{eq:output_attn}
    & ~ \Softmax((K^{(j)})^\top Q^{(j)}) \nonumber\\
    = & ~
    \Softmax
    (\beta
    \begin{bmatrix}
        2L_{0}k_j & 2L_{1}k_j &
        \cdots &
        2L_{n-1}k_j & \cdots & 2L_{P}k_j\\
        -L^2_{0} & -L^2_{1} &
        \cdots &
        -L^2_{n-1} & \cdots & -L^2_{P}
    \end{bmatrix}^\top
    \begin{bmatrix}
        x_1 & x_2 & \cdots & x_n & 0_{d\times (P-n+1)} \\
        1 & 1 & \cdots & 1 & 0_{1 \times (P-n+1)}
    \end{bmatrix}
    )\nonumber\\
    = & ~
    \Softmax
    \left(\beta
    \begin{bmatrix}
        2L_0k_j^\top x_1 -L^2_0 & 2L_0k_j^\top x_2 -L^2_0 & \cdots & 2L_0k_j^\top x_n -L^2_0 & 0_{d\times (P-n+1)} \\
        2L_1k_j^\top x_1 -L^2_1 & 2L_1k_j^\top x_2 -L^2_1 & \cdots & 2L_1k_j^\top x_n -L^2_1 & 0_{d\times (P-n+1)} \\
        \vdots & \vdots & & \vdots & \vdots \\
        2L_Pk_j^\top x_1 -L^2_P & 2L_Pk_j^\top x_2 -L^2_P & \cdots & 2L_Pk_j^\top x_n -L^2_P & 0_{d\times (P-n+1)} \\
    \end{bmatrix}
    \right)\nonumber\\
    = & ~
    \Softmax
    \left(\beta
    \begin{bmatrix}
        -(k_j^\top x_1-L_0)^2+(k_j^\top x_1)^2 & \cdots & -(k_j^\top x_n-L_0)^2+(k_j^\top x_n)^2 & 0_{d\times (P-n+1)} \\
        -(k_j^\top x_1-L_1)^2+(k_j^\top x_1)^2 & \cdots & -(k_j^\top x_n-L_1)^2+(k_j^\top x_n)^2 & 0_{d\times (P-n+1)} \\
        \vdots &  & \vdots & \vdots \\
        -(k_j^\top x_1-L_P)^2+(k_j^\top x_1)^2 & \cdots & -(k_j^\top x_n-L_P)^2+(k_j^\top x_n)^2 & 0_{d\times (P-n+1)} \\
    \end{bmatrix}
    \right) \nonumber\\
    = & ~
    \Softmax
    \left(\beta
    \begin{bmatrix}
        -(k_j^\top x_1-L_0)^2 & \cdots & -(k_j^\top x_n-L_0)^2 & 0_{d\times (P-n+1)} \\
        -(k_j^\top x_1-L_1)^2 & \cdots & -(k_j^\top x_n-L_1)^2 & 0_{d\times (P-n+1)} \\
        \vdots &  & \vdots & \vdots \\
        -(k_j^\top x_1-L_P)^2 & \cdots & -(k_j^\top x_n-L_P)^2 & 0_{d\times (P-n+1)} \\
    \end{bmatrix}
    \right) \in\R^{(P+1)\times(P+1)},
\end{align}
where the last line is by the following property of $\Softmax$ over each column ($C$ corresponds to $(k_j^\top x_i)^2$ at each column)
\begin{align*}
    \Softmax(v) = \Softmax(v+1_{P+1}\cdot C),
\end{align*}
for any vector $v\in \R^{P+1}$ and $C\in \R$.

This yields
\begin{align*}
    \Softmax((K^{(j)})^\top Q^{(j)})_{r,c}
    =
    \frac{ e^{-\beta(L_r-k_j^\top x_c)^2}}{\sum_{s=0}^P e^{-\beta(L_s-k_j^\top x_c)^2}},
\end{align*}
for every $r\in[P+1]$ and $c\in[n]$.

Construct $W_V^{(j)}$ to be
\begin{align*}
W_V^{(j)} :=
\onehot{3n}{j}
\begin{bmatrix}
        0_{1\times (2d+1)} & 1 & 0
    \end{bmatrix}\in\R^{3n\times (2d+3)}.
\end{align*}
$W_V^{(j)}$ also writes as
\begin{align*}
W_V^{(j)} =
\begin{bmatrix}
        0_{3n\times d} & 0_{3n\times d} & 0_{3n} & \onehot{3n}{j} & 0_{3n}
\end{bmatrix}.
\end{align*}

Thus $V^{(j)}$ becomes
\begin{align*}
    V := & ~
    W_V^{(j)}
    \li_{j}^K
    \left(
    \begin{bmatrix}
        X\\
        W_K^\top\\
        W_Q^\top\\
        W_V^\top
    \end{bmatrix}
    \right)\\
    = & ~
    \begin{bmatrix}
        0_{3n\times d} & 0_{3n\times d} & 0_{3n} & \onehot{3n}{j} & 0_{3n}
    \end{bmatrix}
    \cdot
    \begin{bmatrix}
        x_1 & x_2 & \cdots & x_n &
        0_d & \cdots & 0_d\\
        2L_{0}k_j & 2L_{1}k_j &
        \cdots &
        2L_{n-1}k_j & 2L_nk_j & \cdots &  2L_{P}k_j\\
        1 & 1 & \cdots & 1 & 0 & \cdots & 0\\
        L_0 & L_1 & \cdots & L_{n-1} & L_n & \cdots & L_P\\
        -L^2_{0} & -L^2_{1} &
        \cdots &
        -L^2_{n-1} & -L^2_n & \cdots & -L^2_{P}
    \end{bmatrix}\annot{By \eqref{eq:linear_ij}}\\
    = & ~
    \underbrace{
    \begin{bmatrix}
        L_0 & L_1 & \cdots & L_{n-1} & L_n & \cdots & L_P
    \end{bmatrix}\onehot{3n}{j}}_{\R^{3n\times (P+1)}}.
\end{align*}

Combine the above equation with\eqref{eq:output_attn} we have
\begin{align}
\label{eq:attn_output2}
    V^{(j)}\Softmax((K^{(j)})^\top Q^{(j)})_{:,c}
    =
    \sum_{r=0}^P
    \frac{ e^{-\beta(L_r-k_j^\top x_c)^2}}{\sum_{s=0}^P e^{-\beta(L_s-k_j^\top x_c)^2}}
    L_r\onehot{3n}{j},
\end{align}
for every $c\in [n]$.

$V^{(j)}\Softmax((K^{(j)})^\top Q^{(j)})_{:,c}$ is a weighted average of $L_r\onehot{3n}{j}$ where the weight of each element is determined by the distance between $L_r$ and $k_j^\top x_c$.

We demonstrate this value is close to $k_j^\top x_c$ by showing the elements in the weighted average far from $k_j^\top x_c$ have negligible weight.

Specifically, we have the following calculations
\begin{align*}
& ~ \| %
V^{(j)}\Softmax((K^{(j)})^\top Q^{(j)})_{:,c}
-
k_j^\top x_c\onehot{3n}{j}
\|_\infty \\
= & ~ %
\|
\sum_{r=0}^P
    \frac{ e^{-\beta(L_r-k_j^\top x_c)^2}}{\sum_{s=0}^P e^{-\beta(L_s-k_j^\top x_c)^2}}
    (L_r-k_j^\top x_c)\onehot{3n}{j}
\|_\infty \annot{By \eqref{eq:attn_output2}}\\
= & ~ %
|
\sum_{r=0}^P
    \frac{ e^{-\beta(L_r-k_j^\top x_c)^2}}{\sum_{s=0}^P e^{-\beta(L_s-k_j^\top x_c)^2}}
    (L_r-k_j^\top x_c)
| \\
\leq & ~ %
|
\sum_{|L_r-k_j^\top x_c| \leq \Delta L}
\frac{ e^{-\beta(L_r-k_j^\top x_c)^2}}{\sum_{s=0}^P e^{-\beta(L_s-k_j^\top x_c)^2}}
    (L_r-k_j^\top x_c)
|
\\
& ~ +
|
\sum_{|L_r-k_j^\top x_c| \geq \Delta L}
\frac{ e^{-\beta(L_r-k_j^\top x_c)^2}}{\sum_{s=0}^P e^{-\beta(L_s-k_j^\top x_c)^2}}
    (L_r-k_j^\top x_c)
| \annot{Sperate the summation}\\
\leq & ~ %
\sum_{|L_r-k_j^\top x_c| \leq \Delta L}
\frac{ e^{-\beta(L_r-k_j^\top x_c)^2}}{\sum_{s=0}^P e^{-\beta(L_s-k_j^\top x_c)^2}}
|
L_r-k_j^\top x_c
|
\\
& ~ +
\sum_{|L_r-k_j^\top x_c| \geq \Delta L}
\frac{ e^{-\beta(L_r-k_j^\top x_c)^2}}{\sum_{s=0}^P e^{-\beta(L_s-k_j^\top x_c)^2}}
|
L_r-k_j^\top x_c
|\\
\leq & ~ %
\frac{\sum_{|L_r-k_j^\top x_c| \leq \Delta L} e^{-\beta(L_r-k_j^\top x_c)^2}}{\sum_{s=0}^P e^{-\beta(L_s-k_j^\top x_c)^2}}\Delta L
+
\frac{\sum_{|L_r-k_j^\top x_c| \geq \Delta L} e^{-\beta(L_r-k_j^\top x_c)^2}}{\sum_{s=0}^P e^{-\beta(L_s-k_j^\top x_c)^2}}2dB^2 \\
\leq & ~ %
\Delta L
+
\frac{\sum_{|L_r-k_j^\top x_c| \geq \Delta L} e^{-\beta\Delta L^2}}{e^{-\beta \frac{\Delta L^2}{4}}}2dB^2
\annot{By $\sum_{|L_r-k_j^\top x_c| \leq \Delta L} e^{-\beta(L_r-k_j^\top x_c)^2} $is part of $ \sum_{s=0}^P e^{-\beta(L_s-k_j^\top x_c)^2}$}\\
= & ~
\Delta L
+
\sum_{|L_r-k_j^\top x_c| \geq \Delta L} e^{-\frac{3}{4}\beta\Delta L^2}2dB^2 \\
\leq & ~ %
\Delta L
+
P e^{-\frac{3}{4}\beta\Delta L^2}2dB^2.
\end{align*}

The above calculations conclude the following result
\begin{align}
\label{eq:attn_kjxc_dif}
\| %
V^{(j)}\Softmax((K^{(j)})^\top Q^{(j)})_{:,c}
-
k_j^\top x_c\onehot{3n}{j}
\|_\infty
\leq
\Delta L
+
P e^{-\frac{3}{4}\beta\Delta L^2}2dB^2.
\end{align}

For any $\epsilon_1>0$, configure $P$ to be larger than $4dB^2/\epsilon_1$.
This means
\begin{align*}
    \Delta L = \frac{2dB^2}{P} \leq \frac{\epsilon_1}{2}.
\end{align*}

Then we set $\beta$ to satisfy
\begin{align*}
\beta \geq \frac{4}{4}\ln(\frac{4dB^2P}{\epsilon_1})B\Delta L.
\end{align*}

This means
\begin{align*}
    P e^{-\frac{3}{4}\beta\Delta L^2}2dB^2
    \leq \frac{\epsilon_1}{2}.
\end{align*}

Thus by \eqref{eq:attn_kjxc_dif} we have
\begin{align*}
   \left\| \sum_{j=1}^n \Attn^K_j\circ\li^K_j\left(
    \begin{bmatrix}
        X \\
        W_K^\top\\
        W_Q^\top\\
        W_V^\top
    \end{bmatrix}
    \right)
    -
    \begin{bmatrix}
        K \\
        0_{n\times n}\\
        0_{n\times n}
    \end{bmatrix}
    \right\|_\infty
    \leq & ~
    \Delta L
+
P e^{-\frac{3}{4}\beta\Delta L^2}2dB^2\\
\leq & ~
\frac{\epsilon_1}{2} +\frac{\epsilon_1}{2} \\
= & ~
    \epsilon_1,
\end{align*}
for any $\epsilon_1>0$.

\paragraph{In-Context Calculation of $Q$ and $V$.}

Let $\li_j^Q$, $\li_j^V$ denote the linear layers preceding $\Attn_j^Q$ and $Attn_j^V$.
These layers are constructed similarly to those of $K$.
We define $\li_j^Q$, $\li_j^V$ as
\begin{align*}
    \li_j^Q(Z)
    := & ~
    \begin{bmatrix}
    0_{d\times d} & 0_{d\times d} & 0_{d\times d} &0_{d\times d}\\
    0_{d\times d} & 0_{d\times d} & I_d & 0_{d\times d}\\
    0_{3\times d} & 0_{3\times d} & 0_{3\times d} & 0_{3\times d}\\
    \end{bmatrix}
    Z
    \begin{bmatrix}
        L_{0}\onehot{n}{j} & L_{1}\onehot{n}{j} &
        \cdots &
        L_{P} \onehot{n}{j}
    \end{bmatrix}
    +\\
    & ~
    \begin{bmatrix}
    I_d & 0_{d\times d} & 0_{d\times 2d}\\
    0_{d\times d} & 0_{d\times d} & 0_{d\times 2d}\\
    0_{3\times d} & 0_{3\times d} & 0_{3\times 2d}\\
    \end{bmatrix}
    Z
    \begin{bmatrix}
        I_n & 0_{n \times (P+1-n)}
    \end{bmatrix}
    +
    \underbrace{
    \begin{bmatrix}
        0_{2d\times (P+1)} \\
        M_1 \\
        M_L
    \end{bmatrix}}_{(2d+3)\times (P+1)},
\end{align*}
and
\begin{align*}
    \li_j^V(Z)
    := & ~
    \begin{bmatrix}
    0_{d\times d} & 0_{d\times d} & 0_{d\times d} &0_{d\times d}\\
    0_{d\times d} & 0_{d\times d} & 0_{d\times d} & I_d\\
    0_{3\times d} & 0_{3\times d} & 0_{3\times d} & 0_{3\times d}\\
    \end{bmatrix}
    Z
    \begin{bmatrix}
        L_{0}\onehot{n}{j} & L_{1}\onehot{n}{j} &
        \cdots &
        L_{P} \onehot{n}{j}
    \end{bmatrix}
    +\\
    & ~
    \begin{bmatrix}
    I_d & 0_{d\times d} & 0_{d\times 2d}\\
    0_{d\times d} & 0_{d\times d} & 0_{d\times 2d}\\
    0_{3\times d} & 0_{3\times d} & 0_{3\times 2d}\\
    \end{bmatrix}
    Z
    \begin{bmatrix}
        I_n & 0_{n \times (P+1-n)}
    \end{bmatrix}
    +
    \underbrace{
    \begin{bmatrix}
        0_{2d\times (P+1)} \\
        M_1 \\
        M_L
    \end{bmatrix}
    }_{(2d+3)\times (P+1)}.
\end{align*}

Similar to $\li_j^K$, they output
\begin{align*}
    & ~ \li_j^Q(Z)
    =
    \begin{bmatrix}
        x_1 & x_2 & \cdots & x_n &
        0_d & \cdots & 0_d\\
        2L_{0}q_j & 2L_{1}q_j &
        \cdots &
        2L_{n-1}q_j & 2L_nq_j & \cdots &  2L_{P}q_j\\
        1 & 1 & \cdots & 1 & 0 & \cdots & 0\\
        L_0 & L_1 & \cdots & L_{n-1} & L_n & \cdots & L_P\\
        -L^2_{0} & -L^2_{1} &
        \cdots &
        -L^2_{n-1} & -L^2_n & \cdots & -L^2_{P}
    \end{bmatrix}
    \\
    & ~
    \li_j^V(Z)
    =
    \begin{bmatrix}
        x_1 & x_2 & \cdots & x_n &
        0_d & \cdots & 0_d\\
        2L_{0}v_j & 2L_{1}v_j &
        \cdots &
        2L_{n-1}v_j & 2L_nv_j & \cdots &  2L_{P}v_j\\
        1 & 1 & \cdots & 1 & 0 & \cdots & 0\\
        L_0 & L_1 & \cdots & L_{n-1} & L_n & \cdots & L_P\\
        -L^2_{0} & -L^2_{1} &
        \cdots &
        -L^2_{n-1} & -L^2_n & \cdots & -L^2_{P}
    \end{bmatrix},
\end{align*}
which only replaces the output of $k_j$ in $\li_j^K$ to $q_j$ and $v_j$ respectively.

The attention in the in-context calculation of $Q$ and $V$ are the same with that in $K$ except for the $W_V^{(j)}$ part.
Denote them as $W_V^{(j;Q)}$ and $W_V^{(j;V)}$.

For $Q$, we have
\begin{align*}
W_V^{(j;Q)}
:=
\onehot{3n}{j+n}
\begin{bmatrix}
        0_{1\times d} & 0_{1\times d} & 0 & 1 & 0
    \end{bmatrix}.
\end{align*}

For $V$, we have
\begin{align*}
W_V^{(j;V)}
:=
    \onehot{3n}{j+2n}
\begin{bmatrix}
        0_{1\times d} & 0_{1\times d} & 0 & 1 & 0
    \end{bmatrix}.
\end{align*}

In the occasion of $K$, we have
\begin{align*}
   \| \sum_{j=1}^n \Attn^K_j\circ\li^K_j\left(
    \begin{bmatrix}
        X \\
        W_K^\top\\
        W_Q^\top\\
        W_V^\top
    \end{bmatrix}
    \right)
    -
    \begin{bmatrix}
        K \\
        0_{n\times n}\\
        0_{n\times n}
    \end{bmatrix}
    \|_\infty
    \leq & ~
    \epsilon_1.
\end{align*}

Similarly, the outputs of $\sum_{j=1}^n\Attn^Q_j\circ\li^Q_j$ and $\sum_{j=1}^n
\Attn^V_j\circ\li^V_j$ satisfy
\begin{align*}
& ~ \| \sum_{j=1}^n
\Attn^Q_j\circ\li^Q_j\left(
    \begin{bmatrix}
        X \\
        W_K^\top\\
        W_Q^\top\\
        W_V^\top
    \end{bmatrix}
    \right)
    -
    \begin{bmatrix}
        0_{n\times n}\\
        Q \\
        0_{n\times n}
    \end{bmatrix}
    \|_\infty
    \leq
    \epsilon_1, \\
& ~ \| \sum_{j=1}^n
\Attn^V_j\circ\li^V_j\left(
    \begin{bmatrix}
        X \\
        W_K^\top\\
        W_Q^\top\\
        W_V^\top
    \end{bmatrix}
    \right)
    -
    \begin{bmatrix}
        0_{n\times n}\\
        0_{n\times n}\\
        V
    \end{bmatrix}
    \|_\infty
    \leq
    \epsilon_1.
\end{align*}

As previously stated in \eqref{eq:attnKj_output}, $\Attn_j, \Attn_{n+j}$ and $\Attn_{2n+j}$ denote $\Attn^K_j, \Attn^Q_j$ and $\Attn^V_j$ respectively.

Then the output of the complete network satisfies
\begin{align*}
\|
    \sum_{j=1}^{3n}
\Attn_j\circ\li_j\left(
    \begin{bmatrix}
        X \\
        W_K^\top\\
        W_Q^\top\\
        W_V^\top
    \end{bmatrix}
    \right)
    -
    \underbrace{
    \begin{bmatrix}
        K \\
        Q \\
        V
    \end{bmatrix}
    }_{3n\times n}
    \|_\infty
    \leq \epsilon_1.
    \annot{notice the non-zero parts of the three counterparts do not intersect}
\end{align*}

\paragraph{In-Context Calculation of Attention.}

We construct $\Attn_s$ to be
\begin{align*}
    \Attn_s(Z)
    :=
    \underbrace{
    \begin{bmatrix}
        0_{n\times 2n} & I_n
    \end{bmatrix}}_{W_V}
    Z
    \Softmax(
    (
    \underbrace{\begin{bmatrix}
        I_n & 0_{n\times 2n}
    \end{bmatrix}}_{W_K}
    Z)^\top
    \underbrace{
    \begin{bmatrix}
        0_{n\times n} & I_n & 0_{n\times n}
    \end{bmatrix}
    }_{W_Q}
    Z
    ).
\end{align*}

We have
\begin{align*}
    \Attn_s\left(
    \begin{bmatrix}
        K\\
        Q\\
        V
    \end{bmatrix}
    \right)
    =
    \underbrace{W_VX}_{n\times n}\Softmax
    \underbrace{\left(
    (W_KX)^\top W_QX
    \right) }_{n\times n}.
\end{align*}

Now we estimate the error due to the difference between the output of $\sum_{j=1}^{3n}
\Attn_j\circ\li_j\left(
\begin{bmatrix}
    X^\top&
    W_K&
    W_Q&
    W_V
\end{bmatrix}^\top
\right)$ and
$
\begin{bmatrix}
    K^\top & Q^\top & V^\top
\end{bmatrix}^\top$.

Define
\begin{align*}
\begin{bmatrix}
    K' \\
    Q' \\
    V'
\end{bmatrix}
:=
    \sum_{j=1}^{3n}
\Attn_j\circ\li_j\left(
    \begin{bmatrix}
        X \\
        W_K^\top\\
        W_Q^\top\\
        W_V^\top
    \end{bmatrix}
    \right).
\end{align*}
This is the same as
\begin{align*}
    \sum_{j=1}^n \Attn^K_j\circ\li^K_j\left(
    \begin{bmatrix}
        X \\
        W_K^\top\\
        W_Q^\top\\
        W_V^\top
    \end{bmatrix}
    \right)
    =
    \begin{bmatrix}
        K' \\
        0_{n\times n}\\
        0_{n\times n}
    \end{bmatrix}, \\
    \sum_{j=1}^n \Attn^Q_j\circ\li^Q_j\left(
    \begin{bmatrix}
        X \\
        W_K^\top\\
        W_Q^\top\\
        W_V^\top
    \end{bmatrix}
    \right)
    =
    \begin{bmatrix}
        0_{n\times n}\\
        Q' \\
        0_{n\times n}
    \end{bmatrix},\\
    \sum_{j=1}^n \Attn^V_j\circ\li^V_j\left(
    \begin{bmatrix}
        X \\
        W_K^\top\\
        W_Q^\top\\
        W_V^\top
    \end{bmatrix}
    \right)
    =
    \begin{bmatrix}
        0_{n\times n}\\
        0_{n\times n}\\
        V'
    \end{bmatrix}.
\end{align*}

In this definition, we have
\begin{align*}
    & ~ \Attn_s\left(
    \begin{bmatrix}
        K'\\
        Q'\\
        V'
    \end{bmatrix}
    \right)
    -
    \Attn_s\left(
    \begin{bmatrix}
        K\\
        Q\\
        V
    \end{bmatrix}
    \right)\\
    = & ~
    V'\Softmax
    \left(
    K'^\top Q'
    \right)
    -
    V\Softmax
    \left(
    K^\top Q
    \right) \\
    = & ~
    (V'-V)\Softmax
    \left(
    K'^\top Q'
    \right)+
    V(\Softmax
    \left(
    K'^\top Q'
    \right)-
    \Softmax
    \left(
    K^\top Q
    \right)).
\end{align*}

This yields that
\begin{align*}
& ~ \left\|
    \Attn_s\left(
    \begin{bmatrix}
        K'\\
        Q'\\
        V'
    \end{bmatrix}
    \right)
    -
    \Attn_s\left(
    \begin{bmatrix}
        K\\
        Q\\
        V
    \end{bmatrix}
    \right)
    \right\|_\infty \\
    = & ~
    \left\|
    (V'-V)\Softmax
    \left(
    K'^\top Q'
    \right)+
    V(\Softmax
    \left(
    K'^\top Q'
    \right)-
    \Softmax
    \left(
    K^\top Q
    \right))
    \right\|_\infty\\
    \leq & ~
    \left\|
    (V'-V)\Softmax
    \left(
    K'^\top Q'
    \right)\right\|_\infty
    +
    \left\|
    V(\Softmax
    \left(
    K'^\top Q'
    \right)-
    \Softmax
    \left(
    K^\top Q
    \right))
    \right\|_\infty \annot{Triangle inequality}\\
    \leq & ~ %
    \epsilon_1
    +
    n\left\|
    V\right\|_\infty
    \left\|\Softmax
    \left(
    K'^\top Q'
    \right)-
    \Softmax
    \left(
    K^\top Q
    \right)
    \right\|_\infty.
\end{align*}

The last inequality holds true because $\Softmax(K'^\top Q')$ has each column summing up to $1$, which means
\begin{align*}
(V'-V)\Softmax
\left(
K'^\top Q'
\right)_{:,j}   ,
\end{align*}
is a weighted sum of the columns in $V'-V$.

Thus we have
\begin{align*}
\|(V'-V)\Softmax
\left(
K'^\top Q'
\right)_{:,j} \|_\infty
\leq & ~
\|V-V'\|_\infty\leq \epsilon_1.
\end{align*}
because the average of columns has a smaller maximal entry than the original columns.

Since $X,W_K,W_Q,W_V$ are all bounded in infinite norm, the multiplication among them is also bounded in infinite norm. We denote this bound as $B_{KQV}$.

Now we calculate the error of the last term in the above inequality
\begin{align*}
    & ~ |(\Softmax
    \left(
    K'^\top Q'
    \right)-
    \Softmax
    \left(
    K^\top Q
    \right))_{i,j}| \\
    = & ~
    |\frac{e^{K_i'\cdot Q_j'}}{\sum_{i'=1}^ne^{K_{i'}' \cdot Q_j'}}
    -
    \frac{e^{K_{i} \cdot Q_j}}{\sum_{i'=1}^ne^{K_{i'}\cdot Q_j}}|\\
    = & ~
    |\frac{e^{K_i'\cdot Q_j'}-e^{K_{i} \cdot Q_j}}{\sum_{i'=1}^ne^{K_{i'}' \cdot Q_j'}}|
    +
    |e^{K_{i} \cdot Q_j}(\frac{1}{\sum_{i'=1}^ne^{K_{i'}'\cdot Q_j'}}-\frac{1}{\sum_{i'=1}^ne^{K_{i'}\cdot Q_j}})|\\
    \leq & ~
    \frac{e^{K_i'\cdot Q_j'}}{\sum_{i'=1}^ne^{K_{i'}' \cdot Q_j'}}|1-e^{K_i'\cdot Q_j'-K_i\cdot Q_j}|
    \\
    & ~ +
    e^{K_iQ_j}\frac{1}{(\sum_{i'=1}^ne^{K_{i'}' \cdot Q_j'})(\sum_{i'=1}^ne^{K_{i'}\cdot Q_j})}|\sum_{i'=1}^ne^{K_{i'}\cdot Q_j}-\sum_{i'=1}^ne^{K_{i'}'\cdot Q_j'}| \\
    \leq & ~
    |1 - e^{\epsilon_1(K_i'+Q_j')+\epsilon_1^2}|+|1-\frac{\sum_{i'=1}^ne^{K_{i'}'\cdot Q_j'}}{\sum_{i'=1}^ne^{K_{i'}\cdot Q_j}}| \\
    \leq & ~
   2|1 - e^{\epsilon_1(K_i'+Q_j')+\epsilon_1^2}| \annot{the largest deviation ratio between $e^{K_{i'}\cdot Q_j}$ and $e^{K_{i'}'\cdot Q_j'}$ are invariant of $i$}\\
   \leq & ~
   2|1-e^{2\epsilon_1(B_{KQV}+\frac{1}{2})}|
,
\end{align*}
where $K_i'$ and $Q_i'$ denote the columns in $K'$ and $Q'$ respectively.

Thus $\epsilon_1$ satisfies
\begin{align*}
    \epsilon_1 \leq & ~ \frac{\ln(1+\frac{\epsilon}{2})}{2B_{KQV}+1}.
\end{align*}

This completes the proof.
\end{proof}

\subsection{Proof of \texorpdfstring{\cref{thm:attn_sim_stats_methods}}{}}
\label{proof:thm:attn_sim_stats_methods}

\begin{theorem}[Restate of \cref{thm:attn_sim_stats_methods}: In-Context Emulation of Statistical Methods]
Let $\calA$ denote the set of all the in-context algorithms that a single-layer attention is able to approximate.
For an $a\in \calA$ (that is, a specific algorithm), let $W_K^a,W_Q^a,W_V^a$ denote the weights of the attention that implements this algorithm.
For any $\epsilon>0$ and any finite set $\calA_0 \in \calA$, there exists a $2$-layer attention $\Attn \circ \Attn_m$ such that
\begin{align*}
    \|
    \sum_{j=1}^{3n}\Attn_s \circ \Attn_j\circ\li_j
    \left(
    \begin{bmatrix}
        X \\
        W^a
    \end{bmatrix}
    \right)
    -
    a(X)
    \|_\infty \leq \epsilon,
    \quad a\in \calA_0,
\end{align*}
where $W^a$ is the $W$ defined as \cref{def:input_attn} using $W_K^a,W_Q^a,W_V^a$.
\end{theorem}

\begin{proof}
Without loss of generality, assume all $W_K^a,W_Q^a,W_V^a$ to be of the same hidden dimension since we are always able to pad them to the same size.
According to \cref{thm:attn_sim_attn}, there exists a network $\sum_{j=1}^n\Attn_s\circ \Attn_j\circ\li_j$ that approximate $a(X)$ with an error no larger than $\epsilon>0$ when given input of the form:
\begin{align*}
    \begin{bmatrix}
        X \\
        W_K^{a\top} \\
        W_Q^{a\top} \\
        W_V^{a\top}
    \end{bmatrix}.
\end{align*}

Then for a set of $a\in\calA_0$, define $P_m := \max_{a\in \calA_0}P_\epsilon^{(a)}$.

By \cref{thm:attn_sim_attn}, there exists a network consisting of a self-attention followed by a multi-head attention with a linear layer and parameter $P$ equals to $P_m$, such that for any $a\in\calA_0$, we have
\begin{align*}
    \|
    \sum_{j=1}^{3n}\Attn_s \circ \Attn_j\circ \li_j
    \left(
    \begin{bmatrix}
        X \\
        W_K^{a\top} \\
        W_Q^{a\top} \\
        W_V^{a\top}
    \end{bmatrix}
    \right)
    -
    a(X)
    \|_\infty \leq \epsilon,
    \quad a\in \calA_0.
\end{align*}
This completes the proof.
\end{proof}

\section{In-Context Application of Statistical Methods by Modern Hopfield Network}
\label{subsec:In-Context Application of Statistical Methods by Hopfield Network}

\begin{definition}[Modern Hopfield Network]
\label{def:hopfield}
    Define $Y=(y_1, \cdots, y_N)^\top \in\R^{d_y\times N}$ as the raw stored pattern, $R=(r_1, \cdots, r_S)^\top \in\R^{R_r\times S}$ as the raw state pattern, and $W_Q\in\R^{d\times d_r}$, $W_K\in\R^{d\times d_y}$ , $W_V\in\R^{d_v\times d}$ as the projection matrices.
    A Hopfield layer $\hop$ is defined as:
    \begin{align}
    \label{eq:hopfield}
        \hop(R; Y,W_Q,W_K,W_V) :=
        \underbrace{W_V}_{d_v\times d} \overbrace{W_KY}^{d\times N}\Softmax(\underbrace{\beta (W_KY)^\top W_QR}_{N\times S})\in\R^{d_v\times S},
    \end{align}
    where $\beta$ is a temperature parameter.

    With $K\in\R^{d\times N}$ denoting $W_KY$, $Q\in\R^{d\times S}$ denoting $W_QR$ and $V\in\R^{d_v\times N}$ denoting $W_VW_KY$, \eqref{eq:hopfield} writes out as:
    \begin{align*}
        \hop(R; Y,W_Q,W_K,W_V) :=
        V\Softmax(
        \beta\cdot K^\top Q
        )\in\R^{d_v\times S}.
    \end{align*}
\end{definition}

\begin{theorem}
\label{thm:hopfield_approx}
    Let $Z = [z_1,z_2,\cdots,z_n]\in \R^{d\times n}$ denote the input from a compact input domain.
    For any linear transformation $l(z) = a^\top z+b:\R^d \to \R$, and any continuous function $f:\R\to \R^o$ where $o$ is the output dimension, there exists a Hopfield network $\hop$ such that
    \begin{align*}
    \|
        \hop(Z) -
        \begin{bmatrix}
            f(l(z_1)) & f(l(z_2)) & \cdots & f(l(z_n))
        \end{bmatrix}
    \|_\infty
    \leq
    \epsilon,
    \end{align*}
    for any $\epsilon>0$.
\end{theorem}

\begin{proof}

We first perform a simple token-wise linear transformation on the input:
\begin{align*}
    \li(Z) :=
    \begin{bmatrix}
        I_{d\times d} \\
        0_{1\times d}
    \end{bmatrix}
    Z
    +
    \begin{bmatrix}
        0_{d\times n}\\
        1_{1\times n}
    \end{bmatrix}
    =
    \begin{bmatrix}
        Z \\
        1_{1\times n}
    \end{bmatrix} \in \R^{(d+1)\times n}.
\end{align*}

We then construct $W_Q$ to be:
\begin{align*}
    W_Q := I_{(d+1)},
\end{align*}
which is an identity matrix of dimension $\R^{(d+1)\times (d+1)}$.

This yields that
\begin{align*}
    Q := W_Q\li(Z)
    =
    \begin{bmatrix}
        Z \\
        1_{1\times n}
    \end{bmatrix}\in \R^{(d+1)\times n}.
\end{align*}

Following the definition of \textit{Interpolations} in \cref{subsec:Simulation of Attention: Another Construction},
$K,V$ are constructed as (here we omit $Y$ since it's not the input):
\begin{align*}
    & ~ K :=
    \begin{bmatrix}
        2L_0a & 2L_1 a & \cdots & 2L_Pa \\
        2L_0b-L_0^2 & 2L_1b-L_1^2 & \cdots & 2L_Pb-L_P^2
    \end{bmatrix}, \\
    & ~
    V :=
    \begin{bmatrix}
        f(L_0) & f(L_1) &\cdots &
        f(L_P)
    \end{bmatrix}.
\end{align*}
By \cref{def:hopfield}, we have
\begin{align*}
 & ~ \hop(Z)
\\
= & ~
\begin{bmatrix}
    f(L_0) & f(L_1) & \cdots & f(L_P)
\end{bmatrix}
\Softmax
\left(\beta
    \begin{bmatrix}
        2l(z_1)L_0-L_0^2 & 2l(z_2)L_0-L_0^2 & \cdots & 2l(z_n)L_0-L_0^2\\
        2l(z_1)L_1-L_1^2 & 2l(z_2)L_1-L_1^2 & \cdots & 2l(z_n)L_1-L_1^2\\
        \vdots & \vdots & & \vdots \\
        2l(z_1)L_P-L_P^2 & 2l(z_2)L_P-L_P^2 & \cdots & 2l(z_n)L_P-L_P^2\\
    \end{bmatrix}
\right).
\end{align*}

This is equivalent to:
\begin{align*}
& ~ \hop(Z) \\
= & ~
\begin{bmatrix}
    f(L_0) & f(L_1) & \cdots & f(L_P)
\end{bmatrix}
\Softmax
\left(-\beta
    \begin{bmatrix}
        (l(z_1)-L_0)^2 & (l(z_2)-L_0)^2 & \cdots & (l(z_n)-L_0)^2\\
        (l(z_1)-L_1)^2 & (l(z_2)-L_1)^2 & \cdots & (l(z_n)-L_1)^2\\
        \vdots & \vdots & & \vdots \\
        (l(z_1)-L_P)^2 & (l(z_2)-L_P)^2 & \cdots & (l(z_n)-L_P)^2\\
    \end{bmatrix}
\right).
\end{align*}

For any column $c \in [n]$ in $\hop(Z)$, we have
\begin{align*}
\hop(Z)_{:,c}
= & ~
\begin{bmatrix}
    f(L_0) & f(L_1) & \cdots & f(L_P)
\end{bmatrix}
\Softmax(
-\beta
\begin{bmatrix}
    (l(z_c) - L_0)^2\\
    (l(z_c) - L_1)^2\\
    \vdots\\
    (l(z_c) - L_P)^2\\
\end{bmatrix}
)\\
= & ~
\sum_{r=1}^P
\frac{e^{-\beta(l(z_c) - L_r)^2}}
{\sum_{r'=1}^P e^{-\beta(l(z_c) - L_{r'})^2}}
f(L_r).
\end{align*}

When $\beta$ is large enough, we have
\begin{align*}
\sum_{(l(z_c) - L_r)^2 \geq \Delta L} \frac{e^{-\beta(l(z_c) - L_r)^2}}
{\sum_{r'=1}^P e^{-\beta(l(z_c) - L_{r'})^2}}
\leq
\sum_{(l(z_i) - L_r)^2 \geq \Delta L}
\frac{e^{-\beta\Delta L}}
{e^{-\beta\frac{\Delta L}{2}}}
\leq
P e^{-\frac{\beta\Delta L}{2}}\leq \epsilon_1,
\end{align*}
for any $\epsilon_1>0$.

This means that the proportion of the $f(L_r)$ in $\hop(Z)_{:,c}$ that deviates from $l(z_c)$ is no larger than $\epsilon_1$.

Since $f$ and $l$ are continuous, and $Z$ comes from a compact domain, $l(z_i)$ comes from a compact domain for all $i\in [n]$.
Thus $f$ is uniformly continuous on its input domain.
This means that for any $\epsilon_2>0$, there exists a $\delta>0$ such that when$(x-y)^2\leq \delta$, $\|f(x)-f(y)\|_\infty\leq \epsilon_2$.

Configuring $\Delta L \leq \delta$ yields:
\begin{align*}
    & ~ \|\hop(Z)_{:,c}-f(l(z_c))\|_\infty \\
    \leq & ~
    \sum_{r=1}^P
\frac{e^{-\beta(l(z_c) - L_r)^2}}
{\sum_{r'=1}^P e^{-\beta(l(z_c) - L_{r'})^2}}
\|
f(L_r)
-
f(l(z_c))
\|_\infty \\
= & ~
\sum_{(l(z_c) - L_r)^2 \geq \Delta L }
\frac{e^{-\beta(l(z_c) - L_r)^2}}
{\sum_{r'=1}^P e^{-\beta(l(z_c) - L_{r'})^2}}
\|
f(L_r)
-
f(l(z_c))
\|_\infty \\
& ~ +
\sum_{(l(z_c) - L_r)^2 \leq \Delta L}
\frac{e^{-\beta(l(z_c) - L_r)^2}}
{\sum_{r'=1}^P e^{-\beta(l(z_c) - L_{r'})^2}}
\|
f(L_r)
-
f(l(z_c))
\|_\infty\\
\leq & ~
\epsilon_1 \cdot 2B + (1-\epsilon_1)\epsilon_2,
\end{align*}
where $B:= \|f\|_{L_\infty}$ is the bound of $f$ in infinite norm.

We set $\epsilon_2 \leq \epsilon/2$, $\epsilon_1 \leq \epsilon/(4B)$.
This yields:
\begin{align*}
\|\hop(Z)_{:,c}-f(l(z_c))\|_\infty
\leq & ~
\epsilon_1 \cdot 2B + (1-\epsilon_1)\epsilon_2 \\
\leq & ~
\frac{\epsilon}{4B}\cdot 2B + 1 \cdot \frac{\epsilon}{2}
=
\epsilon.
\end{align*}
This completes the proof.
\end{proof}

\begin{theorem}
\label{thrm:hopfield_sim_f}
Define
\begin{align*}
    X := \begin{bmatrix}
        x_1 & x_2 & \cdots & x_n \\
        y_1 & y_2 & \cdots & y_n
    \end{bmatrix} \in\R^{(d+1)\times n}
    \quad\text{and}\quad
    W :=
    \begin{bmatrix}
        w & w & \cdots & w
    \end{bmatrix}\in\R^{d\times n},
\end{align*}

where $x_i \in \R^d$ and $y_i \in \R$ are the input-output pairs.
$w\in \R^d$ is the linear coefficient to optimize.
Suppose $x_i,y_i$ and $w$ are bounded by $B$ in infinite norm.

For any continuous function $f:\R \to \R$, there exists a Hopfield layer $\hop$ with linear connections such that
\begin{align*}
    \|
    \hop(W;X)
    -
    \begin{bmatrix}
        f(w^\top x_1-y_1)x_1 &
        f(w^\top x_2-y_2)x_2 &
        \cdots
        &
        f(w^\top x_n-y_n)x_n
    \end{bmatrix}
    \|_\infty
    \leq
    \epsilon,
\end{align*}
for any $\epsilon > 0$.
\end{theorem}

\begin{proof}

Before plugging input $W$ to the Hopfield layer, we pass it through a linear transformation $\li_w$:
\begin{align*}
    \li_w(W) :=
    \begin{bmatrix}
        I_d \\
        0_{(d+n+2)\times d}
    \end{bmatrix}
    W
    +
    \begin{bmatrix}
        0_{d\times n} \\
        -1_{1 \times n}\\
        0_{d\times n}\\
        -1_{1 \times n}\\
        I_n
    \end{bmatrix}
    =
    \begin{bmatrix}
        W\\
        -1_{1 \times n}\\
        0_{d\times n}\\
        -1_{1 \times n}\\
        I_n
        \end{bmatrix}\in\R^{(2d+n+2)\times n}.
\end{align*}

We also pass $X$ through a linear transformation $\li_x$:
\begin{align*}
    & ~ \li_x(X) \\
    := & ~
    \sum_{i=1}^n
    \underbrace{\begin{bmatrix}
        I_{d+1}\\
        0_{(d+1+n)\times (d+1)}
    \end{bmatrix}}_{(2d+n+2)\times (d+1)}
    \underbrace{X}_{(d+1)\times n}
    \underbrace{\begin{bmatrix}
        0_{n\times (i-1)(P+1)} & 2L_0\onehot{n}{i} & 2L_1\onehot{n}{i} & \cdots & 2L_P\onehot{n}{i} & 0_{n\times(n-i)(P+1)}
    \end{bmatrix}}_{n \times n(P+1)} \\
    & ~ +
    \sum_{i=1}^n
    \underbrace{\begin{bmatrix}
        0_{(d+1)\times d} & 0_{(d+1)}\\
        I_d & 0_d\\
        0_{(n+1)\times d} & 0_{n+1}
    \end{bmatrix}}_{(2d+n+2)\times (d+1)}
     X
     \begin{bmatrix}
        0_{n\times (i-1)(P+1)} & f(L_0)\onehot{i}{i} & f(L_1)\onehot{i}{i} & \cdots & f(L_P)\onehot{i}{i} & 0_{n\times(n-i)(P+1)}
    \end{bmatrix}
     \\
     & ~ +
     \underbrace{\begin{bmatrix}
     0_{(2d+1)\times (P+1)} & \cdots &0_{(2d+1)\times (P+1)} \\
         S & \cdots & S \\
         ({2dB^2+B-\ln\epsilon_0})\onehot{n}{1}1_{1\times (P+1)}
         &
         \cdots &
         ({2dB^2+B-\ln\epsilon_0})\onehot{n}{n}1_{1\times (P+1)}
     \end{bmatrix}}_{(2d+n+2)\times n(P+1)}\\
     = & ~
     \begin{bmatrix}
         T_1 & T_2 & \cdots & T_n
     \end{bmatrix},
\end{align*}
where
\begin{align*}
    & ~1_{1\times (P+1)} := \begin{bmatrix}
        1 & 1 & \cdots & 1
    \end{bmatrix} \in\R^{1\times (P+1)}, \\
    & ~S :=
    \begin{bmatrix}
        -L_0^2 & -L_1^2 & \cdots & L_P^2
    \end{bmatrix}\in\R^{1\times (P+1)},\\
    & ~T_i :=
    \begin{bmatrix}
        2L_0x_i & 2L_1x_i & \cdots & 2L_Px_i \\
        2L_0y_i & 2L_1y_i & \cdots & 2L_Py_i \\
        f(L_0)x_i & f(L_1)x_i & \cdots & f(L_P)x_i\\
        -L_0^2 & -L_1^2 & \cdots & -L_P^2 \\
        ({2dB^2+B-\ln\epsilon_0})\onehot{n}{i} & ({2dB^2+B-\ln\epsilon_0})\onehot{n}{i} & \cdots & ({2dB^2+B-\ln\epsilon_0})\onehot{n}{i}
    \end{bmatrix}\in\R^{(2d+n+2)\times (P+1)}.
\end{align*}

Here $\epsilon_0$ is a parameter that we will designate later according to $\epsilon$.

Now construct $W_K,W_Q,W_V$ to be:
\begin{align*}
W_Q := & ~  I_{2d+n+2}, \\
W_K := & ~
I_{2d+n+2},\\
W_V^\top:= & ~
\begin{bmatrix}
    0_{d \times (d+1)} & I_d & 0_{d \times (n+1)}\\
\end{bmatrix}\in\R^{d \times (2d+n+2)}.
\end{align*}

Therefore, by \cref{def:hopfield}, the output becomes:
\begin{align*}
    \hop(\li_w(W);\li_x(X))
    = & ~
    W_V\li_x(X)
    \Softmax(\beta
    \li_x(X)^\top
    \li_w(W)
    ),
\end{align*}
where
\begin{align*}
\Softmax(
\li_x(X)^\top
\li_w(W)
)
= & ~
\Softmax
(\beta
\begin{bmatrix}
         T_1 & T_2 & \cdots & T_n
\end{bmatrix}^\top
\begin{bmatrix}
        W\\
        -1_{1 \times n}\\
        0_{d\times n}\\
        -1_{1 \times n}\\
        I_n
    \end{bmatrix}
).
\end{align*}

This is equivalent to:
\begin{align*}
(\li_x(X)^\top
\li_w(W)
)_{:,c}
= & ~
\begin{bmatrix}
    T_1^\top \\
    T_2^\top\\
    \vdots \\
    T_n^\top
\end{bmatrix}
\cdot
\begin{bmatrix}
w \\
-1\\
0_d\\
-1\\
\onehot{n}{c}
\end{bmatrix}\\
= & ~
\begin{bmatrix}
    M_{1,c} \\
    M_{2,c} \\
    \vdots \\
    M_{n,c}
\end{bmatrix},
\end{align*}
where
\begin{align*}
    M_{i,c} := & ~
    T_i^\top
    \cdot
    \begin{bmatrix}
    w\\
    -1\\
    0_d\\
    -1\\
    \onehot{n}{i}
    \end{bmatrix}\\
    = & ~
    \begin{bmatrix}
        2L_0x_i^\top w-2L_0y_i-L_0^2+ ({2dB^2+B-\ln\epsilon_0}) \one_{i=c} \\
        2L_1x_i^\top w-2L_1y_i-L_1^2+ ({2dB^2+B-\ln\epsilon_0}) \one_{i=c}\\
        \cdots \\
        2L_Px_i^\top w-2L_Py_i-L_P^2+ ({2dB^2+B-\ln\epsilon_0}) \one_{i=c}
    \end{bmatrix},
\end{align*}
where $i\in[n]$ and $c\in[n]$, and $\one_{i=c}$ represents the indicator function of $i=c$.

This means that
\begin{align*}
    & ~ \Softmax(
    \beta
    \li_x(X)^\top
\li_w(W)
    )_{:,c} \\
    = & ~
    \Softmax
    (
    \beta
    \begin{bmatrix}
    M_{1,c} \\
    M_{2,c} \\
    \vdots \\
    M_{n,c}
    \end{bmatrix}
    ) \\
    = & ~ \beta
    \sum_{i=1}^n\sum_{j=1}^P
    \frac{\exp{
    2L_{j}x_i^\top w-2L_{j}y_i-L_0^2+ ({2dB^2+B-\ln\epsilon_0}) \one_{i=c}}
    }{\sum_{i'=1}^n\sum_{j'=0}^P
    \exp{2L_{j'}x_{i'}^\top w-2L_{j'}y_{i'}-L_{j'}^2+ ({2dB^2+B-\ln\epsilon_0}) \one_{i=c}}
    }\onehot{nP}{(i-1)P+j}.
\end{align*}

Thus we have (without loss of generality, we ignore the $\beta$ parameter in $\Softmax$):
\begin{align*}
& ~ \hop(\li_w(W);\li_x(X))_{:,c}\\
= & ~
W_V\li_x(X)
    \Softmax(
\li_x(X)^\top
\li_w(W)
)_{:,c} \annot{$W_v$ only retrieves the $(d+2)$-th row in $T_i$}\\
= & ~
\begin{bmatrix}
    F_1 &\cdots & F_n
\end{bmatrix}
\sum_{i=1}^n\sum_{j=1}^P
    \frac{\exp{
    2L_{j}x_i^\top w-2L_{j}y_i-L_j^2+ ({2dB^2+B-\ln\epsilon_0}) \one_{i=c}}
    }{\sum_{i'=1}^n\sum_{j'=0}^P
    \exp{2L_{j'}x_{i'}^\top w-2L_{j'}y_{i'}-L_{j'}^2+ ({2dB^2+B-\ln\epsilon_0}) \one_{i=c}}
    }\onehot{nP}{(i-1)P+j}
\\
= & ~
\sum_{i=1}^n\sum_{j=0}^P
    \frac{\exp{
    2L_{j}x_i^\top w-2L_{j}y_i-L_j^2+ ({2dB^2+B-\ln\epsilon_0}) \one_{i=c}}
    }{\sum_{i'=1}^n\sum_{j'=0}^P
    \exp{2L_{j'}x_{i'}^\top w-2L_{j'}y_{i'}-L_{j'}^2+ ({2dB^2+B-\ln\epsilon_0}) \one_{i=c}}
    }f(L_j)x_i,
\end{align*}
where $F$ is:
\begin{align*}
F_i
:=
\begin{bmatrix}
f(L_0)x_i & f(L_1)x_i & \cdots & f(L_P)x_i
\end{bmatrix}.
\end{align*}

For every $i \in [n]$, if $i\neq c$, we have
\begin{align*}
    & ~ \sum_{j=0}^P
    \frac{\exp{
    2L_{j}x_i^\top w-2L_{j}y_i-L_j^2+ ({2dB^2+B-\ln\epsilon_0}) \one_{i=c}}
    }{\sum_{i'=1}^n\sum_{j'=0}^P
    \exp{2L_{j'}x_{i'}^\top w-2L_{j'}y_{i'}-L_{j'}^2+ ({2dB^2+B-\ln\epsilon_0}) \one_{i=c}}
    } \\
    = & ~
    \sum_{j=0}^P
    \frac{\exp{
    2L_{j}x_i^\top w-2L_{j}y_i-L_j^2
    }}{\sum_{i'=1}^n\sum_{j'=0}^P
    \exp{2L_{j'}x_{i'}^\top w-2L_{j'}y_{i'}-L_{j'}^2+ ({2dB^2+B-\ln\epsilon_0}) \one_{i=c}}
    }\\
    < & ~
    \sum_{j=0}^P
    \frac{\exp{
    2L_{j}x_i^\top w-2L_{j}y_i-L_j^2
    }}{\sum_{j'=0}^P
    \exp{2L_{j'}x_{i'}^\top w-2L_{j'}y_{i'}-L_{j'}^2+ ({2dB^2+B-\ln\epsilon_0})}}
    \annot{only taking the $i' = c$ part}
    \\
    < & ~
    \sum_{j=0}^P
    \frac{\exp{
    2dB^2+B
    }}{P
    \exp({2dB^2+B-\ln\epsilon_0})}
    =
    \epsilon_0.
\end{align*}

For $i = c$, since
\begin{align*}
    \sum_{i\neq c}^n\sum_{j=0}^P
    \frac{\exp{
    2L_{j}x_i^\top w-2L_{j}y_i-L_j^2+ ({2dB^2+B-\ln\epsilon_0}) \one_{i=c}}
    }{\sum_{i'=1}^n\sum_{j'=0}^P
    \exp{2L_{j'}x_{i'}^\top w-2L_{j'}y_{i'}-L_{j'}^2+ ({2dB^2+B-\ln\epsilon_0}) \one_{i=c}}
    }\leq (n-1)\epsilon_0,
\end{align*}
we have
\begin{align*}
    & ~
    \frac{\sum_{j=0}^P\exp{
    2L_{j}x_c^\top w-2L_{j}y_c-L_j^2+ ({2dB^2+B-\ln\epsilon_0})}
    }{\sum_{i'=1}^n\sum_{j'=0}^P
    \exp{2L_{j'}x_{i'}^\top w-2L_{j'}y_{i'}-L_{j'}^2+ ({2dB^2+B-\ln\epsilon_0}) \one_{i=c}}
    }\\
    = & ~
    \sum_{j=0}^P
    \frac{\exp{
    2L_{j}x_c^\top w-2L_{j}y_c-L_j^2+ ({2dB^2+B-\ln\epsilon_0})}
    }{\sum_{i'=1}^n\sum_{j'=0}^P
    \exp{2L_{j'}x_{i'}^\top w-2L_{j'}y_{i'}-L_{j'}^2+ ({2dB^2+B-\ln\epsilon_0}) \one_{i=c}}
    }\\
    \geq & ~  1-(n-1)\epsilon_0.
\end{align*}

Thus for the parts in the weighted sum output that corresponds to rows in $M_{:,c}$ in the attention score matrix, we have
\begin{align*}
    & ~ \|
    \sum_{j=0}^P
    \frac{\exp{
    2L_{j}x_c^\top w-2L_{j}y_c-L_j^2+ ({2dB^2+B-\ln\epsilon_0})}
    }{\sum_{i'=1}^n\sum_{j'=0}^P
    \exp{2L_{j'}x_{i'}^\top w-2L_{j'}y_{i'}-L_{j'}^2+ ({2dB^2+B-\ln\epsilon_0}) \one_{i=c}}
    }f(L_j)x_c
    -f(x_c^\top w-y_c)x_c\|_\infty \\
    = & ~ %
    \|
    \sum_{j=0}^P
    \frac{\exp{
    2L_{j}x_c^\top w-2L_{j}y_c-L_j^2+ ({2dB^2+B-\ln\epsilon_0})}
    }{\sum_{j'=0}^P\exp{
    2L_{j'}x_c^\top w-2L_{j'}y_c-L_{j'}^2+ ({2dB^2+B-\ln\epsilon_0})}}(f(L_j)x_c-f(x_c^\top w-y_c)x_c) \\
    & ~ \quad\quad \cdot
    \frac{\sum_{j'=0}^P\exp{
    2L_{j'}x_c^\top w-2L_{j'}y_c-L_{j'}^2+ ({2dB^2+B-\ln\epsilon_0})}}
    {\sum_{i'=1}^n\sum_{k=0}^P
    \exp{2L_{k}x_{i'}^\top w-2L_{k}y_{i'}-L_{k}^2+ ({2dB^2+B-\ln\epsilon_0}) \one_{i=c}}
    }\\
    & ~ -
    (1-\frac{\sum_{j'=0}^P\exp{
    2L_{j'}x_c^\top w-2L_{j'}y_c-L_{j'}^2+ ({2dB^2+B-\ln\epsilon_0})}}
    {\sum_{i'=1}^n\sum_{k=0}^P
    \exp{2L_{k}x_{i'}^\top w-2L_{k}y_{i'}-L_{k}^2+ ({2dB^2+B-\ln\epsilon_0}) \one_{i=c}}
    } )
    f(x_c^\top w-y_c)x_c
    \|_\infty\\
    \leq & ~ %
    \sum_{j=0}^P
    \frac{\exp{
    2L_{j}x_c^\top w-2L_{j}y_c-L_j^2+ ({2dB^2+B-\ln\epsilon_0})}
    }{\sum_{j'=0}^P\exp{
    2L_{j'}x_c^\top w-2L_{j'}y_c-L_{j'}^2+ ({2dB^2+B-\ln\epsilon_0})}}|f(L_j)-f(x_c^\top w-y_c)|\cdot d\|x_c\|_\infty
    \\
    & ~ -
    (1-\frac{\sum_{j'=0}^P\exp{
    2L_{j'}x_c^\top w-2L_{j'}y_c-L_{j'}^2+ ({2dB^2+B-\ln\epsilon_0})}}
    {\sum_{i'=1}^n\sum_{k=0}^P
    \exp{2L_{k}x_{i'}^\top w-2L_{k}y_{i'}-L_{k}^2+ ({2dB^2+B-\ln\epsilon_0}) \one_{i=c}}
    })|f(x_c^\top w-y_c)|\|x_c\|_\infty\\
    \leq & ~ %
    \sum_{j=0}^P
    \frac{\exp{
    2L_{j}x_c^\top w-2L_{j}y_c-L_j^2+ ({2dB^2+B-\ln\epsilon_0})}
    }{\sum_{j'=0}^P\exp{
    2L_{j'}x_c^\top w-2L_{j'}y_c-L_{j'}^2+ ({2dB^2+B-\ln\epsilon_0})}}|f(x_c^\top w-y_c)|\|x_c\|_\infty \\
    & ~+
    (n-1)\epsilon_0 B_f\|x_c\|_\infty\\
    = & ~ %
    \sum_{j=0}^P
    \frac{\exp{
    2L_{j}x_c^\top w-2L_{j}y_c-L_j^2}
    }{\sum_{j'=0}^P\exp{
    2L_{j'}x_c^\top w-2L_{j'}y_c-L_{j'}^2}}|f(L_j)-f(x_c^\top w-y_c)|\|x_c\|_\infty
    +
    (n-1)\epsilon_0 B_f\|x_c\|_\infty\\
    =& ~
    \sum_{j=0}^P
    \frac{\exp{
    -\beta(x_c^\top w -y_c -L_j)^2}
    }{\sum_{j'=0}^P\exp{
    -\beta(x_c^\top w -y_c -L_{j'})^2}}|f(L_j)-f(x_c^\top w-y_c)|\|x_c\|_\infty+
    (n-1)\epsilon_0 B_f\|x_c\|_\infty,
\end{align*}
where we define $B_f:=|f|$ as the bound for $f$.

For any $\epsilon_1>0$, set $\Delta L$ to be sufficiently small such that
\begin{align*}
    |f(x)-f(y)|\leq \epsilon_1,
\end{align*}
when $|x-y|\leq \Delta L$.

Then when $\beta$ is sufficiently large, we have
\begin{align*}
    \sum_{|L_i - (x_c^\top w -y_c)|>\Delta L}
    \frac{\exp{
    -\beta(x_c^\top w -y_c -L_j)^2}
    }{\sum_{j'=0}^P\exp{
    -\beta(x_c^\top w -y_c -L_{j'})^2}}\leq\epsilon_2,
\end{align*}
for any $\epsilon_2>0$.

Thus
\begin{align*}
    & ~ \sum_{j=0}^P
    \frac{\exp{
    -\beta(x_c^\top w -y_c -L_j)^2}
    }{\sum_{j'=0}^P\exp{
    -\beta(x_c^\top w -y_c -L_{j'})^2}}|f(L_j)-f(x_c^\top w-y_c)| \\
    = & ~
    \sum_{|L_i - (x_c^\top w -y_c)|>\Delta L}
    \frac{\exp{
    -\beta(x_c^\top w -y_c -L_j)^2}
    }{\sum_{j'=0}^P\exp{
    -\beta(x_c^\top w -y_c -L_{j'})^2}}|f(L_j)-f(x_c^\top w-y_c)| \\
    & ~ +
    \sum_{|L_i - (x_c^\top w -y_c)|\leq\Delta L}
    \frac{\exp{
    -\beta(x_c^\top w -y_c -L_j)^2}
    }{\sum_{j'=0}^P\exp{
    -\beta(x_c^\top w -y_c -L_{j'})^2}}|f(L_j)-f(x_c^\top w-y_c)| \\
    \leq & ~
    \epsilon_2\cdot 2B_f
    +
    \epsilon_1.
\end{align*}

This completes the proof.
\end{proof}

\begin{corollary}[In-Context GD of Hopfield Layer]
\label{cor:in-context-GD-hop}

Define
\begin{align*}
    X := \begin{bmatrix}
        x_1 & x_2 & \cdots & x_n \\
        y_1 & y_2 & \cdots & y_n
    \end{bmatrix} \in\R^{(d+1)\times n}
    \quad\text{and}\quad
    W :=
    \begin{bmatrix}
        w & w & \cdots & w
    \end{bmatrix}\in\R^{d\times n},
\end{align*}
where $x_i \in \R^d$ and $y_i \in \R$ are the input-output pairs.
$w\in \R^d$ is the linear coefficient we aim to optimize.
For any differentiable loss function $\ell:\R \to \R$, There exists a Hopfield layer $\hop$ with linear connections such that
\begin{align*}
    \|
    \hop(W;X)
    -
    \begin{bmatrix}
        \nabla \ell(w^\top x_1-y_1)x_1 &
        \nabla \ell(w^\top x_2-y_2)x_2 &
        \cdots
        &
        \nabla \ell(w^\top x_n-y_n)x_n
    \end{bmatrix}
    \|_\infty
    \leq
    \epsilon,
\end{align*}
for any $\epsilon > 0$.

\end{corollary}

\begin{proof}
    Replacing the continuous function $f$ in \cref{thrm:hopfield_sim_f} with $\nabla \ell$ completes the proof.
\end{proof}

\clearpage
\def\arxivfont{\rm}
\bibliographystyle{plainnat}

\bibliography{refs}

\end{document}